\documentclass[10pt,journal,compsoc]{IEEEtran}
\usepackage{graphicx}
\usepackage{psfrag}
\usepackage{amsmath,amssymb}
\usepackage{math}

\title{Human Motion Tracking by Registering an Articulated Surface to 3-D Points and Normals}
\author{Radu Horaud, Matti Niskanen, Guillaume Dewaele, and Edmond Boyer
\thanks{R. Horaud, G. Dewaele, and E. Boyer are with INRIA Grenoble Rh\^one-Alpes, 655 avenue de l'Europe, 38330 Montbonnot
  Saint-Martin, France.}
\thanks{M. Niskanen is with the Machine Vision Group, Infotech Oulu, University
  of Oulu, PO Box 4500, Finland. M. Niskanen acknowledges financial
  support from the Seppo Synjkangas foundation.}
  }

\begin{document}

\IEEEtitleabstractindextext{
\begin{abstract}
We address the problem of human motion tracking by registering a
surface to 3-D data. We propose a method that iteratively
computes two things: Maximum likelihood estimates for both the kinematic and
free-motion parameters of a kinematic human-body representation, as well as
probabilities that the data are assigned either to a body part, or to an
outlier cluster. We introduce a new metric between observed points
and normals on one side, and a parameterized surface on the other side,
the latter being defined as a 
blending over a set of ellipsoids. We claim that this metric is well
suited when one deals with either visual-hull or visual-shape
observations. We illustrate the method by
tracking human motions using sparse visual-shape data (3-D surface points
and normals) gathered from
imperfect silhouettes.
\end{abstract}

}
\maketitle

\section{Introduction} 
\label{section:introduction}

We address the problem of recovering articulated
human-motion parameters using 3-D data gathered from multiple image
sequences. We advocate that this
type of data has several advantages 
over 2-D data: it is less ambiguous and it is less sensitive to
self-occlusions. 
3-D features may be obtained
by stereo 
\cite{PlankersFua2003,Demirdjian2004,DDHF06}.
Alternatively one can capitalize on 3-D shape from silhouettes. In general,
2-D silhouettes are explicitly associated with a 3-D smooth surface
\cite{PlankersFua2003,IlicSalzmannFua2007,KRH08}.
Another way to use
silhouettes is to infer volumetric representations and
to fit articulated models to the voxels thus obtained
\cite{CheungBakerKanadeII2005,StarckHilton2003},
or to extract skeletal representations from these voxels \cite{BrostowECCV2004}.
It is also possible to infer 3-D surfaces
from silhouettes, namely the {\em visual hull} \cite{BF03} or the
{\em visual shape}
\cite{FLB06}. The advantage of surface-from-silhouettes is that it
allows the recovery of both 3-D surface points and
surface normals. Moreover, there is no matching process associated
with the reconstruction algorithm. Visual hull algorithms have been
proved to be extremely useful for recovering 3-D meshes which, in
turn, are very useful for surface rendering. The drawback is that they
need perfect silhouettes. Alternatively, visual shape methods (such as
the one described in \cite{FLB06}) produce
{\em sparse} surface descriptions (points and normals)
and can operate on imperfect silhouettes. 

In this short paper we present a new method for tracking human motion based
on fitting an articulated implicit surface to 3-D points and
normals. There are two important contributions. First, we introduce
a new distance between an observation (a point and a normal) and an
ellipsoid. We show that this can be used to define an implicit surface
as a blending over a set of 
ellipsoids which are linked together to from a kinematic
chain. Second, we
exploit the analogy between the distance from a set of observations
to the implicit surface and the negative log-likelihood of a
mixture of Gaussian distributions. This allows us to cast the problem
of implicit surface fitting into the problem of maximum likelihood (ML)
estimation with hidden variables. We argue that outliers are best
described by a uniform component that is added to the mixture, and we
formally derive the associated EM algorithm. 

Casting the data-to-model association problem into
ML with hidden variables has already been addressed in the past within
the framework of point registration
\cite{Wells97}, \cite{LuoHancock2001}, \cite{DDHF06}. In
\cite{Demirdjian2004} observations are deterministically and
iteratively assigned to each individual body part. We
appear to be the first to apply a probabilistic data-to-model
association framework to the problem of
fitting a blending of ellipsoids to a set of 3-D observations and to
explicitly model outliers within this context.

The remainder of the paper is organized as
follows. Section~\ref{section:object-modelling} describes how to
compute a distance between a 3-D observation (point and normal) and an
ellipsoid, and how to build an implicit articulated surface based on
this distance. Finally, it introduces the concept of a {\em
  probabilistic implicit surface}. 
Section~\ref{section:fitting} describes the formal derivation of the
EM algorithm in the case of implicit surface fitting.
Section~\ref{section:data-gathering} describes
experiments with simulated data and with
multiple-camera video data. 

\section{Modeling articulated objects}
\label{section:object-modelling} 

In order to model articulated objects such as human bodies we must
define a number of {\em open kinematic chains} that link the various
{\em body parts}. 
We will use ellipsoids for modeling these parts. Since
we measure 3-D data (point and orientation vectors) we must properly
define a metric that measures the discrepancy between the
data and the model. This metric will be used to define a distance
function as well as a probabilistic implicit surface. 

\subsection{The distance from a 3-D datum to an ellipsoid}

One convenient way to describe 3-D ellipsoids is to use an implicit
equation and to embed the
3-D Euclidean space into the 3-D projective space. 
This yields a 4$\times$4 full-rank symmetric matrix $\Qmat$:
\begin{equation}
\Qmat = 
\left[ \begin{array}{cc}
\Qbarre & \qvect \\
\qvect\tp & q_{44}
\end{array} \right]
=
\left[ \begin{array}{cc}
\Rmat\Dmat\Rmat\tp & -\Rmat\Dmat\Rmat\tp\tvect \\
-\tvect\tp\Rmat\Dmat\Rmat\tp & \tvect\tp\Rmat\Dmat\Rmat\tp\tvect
-1
\end{array} \right]
\label{eq:Qmat-definition}
\end{equation}
where $\Dmat=\diag[a^{-2},b^{-2},c^{-2}]$ is a $3\times$3 diagonal
matrix, $\Rmat$ is a $3\times 3$ rotation matrix and $\tvect$ is a 3-D
translation vector. In practice $b=c$ and 
we choose $a\geq b$. 
We denote by $\Xvect$ the homogeneous coordinates of a point $\xvect$ lying
onto the surface of the ellipsoid,
$\Xvect\tp\Qmat\Xvect=0$. The adjoint matrix 
$\Qmat^\star=\Qmat\tpinv$ defines the {\em
  dual 
  ellipsoid}. The family $\Pvect$ of planes which are tangent to the ellipsoid $\Qmat$
satisfy the constraint
$\Pvect\tp\Qmat\inverse\Pvect=0$ since $\Qmat\tp=\Qmat$. 
We denote by $\pvect$ the 3-D vector which is orthogonal to the plane
$\Pvect$ (therefore $\pvect$ is normal to the ellipsoid at point
$\xvect$):
$\pvect = \Qbarre \xvect + \qvect$,
where the notations of eq.~(\ref{eq:Qmat-definition}) are used.

\begin{figure}[t!]
\begin{center}
\includegraphics[width=0.95\columnwidth]{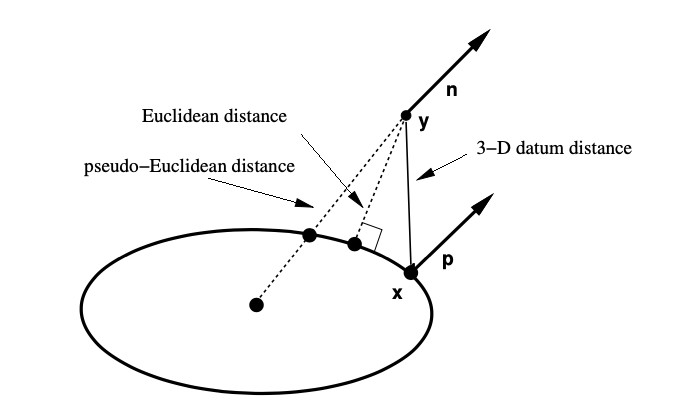}\\
\caption{The distance from the datum $\cc{Y}=(\yvect,\nvect)$ to the
  ellipsoid $\Qmat$ is estimated by seeking the point $\xvect\in\Qmat$
  such that the normal $\pvect$ at $\xvect$ is aligned with vector
  $\nvect$. }
\label{fig:Euclidean-distance}
\end{center}
\end{figure}

The
{\em algebraic distance} from a 3-D point $\Yvect$ to the
surface of an ellipsoid was used in \cite{PlankersFua2003}, defined by
$q(\Yvect)=\Yvect\tp\Qmat\Yvect$. The value of $q$ 
varies from $-1$ at the center of the ellipsoid, to $0$ on its surface, and then to
$+\infty$ outside the ellipsoid as the point is farther away from
the surface. 
The {\em Euclidean distance} from a point to an ellipsoid
requires to solve a six-degree polynomial. In
\cite{DDHF06} an approximation of the Euclidean
distance is used, i.e, a {\em pseudo-Euclidean distance}, as
shown on
Figure~\ref{fig:Euclidean-distance}.  

\begin{figure}[htb]
\begin{center}
\includegraphics[width=0.95\columnwidth]{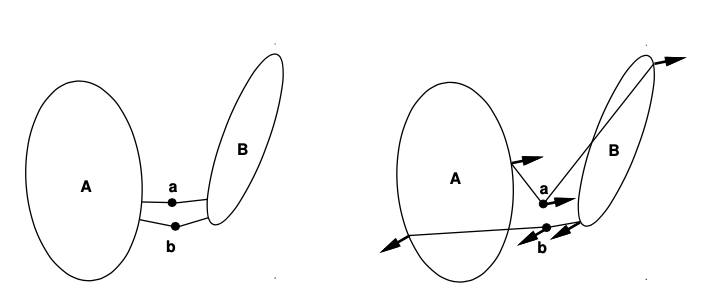}
\caption{The classical Euclidean distance
  from a point to an ellipsoid (left) does not assign a point to an
  ellipsoid in an unambiguous way. The 3-D datum distance (right)
  assigns without ambiguity $\avect$ to $\Amat$ and $\bvect$ to $\Bmat$.
} 
\label{fig:better-distance}
\end{center}
\end{figure}

An observation will be
refered to as a 3-D {\em datum} and consists of both a
3-D point and a 3-D vector. 
We define a metric between such a 3-D datum and an ellipsoid as
follows. Let $\Yvect\tp=(\yvect\tp\;1)$ be 
the homogeneous coordinates of an observed point, and let $\nvect$ be
a 3-D observed
vector. An observation or a 3-D datum is denoted by $\cc{Y} = (\yvect,\nvect)$.
We seek an ellipsoid point $\Xvect=(\xvect\tp\;1)$ under the
constraint that the vector $\pvect$ (normal to the ellipsoid at $\xvect$) is aligned with
$\nvect$, e.g. Figure~\ref{fig:Euclidean-distance}. In other words, we
seek an association between $\cc{X}=(\xvect,\pvect)$ and
$\cc{Y}=(\yvect,\nvect)$. Figure~\ref{fig:better-distance} compares the distance used in
this paper with the Euclidean distance from a point to an ellipsoid.

Let $d_{\cc{E}}(\cc{Y},\cc{X})$ be the
Euclidean distance from the datum-point $\yvect$ to the
ellipsoid-point $\xvect$ under the constraint that the datum-vector $\nvect$
and the ellipsoid-vector $\pvect$ are parallel:
\begin{equation}
d_{\cc{E}}(\cc{Y},\cc{X}) = \| \xvect - \yvect \|_2
\mbox{ with }
\nvect \times \pvect = 0
\label{eq:Eucl-dist-definition}
\end{equation}
where $\|\avect\|_2$ denotes the Euclidean norm.
We seek a solution for $\xvect$ under the constraints that $\pvect$ 
and $\nvect$ are parallel and yield the same orientation. Using
eq.~(\ref{eq:Qmat-definition}) we obtain the following set of constraints:
\begin{eqnarray}
\label{eq:quadratic-constraint}
\xvect\tp\Qbarre\xvect + 2\xvect\tp\qvect+q_{44} & = & 0 \\
\label{eq:orientation-constraint}
 \Qbarre \xvect + \qvect & = & \lambda \nvect
%\label{eq:vector-alignment}
%\pvect\tp\nvect & > & 0
\end{eqnarray}

From eq.~(\ref{eq:orientation-constraint}) we obtain
$\xvect = \Qbarre\;\inverse \left(\lambda \nvect - \qvect \right)$.
By substitution in eq.~(\ref{eq:quadratic-constraint}) we obtain two solutions for
$\lambda$. From $\pvect\tp\nvect > 0$ we have $\lambda>0$
and
$\lambda = (\nvect\tp\Rmat\Dmat\inverse\Rmat\tp\nvect)
^{-1/2}$.
Therefore, the point onto the ellipsoid where its normal $\pvect$ is aligned with
$\nvect$ is given by:
\begin{equation}
\label{eq:x-definition}
\xvect = \lambda \; \Rmat\Dmat\inverse\Rmat\tp\nvect
+  \tvect
\end{equation}
It will be convenient to use the 
Mahalanobis distance:
\begin{equation}
\label{eq:dist-to-ellipsoid-M}
d^2_{\cal M} (\yvect,\xvect) =  (\yvect -
  \xvect(\Rmat,\tvect,\nvect))\tp\mm{\Sigma}\inverse (\yvect - \xvect(\Rmat,\tvect,\nvect)) 
\end{equation}

\subsection{Kinematic chains and human-body modeling}
\label{subsection:kinematic-chains}

Articulated motion has a long history in mechanics, biomechanics,
robotics, and computer vision. A human body can be
described by a number of {\em open kinematic
chains} that share a common {\em root}. Such an open chain is composed
of a number of rigid objects and two
consecutive rigid objects in the chain are mechanically linked to form
a joint. Rotational (or spherical) joints are the most convenient
representations and they are well suited for human body modeling.
Each such
joint may have one, two, or three rotational degrees of freedom. Therefore,
within such a chain, a body part $\Qmat$ is
linked to a {\em root} body part $\Qmat_{r}$ through
a {\em constrained motion}, i.e., a kinematic chain with a number of
rotational degrees of freedom. Since each joint may have several
degrees of freedom, the total number of rotational parameters of a
chain is larger than the number of rigid parts composing the chain.
Moreover, the root
body part undergoes a {\em free motion} itself, i.e., a rigid
displacement with up to six degrees of freedom: three
rotations and three translations. 

Therefore the motion of a body-part (or ellipsoid) $\Qmat$
is composed of the root's free motion 
followed by the chain's constrained motion. We will denote the motion
of $\Qmat$ by the 4$\times$4 homogeneous matrix $\Tmat$ which in turn
is parameterized by the joint and free-motion parameter vector $\vv{\Lambda}$:

\begin{equation}\label{eq:motion-ellipsoid}
\Tmat(\vv{\Lambda}) = 
\left[
\begin{array}{cc}
\Rmat(\vv{\Lambda}) & \tvect(\vv{\Lambda}) \\
\zerovect & 1
\end{array}
\right]
\end{equation}

A complete human-body model may be described with five kinematic
chains that share a common root body-part. In this paper we use the
following {\em simplified}
human-body model. There are 14 body parts and 11 joints (2 ankles, 2
knees, 2 hips, 2 elbows, 2 shoulders, and a neck) with
$22$ rotational degrees of freedom (there are two degrees of freedom
for each joint). We also consider 3 rotations and 3
translations for the free motion. Hence, there is a total of 26
degrees of freedom.

\begin{figure}[t!]
\begin{center}
\begin{tabular}{cccc}
\includegraphics[width=0.21\columnwidth]{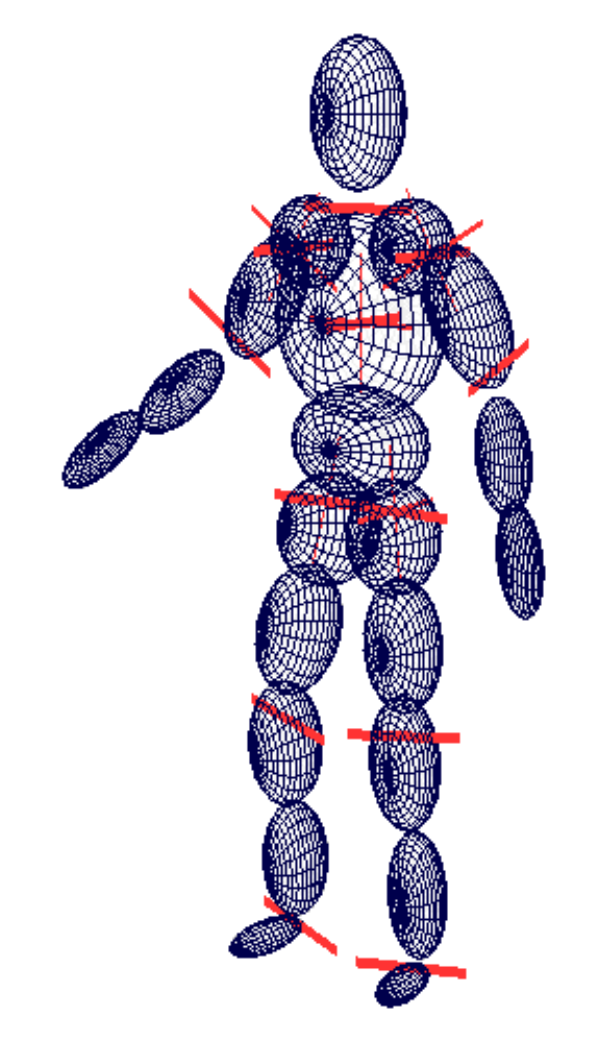} &
\includegraphics[width=0.21\columnwidth]{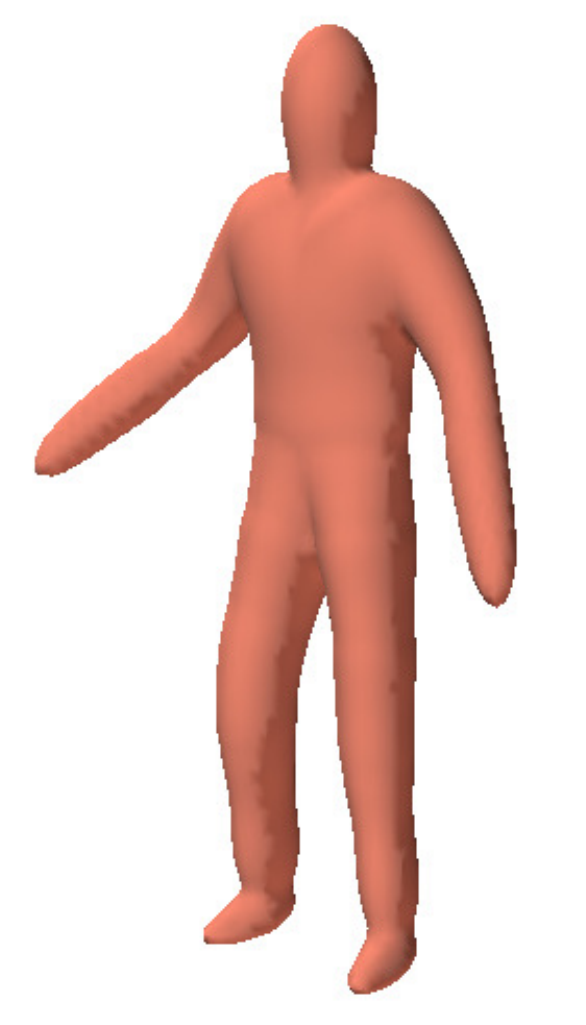} &
\includegraphics[width=0.16\columnwidth]{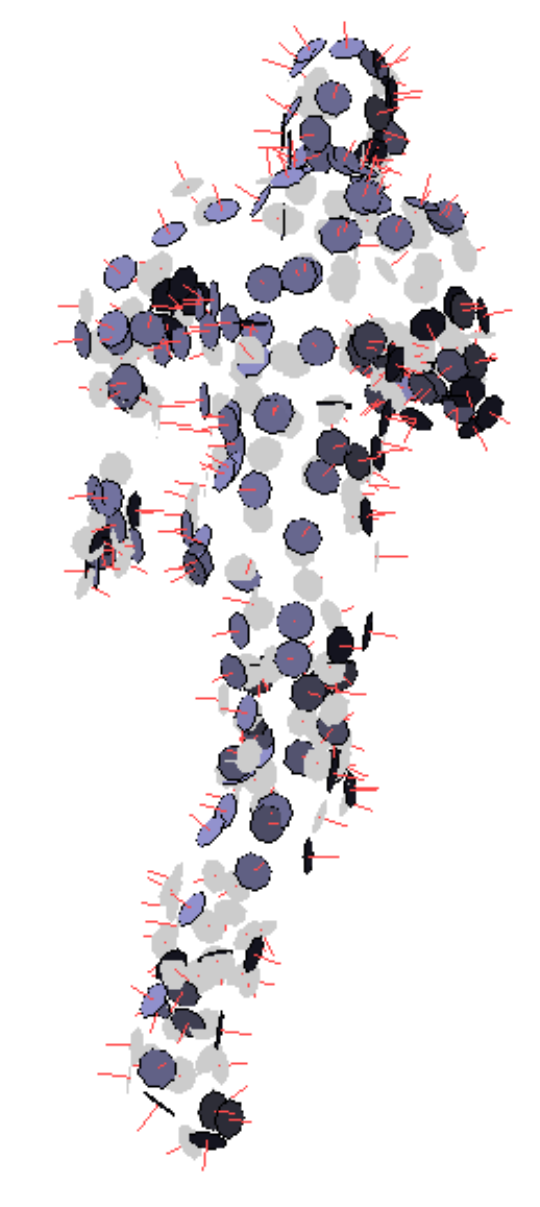} &
\includegraphics[width=0.16\columnwidth]{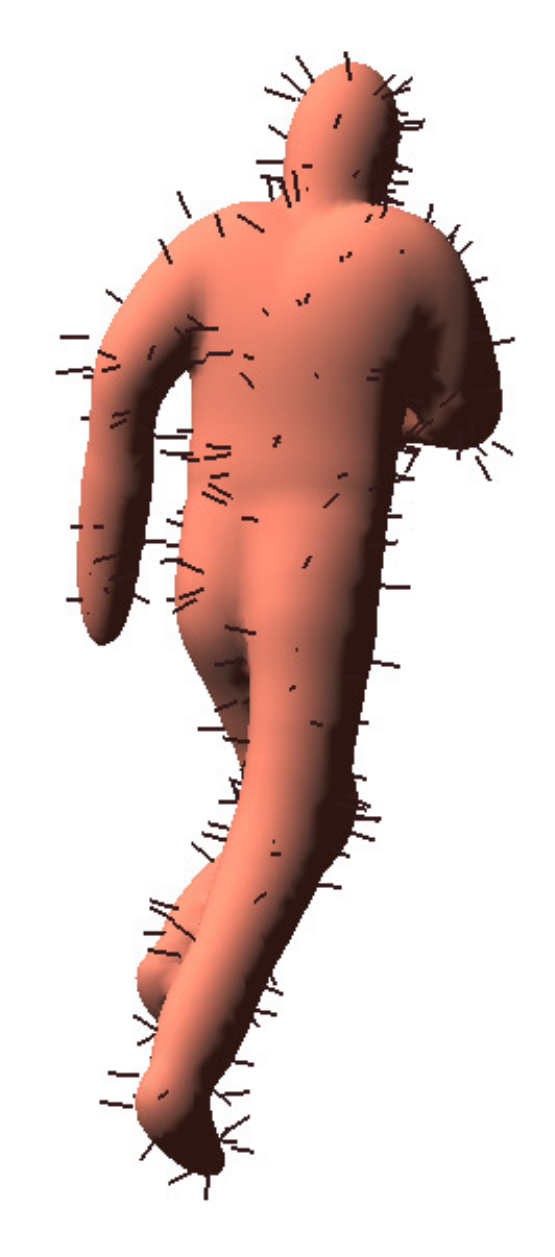} \\
\end{tabular}
\caption{From left to right: The set of 21 ellipsoids used to model 14
body parts with 11 joints and 2 rotations per joint. The implicit
surface defined as a blending of
these ellipsoids. A set of 3-D ``surface'' observations (points and
normals) and the articulated implicit surface that has been fitted to
these observations.}
\label{fig:model}
\end{center}
\end{figure}

As detailed above, body parts are
described by one or several ellipsoids: The feet and the thighs are
described by two ellipsoids, the torso is described by three
ellipsoids, and all the other body parts are described by a single
ellipsoid, hence there are 21 ellipsoids, e.g.,
Figure~\ref{fig:model}. The
body parts 
are denoted by $\Qmat_p, 1 \leq p \leq P$, where, for convenientce, $\Qmat_1$
corresponds to the common root body part. 

\subsection{Articulated implicit surfaces}
\label{subsection:articulated-implicit-surfaces}

In addition to using a collection of kinematically linked ellipsoids,
we will fuse them in order to define a smooth surface $\mathcal{S}$ over the entire
body. This surface will be described by the implicit equation
$f(\yvect)=C$ with $C=1$. The {\em contribution} of an ellipsoid $\Qmat_p$ is
defined by:
\begin{equation}
\label{eq:exp-distance}
f_p(\yvect) = \exp \left( -
  \frac{d^{2}_{\mathcal{M}}(\yvect,\xvect_p)}{\nu_p^2} \right)
\end{equation}
where $d_{\mathcal{M}}$ is the Mahalanobis distance from $\yvect$
to the ellipsoid defined by eq.~(\ref{eq:dist-to-ellipsoid-M}), the
point $\xvect_p$ lies onto the ellipsoid, and $\nu_p^2$ is 
a parameter that tunes the spatial influence of the
ellipsoid. 

An implicit surface is defined as a level set of the following
implicit function which is the fusion (or blending) of $P$ ellipsoids
verifying:
\begin{equation}
\label{eq:implicit-surface}
f(\yvect) = \sum_{p=1}^{P} f_p(\yvect)
\end{equation}
The class of implicit surfaces defined as above, i.e.,
$\yvect\in\mathcal{S}\Leftrightarrow f(\yvect)=C$, has successfully been used in
computer graphics and in computer vision in conjunction with
the
algebraic distance \cite{PlankersFua2003} and with the
pseudo-Euclidean distance
\cite{DDHF06}. Within this paper we extend this concept to the 
distance defined above. As it will be detailed below, this
is well suited to cast the problem of implicit surface fitting into
the framework of maximum likelihood in the presence of outliers.

In order to track articulated objects, the task at hand consists of
fitting the articulated implicit surface just described to a set of
observations. For this purpose, we first define a distance from a set
of observations to the implicit surface. We have to solve the
equation $f(\yvect)=1$, where $\yvect$ is, as before, an observed 3-D
point. One may notice that the first order Taylor expansion of $\ln a$
at $a=1$ is: $\ln a = a-1 + O(a^2)$. We retain the following
approximation of the distance
from a set of $I$ observations to the articulated implicit
surface formed by $P$ ellipsoids and parameterized by the kinematic variables $\vv{\Lambda}$:
\begin{equation}
F(\vv{\Lambda}) =  -\nu^2 \sum_{i=1}^{I} \ln 
    \sum_{p=1}^{P} \exp \left( -
  \frac{
 d_{ip}^2
}{\nu^2} \right) 
\label{eq:observations-to-surface}
\end{equation}
where:
\[
d_{ip}^2=(\yvect_i -
  \xvect_{ip}(\vv{\Lambda},\nvect_i))\tp\mm{\Sigma_p}\inverse (\yvect_i -
  \xvect_{ip}(\vv{\Lambda},\nvect_i))
\]
For convenience, we set $\nu=\nu_1 = \ldots = \nu_P$. The
notation $\xvect_{ip}$ means that the 3-D point $\xvect$ lies on
ellipsoid $p$ {\bf and} is associated with observation $i$.
It is
worthwhile to notice that, whenever a set of observations is closed to one of the
ellipsoids, the distance function is strictly equal to the sum of Mahalanobis
distances from each such observation to the ellipsoid. 

\subsection{Probabilistic implicit surfaces}

In this section we introduce a probabilistic interpretation of
eq.~(\ref{eq:observations-to-surface}). For this purpose we denote by
$z_i$ 
%$1\leq i\leq I$ 
a random variable that assigns an observation $i$
to an ellipsoid $p$, namely the notation $z_i=p$ means that the $i^{th}$
observation is assigned to the $p^{th}$ ellipsoid. There are as many
\textit{hidden} variables $z_i$ as observations:
$i\in\{1,\ldots,I\}$. The set of all the hidden variables is denoted
by $\mathcal{Z}=\{z_1,\ldots,z_I\}$.

The likelihood of an
observed 3-D point, given its assignment to an ellipsoid and given an observed
3-D normal, is drawn from a Gaussian distribution:
\begin{equation}
P(\yvect_i|z_i=p, \nvect_i) = \mathcal{N} (\yvect_i|
\xvect_{ip}(\vv{\Lambda},\nvect_i), \mm{\Sigma}_p)
\label{eq:observation-normal}
\end{equation}
In practice the data are corrupted by noise and by errors and
therefore there are observations which should not be assigned to an 
ellipsoid. For this reason we introduce an {\em outlier class} denoted by
$P+1$, and we assume
that the likelihood of an observation given that is classified as an outlier
is a uniform distribution over the volume $V$ of the working space:
\begin{equation}
P(\yvect_i|z_i=P+1, \nvect_i) = \mathcal{U} (\yvect_i|V,0) = \frac{1}{V}
\label{eq:observation-uniform}
\end{equation}
Therefore, one can write the likelihood of an observation as a mixture
of $P$ Gaussian components and one uniform component:
\begin{equation}
P(\yvect_i|\nvect_i) = \sum_{p=1}^{P+1} \pi_p P(\yvect_i|z_i=p,
\nvect_i)
\label{eq:likelihood-y}
\end{equation}
The notation:
\begin{equation}
\pi_p=P(z_i=p| \nvect_i)
\label{eq:prior}
\end{equation}
denotes the priors, the proportions, or the mixing parameters, and they
obey the obvious constraint
$\sum_{p=1}^{P+1}\pi_p=1$. Notice that this prior probability depends
on the observed vector $\nvect_i$. In this paper we do not treat these
observed vectors as random variables.
By assuming independent and identically
distributed observations one can write the joint likelihood of all the
observations as: 
\[
P(\mathcal{Y}_1,\ldots,\mathcal{Y}_I) = P(\yvect_1,\nvect_1,\ldots,\yvect_I,\nvect_I) = \prod_{i=1}^{I}
P(\yvect_i|\nvect_i)P(\nvect_i)
\]
Using Bayes' formula and the equations above, the negative log-likelihood writes:
\begin{eqnarray}
&-&\ln P_{\vv{\Lambda}}(\mathcal{Y}_1,\ldots,\mathcal{Y}_I) =
\nonumber \\
&-& \sum_{i=1}^{I} \ln \left(
%  \pi_{P+1}\mathcal{U} (\yvect_i|V,0)+
\sum_{p=1}^{P}  \pi_p \mathcal{N} (\yvect_i|
\xvect_{ip}(\vv{\Lambda},\nvect_i), \mm{\Sigma}_p) \right. \nonumber \\
&+& \left. \pi_{P+1} \mathcal{U}(\yvect_i|V,0)
  \phantom{\sum_{p=1}^{P}} \right) 
\label{eq:likelihood-observations}
\end{eqnarray}
Notice that there is a strong analogy between
eqs.~(\ref{eq:observations-to-surface}) and
(\ref{eq:likelihood-observations}): the former is a distance between a
set of $I$ observations and an articulated implicit surface while the
latter is the joint likelihood of the same observation set, where the
likelihood is a mixture of $P$ normal
distributions plus a uniform distribution that captures
the bad observations. This analogy will be exploited in the next section in order
to cast the estimation of the kinematic parameters in the framework of maximum
likelihood with hidden variables via the EM algorithm.

\section{Robust tracking with the EM algorithm} 
\label{section:fitting}

Because of the presence of the hidden variables,
$\mathcal{Z}=\{z_1,\ldots,z_I\}$, the maximum-likelihood estimation
problem, i.e.,
eq.~(\ref{eq:likelihood-observations}) does not have a simple
solution. The most convenient way to maximize the likelihood of a mixture of
distributions is to use the EM algorithm. The latter has been
thoroughly studied in the context of data clustering \cite{FraleyRaftery2002}.
In this paper we formally
derived an expectation-maximization scheme in the particular case of
{\em robustly} fitting an implicit surface to 
a set of 3-D observations. It is worthwhile to notice that the
formulae below are valid independently of the distance function being
used, i.e., Figure~\ref{fig:Euclidean-distance}.

First we derive the posterior
class probabilities conditioned by the observations, namely:
\[
P(z_i=p|\yvect_i,\nvect_i) =
\frac{P(z_i=p,\yvect_i,\nvect_i)}{P(\yvect_i,\nvect_i)}
\]
We denote these posteriors by $t_{ip}$ and with the notations
introduced in the prevous section we obviously obtain:
\begin{equation}
t_{ip} = \frac{ \pi_p P(\yvect_i|z_i=p,\nvect_i)}{P(\yvect_i|\nvect_i)}
\label{eq:posteriors}
\end{equation}

Second we consider the joint probability of the set of observations
$\mathcal{Y}=\{\mathcal{Y}_1,\ldots,\mathcal{Y}_I\}$ and 
of their assignments $\mathcal{Z}$ which yield the following expression:
\[
P(\mathcal{Y},\mathcal{Z})=\prod_{i=1}^{I} \prod_{p=1}^{P+1} \big(
P(\yvect_i|z_i=p,\nvect_i)P(z_i=p|\nvect_i)
\big)^{\delta_p(z_i)} P(\nvect_i)
\]
with the following definition for the function $\delta_p(z_i)$:
\[
\delta_p(z_i) = \left\{ \begin{array}{ccc} 1 & \text{if} & z_i=p\\ 0
    && \text{otherwise} \end{array} \right.
\]

Third we derive the expression of the {\em conditional
  expectation} of the log-likelihood taken over $\mathcal{Z}$, which in this case yields:
\begin{eqnarray*}
E[\ln P(\mathcal{Y},\mathcal{Z}) | \mathcal{Y} ] = \\
\sum_{i=1}^{I}
\sum_{p=1}^{P+1}
E[\delta_p(z_i)| \mathcal{Y} ] ( \ln P(\yvect_i|z_i=p,\nvect_i) \\
+ \ln \pi_p ) + (P+1)\sum_{i=1}^{I} \ln P(\nvect_i)
\end{eqnarray*}
One may notice that:
\[
E[\delta_p(z_i)| \mathcal{Y} ] = \sum_{p=1}^{P+1}
\delta_p(z_i=p)P(z_i=p|\yvect_i,\nvect_i) = t_{ip}
\]

By using the expressions of the normal and uniform distributions, and
by grouping constant terms we obtain:
\begin{eqnarray}
&& E[\ln P(\mathcal{Y},\mathcal{Z}) | \mathcal{Y} ] = \nonumber \\
&-& \frac{1}{2}
\sum_{i=1}^{I}
\left(
\sum_{p=1}^{P} t_{ip} \big( 
(\yvect_i -
  \xvect_{ip})\tp\mm{\Sigma_p}\inverse (\yvect_i -
  \xvect_{ip})
\right. 
\nonumber \\
\label{eq:EM}
& + & 
\left. \ln\det\mm{\Sigma_p} - \ln\pi_p\big) +
t_{iP+1}\ln\pi_{P+1} \vphantom{\sum_{p=1}^{P}}
\right) 
+ \text{const}
\end{eqnarray}

The maximization of eq.~(\ref{eq:EM}) (or equivalently the
minimization of its negative) will be carried out via the EM algortihm
(expectation-maximization). There are, however three notable differences
between the standard EM for Gaussian mixtures \cite{Bishop2006} and
our formulation:
\begin{itemize}
\item We added a {\em uniform-noise} component to the mixture. The
  role of this
  component is to ``capture'' outliers and hence to avoid that they influence the
  estimation of the model parameters;
  
\item The means of the Gaussian components, $\xvect_{ip}$ are
  parameterized by the kinematic parameters that control the
  articulated motion of each ellipsoid; This has an important
  consequence because the M-step of the algorithm will incorporate
  a non-linear minimization procedure over the kinematic joints.

\item At the start of the algorithm each observation is associated
  with  all the ellipsoids. As
  the algorithm proceeds, each observation is eventually associated
  with one of the ellipsoids. Due to occlusions, missing data, etc.,
  there may be ellipsoids with no associated observation.
Therefore, there is a risk that the
  corresponding covariance becomes infinitely small. To overcome this
  problem {\em we use a unique covariance matrix common to all the
  densities in the mixture}.
\end{itemize}
Since we formally derived
eq.~(\ref{eq:EM}), the EM algorithm outlined below guarantees
likelihood maximization.
To summarize, the advantages of this formulation
are (i)~fast convergence properties of EM and (ii)~the
fact that \emph{it 
minimizes the negative log-likelihood} given by
eqs.~(\ref{eq:likelihood-observations}) and (\ref{eq:EM}).
In practice, the following EM procedure can be used for robust
tracking of an articulated implicit surface:

%fbox{\begin{minipage}{0.90\textwidth}
%\begin{description} \item[Robust tracking with EM:] \end{description}
\begin{enumerate}
\item {\bf Initialization.} Compute the locations of the ellipsoid points $\xvect_{ip}^{(q)}$ from the
  current kinematic parameters $\vv{\Lambda}^{(q)}$ using
  eq.~(\ref{eq:x-definition}). Similarly, initialize the covariance
  matrix $\mm{\Sigma}^{(q)}$ common to all the ellipsoids. Initialize the
  priors, or the mixing parameters
  $\pi_1^{(q)}=\ldots=\pi_{P+1}^{(q)}=1/(P+1)$;
\item {\bf E step.} Evaluate the posterior probabilities $t_{ip}^{(q)}$ using the
  current parameter values, through
   eq.~(\ref{eq:posteriors});
\item {\bf M step.} Estimate new values for the kinematic parameters $\vv{\Lambda}^{(q+1)}$:
\begin{eqnarray*}
%& \vv{\Lambda}^{(q+1)} = & \\
\arg \min_{\vv{\Lambda}}
\frac{1}{2}
\sum_{i=1}^{I}
\sum_{p=1}^{P} t_{ip}^{(q)}
(\yvect_i -
  \xvect_{ip}(\vv{\Lambda}))\tp{\mm{\Sigma}^{(q)}}\inverse (\yvect_i -
  \xvect_{ip}(\vv{\Lambda}))
\end{eqnarray*}
Update the covariance matrix %$\mm{\Sigma}^{(q+1)}$ 
and the priors:
\begin{eqnarray*}
\mm{\Sigma}^{(q+1)} & & = \\
&& \frac{1}{\sum_{p=1}^{P}T_p} \sum_{i=1}^{I}\sum_{p=1}^{P}
 t_{ip}^{(q)}(\yvect_i -
  \xvect_{ip}(\vv{\Lambda}^{(q+1)}))\\
&& (\yvect_i - \xvect_{ip}(\vv{\Lambda}^{(q+1)}))\tp\\
\pi_p^{(q+1)} && =
%\frac{N_p}{I}\\
\frac{1}{I} \sum_{i=1}^{I}  t_{ip}^{(q)}
% N_p &=& \sum_{i=1}^{I}  t_{ip}^{(q)}
\end{eqnarray*}
\item {\bf Maximum likelihood.} Evaluate the log-likelihood, i.e.,
  eq.~(\ref{eq:likelihood-observations}) and check for convergence.
\end{enumerate}
%\end{minipage}}

\section{Experimental results}
\label{section:data-gathering}

The tracking algorithm described in the previous section is not tight to
any particular method for extracting the 3-D data. In
practice we used 3-D points and normals that lie on the {\em visual
  shape} \cite{FLB06}. Notice that the visual-shape
algorithm does not require perfect silhouettes and provides as output
a sparse set of 3-D points and normals,
not a 3-D mesh. The
visual-shape method uses multiple-view epipolar geometry in
conjunction with the 
assumption that the object's surface is locally continuous and twice
differentiable (see 
\cite{FLB06} for details).

\subsection{Experiments with simulated data} 
\label{section:experiments-simulated}

We used an animation package to build a human body, to simulate
various human motions, and to render image silhouettes. The simulator
uses its own shape representation, that is different than ours,
but it allows the user to define her/his own kinematic
model. Therefore we used the same kinematic model
with the same number of degrees of freedom as
the one described in section~\ref{subsection:kinematic-chains}.
Nevertheless, we have not attempted to finely tune the shape
parameters of our model to the simulated data.
We simulated a setup
composed of seven calibrated
cameras.
Sequences of image silhouettes
were generated from the 3-D model and rendered with the cameras'
parameters. We computed 3-D points and normals from these silhouettes
and then we 
applied our method to these data sets. The articulated-motion
parameters were recovered using our tracker. In order to assess the
merits of the data-to-model fitting process, we added 20\% of outliers
uniformly distributed in the volume of the working space. 
These simulations allowed us to
(i)~assess the quality of the tracker with
respect to ground-truth joint trajectories, (ii)~analyse the behavior
of the method in the
presence of various perturbations that alter the quality of the data, (iii)~quantify the
merit of using both 3-D points and normals instead of using only points, and (iv)~determine the optimal
number of observations needed for reliable estimation of articulated
motion. 

%\begin{figure}[htb]
%\begin{center}
% Here STARTS the simulated experiment under the form of 3x3 images
%\begin{center}
%\begin{tabular}{cccccc}
%\includegraphics[width=0.15\columnwidth]{sil33-6-eps-converted-to.pdf} &
%\frame{\includegraphics[height=20.0mm]{pos33_6-eps-converted-to.pdf}} &
%\includegraphics[width=0.15\columnwidth]{sil34-6-eps-converted-to.pdf} &
%\frame{\includegraphics[height=20.0mm]{pos34_6-eps-converted-to.pdf}} &
%\includegraphics[width=0.15\columnwidth]{sil90-6-eps-converted-to.pdf} &
%\frame{\includegraphics[height=20.0mm]{pos90_6-eps-converted-to.pdf}} \\
%\end{tabular}
%\caption{The silhouettes associated with frames 33, 34, and 90 of
%one of the simulated image sequences, and the articulated and implicit
%surfaces fitted to the corresponding 3-D data. 
%The "needles" illustrate the normals associated
%with the 3-D observations.}
%\label{fig:synthdata}
%\end{center}
%\end{figure}

\begin{figure}[t!]
\begin{center}
% Here STARTS the simulated experiment under the form of 3x3 images
%\begin{center}
\begin{tabular}{cc}
\includegraphics[width=0.45\columnwidth]{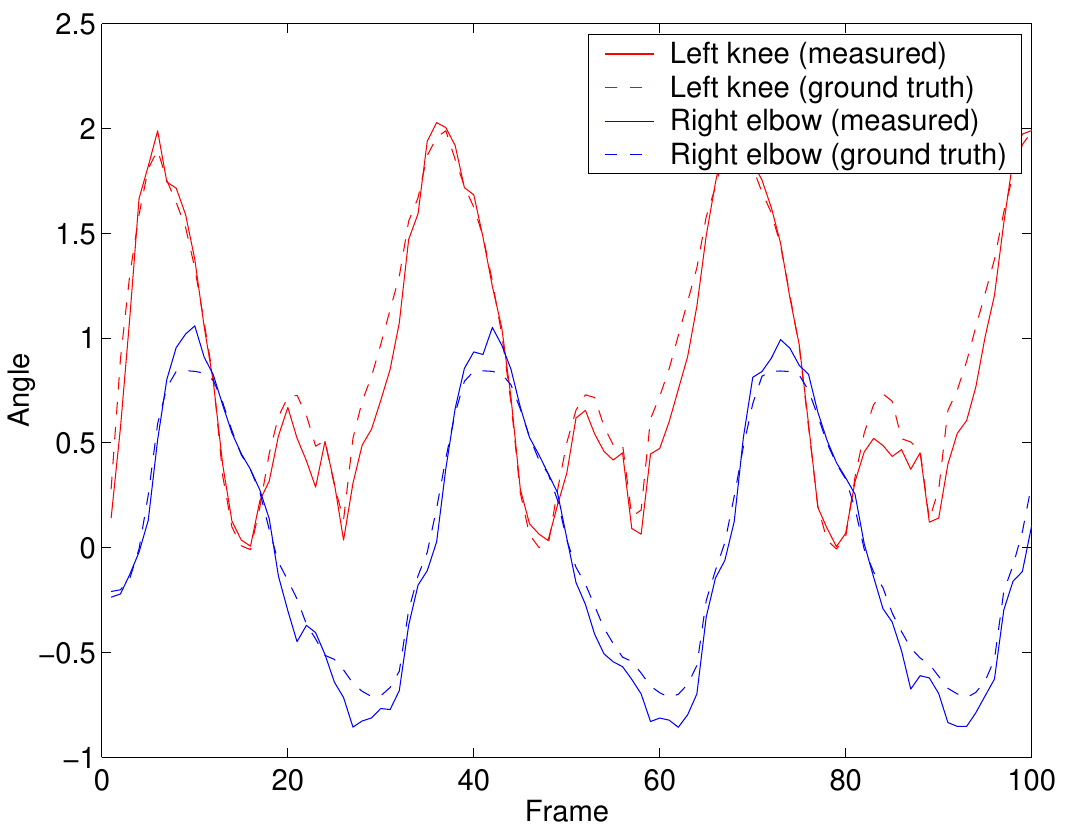} &
\includegraphics[width=0.45\columnwidth]{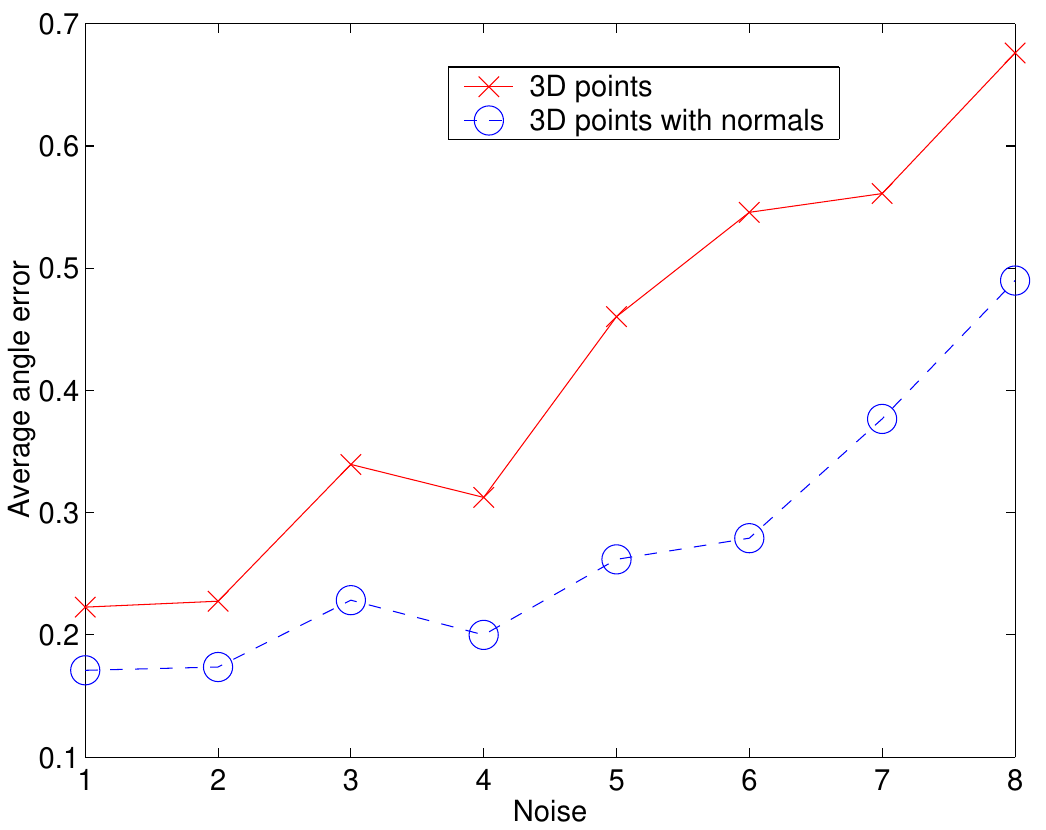} \\
\includegraphics[width=0.45\columnwidth]{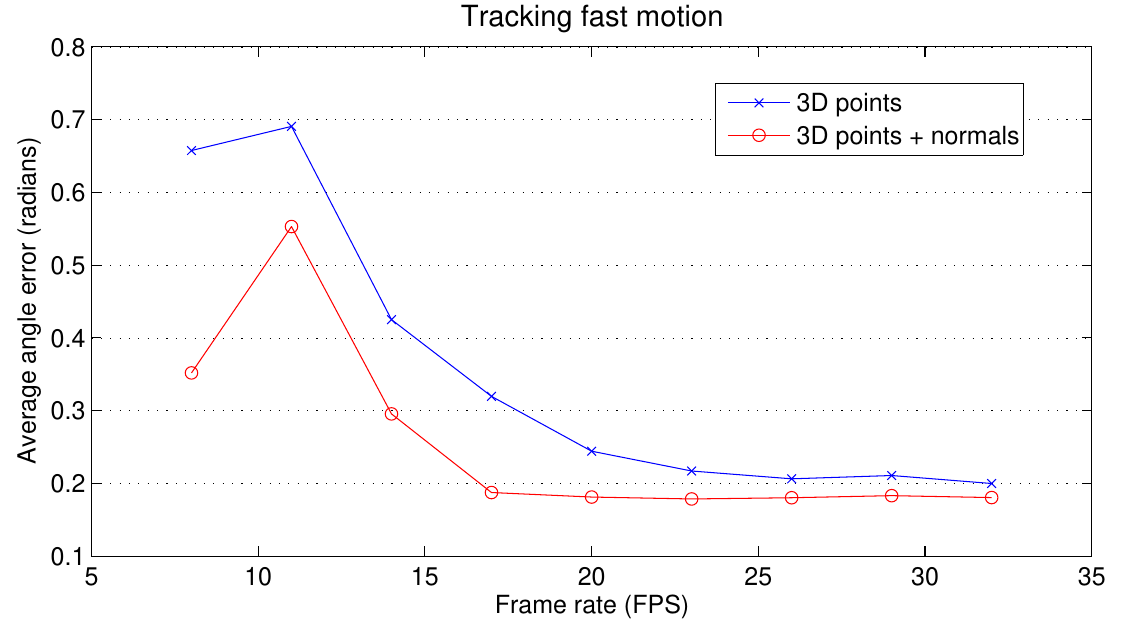} &
\includegraphics[width=0.45\columnwidth]{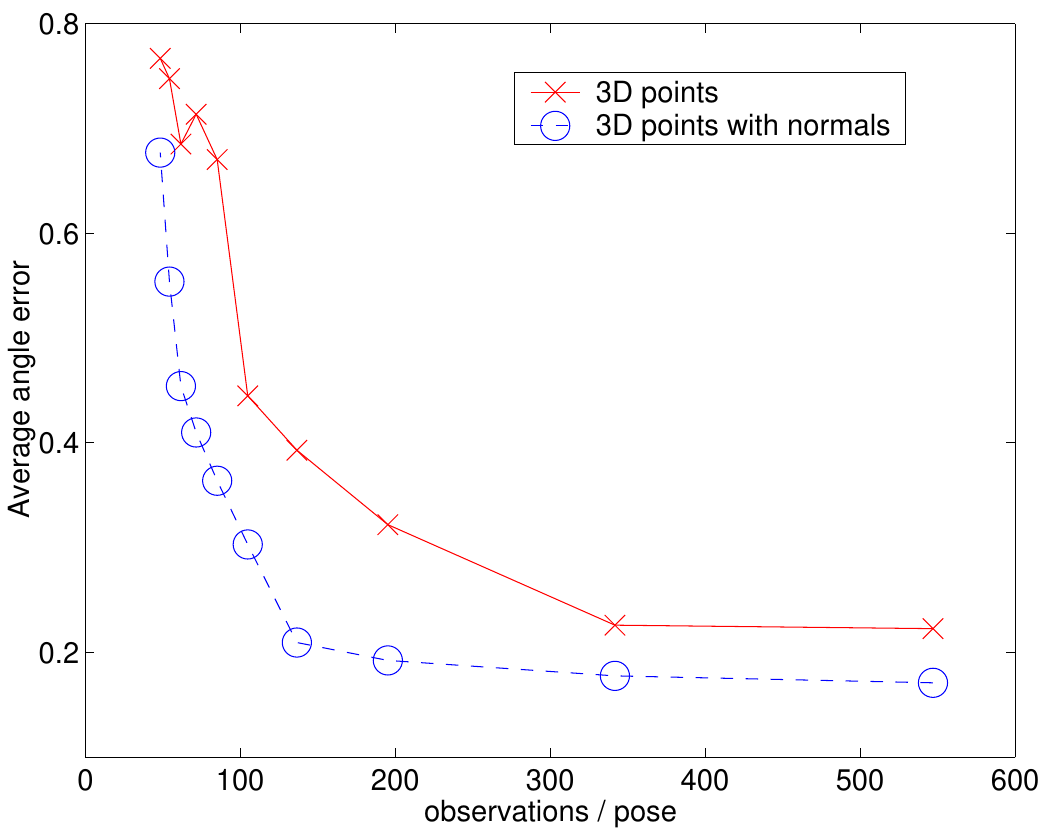} \\
\end{tabular}
\caption{The error between simulated angle values and
  estimated ones (measured in radians), from left ot right: Ground-truth and
  measured trajectories over 100 frames. Average angle
  error as a function of silhouette noise for points and for points
  and normals. Average angle error as a function of frame
rate. Average angle error as a function of the number of
observations being used.} 
\label{fig:graphs}
\end{center}
\end{figure}
\begin{figure*}[h]
\begin{center}
\begin{tabular}{ccccccc}
% first and fourth images
\includegraphics[width=0.12\textwidth]{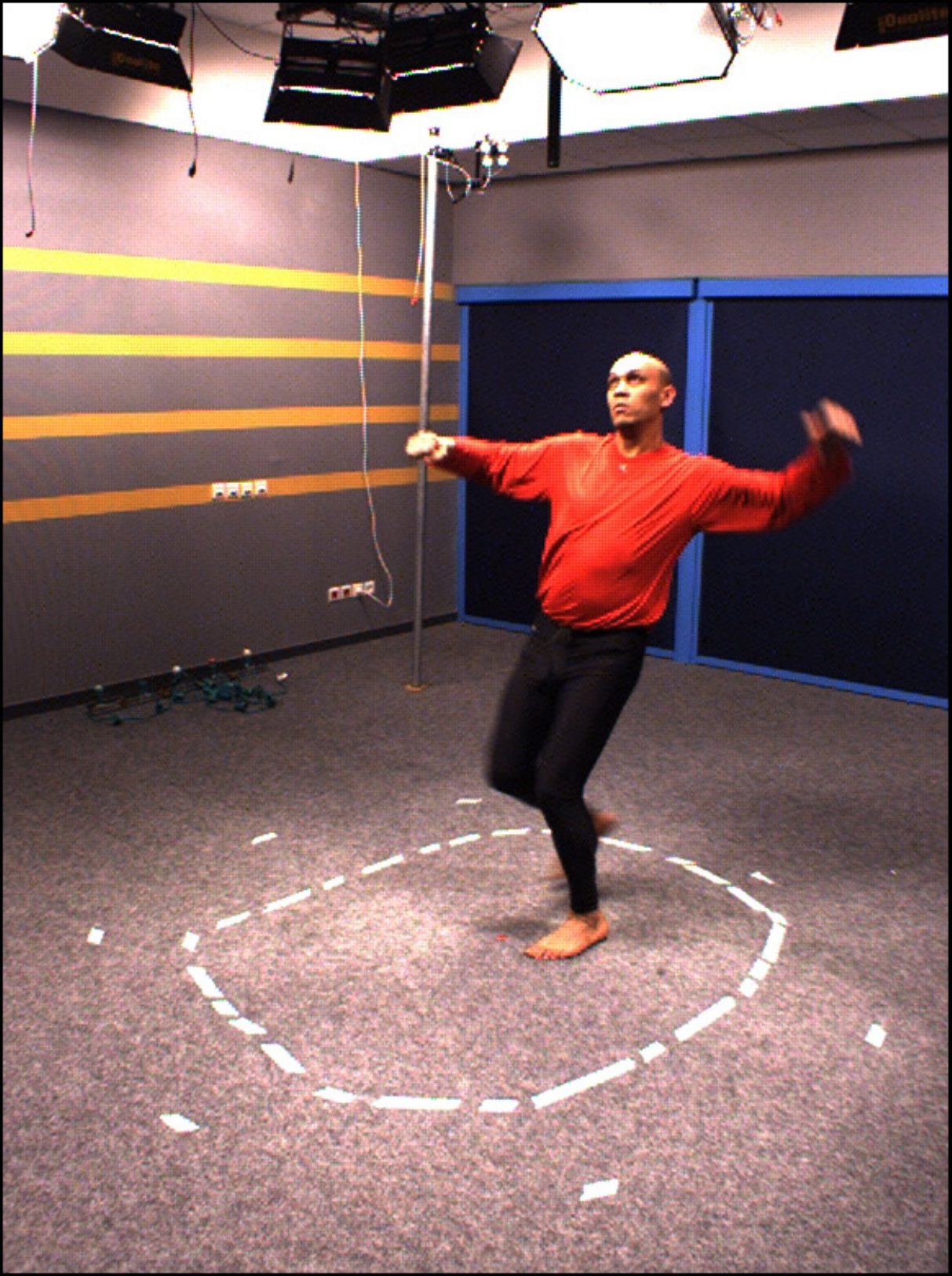} &
\includegraphics[width=0.12\textwidth]{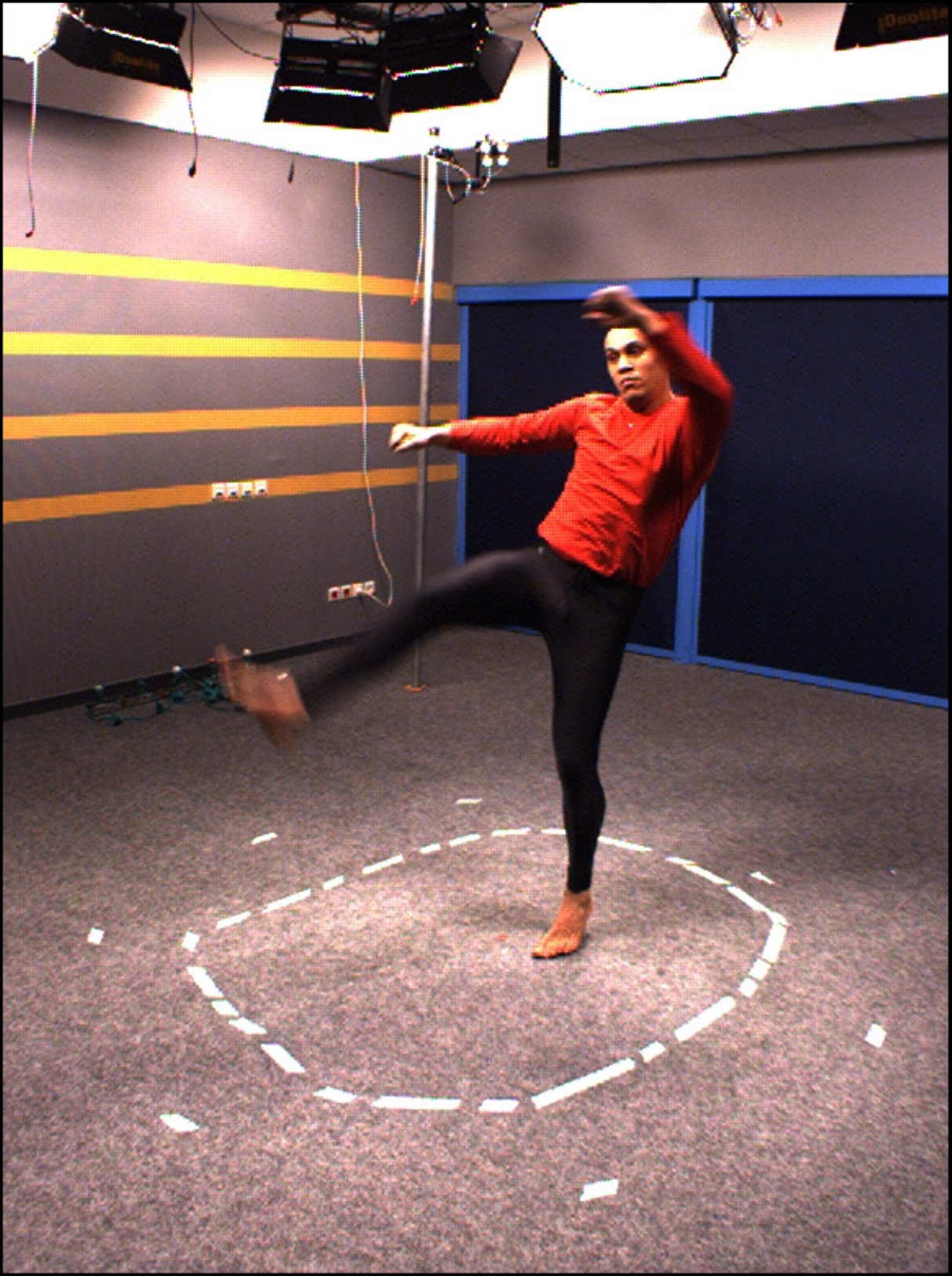} &
\includegraphics[width=0.12\textwidth]{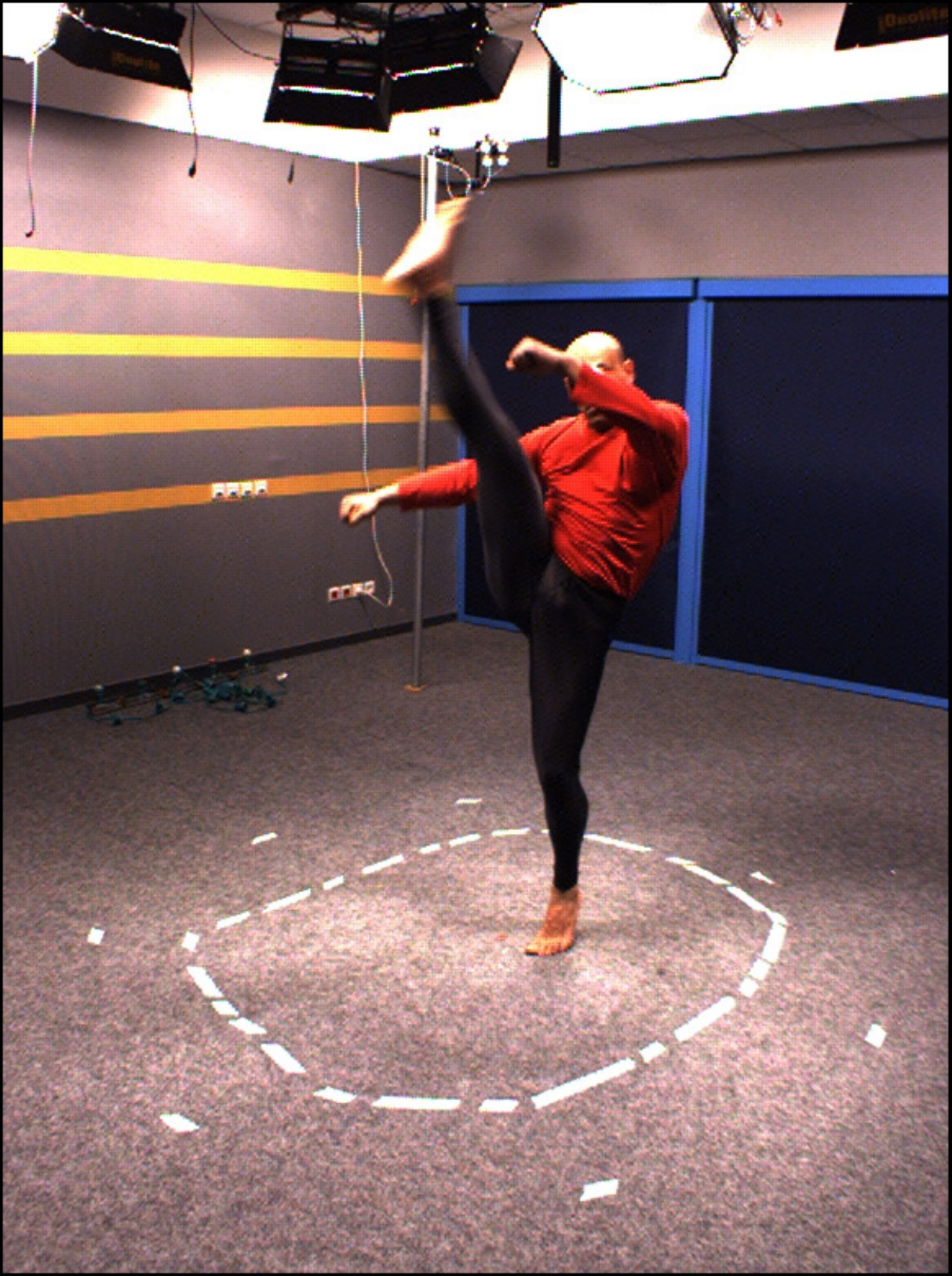} &
\includegraphics[width=0.12\textwidth]{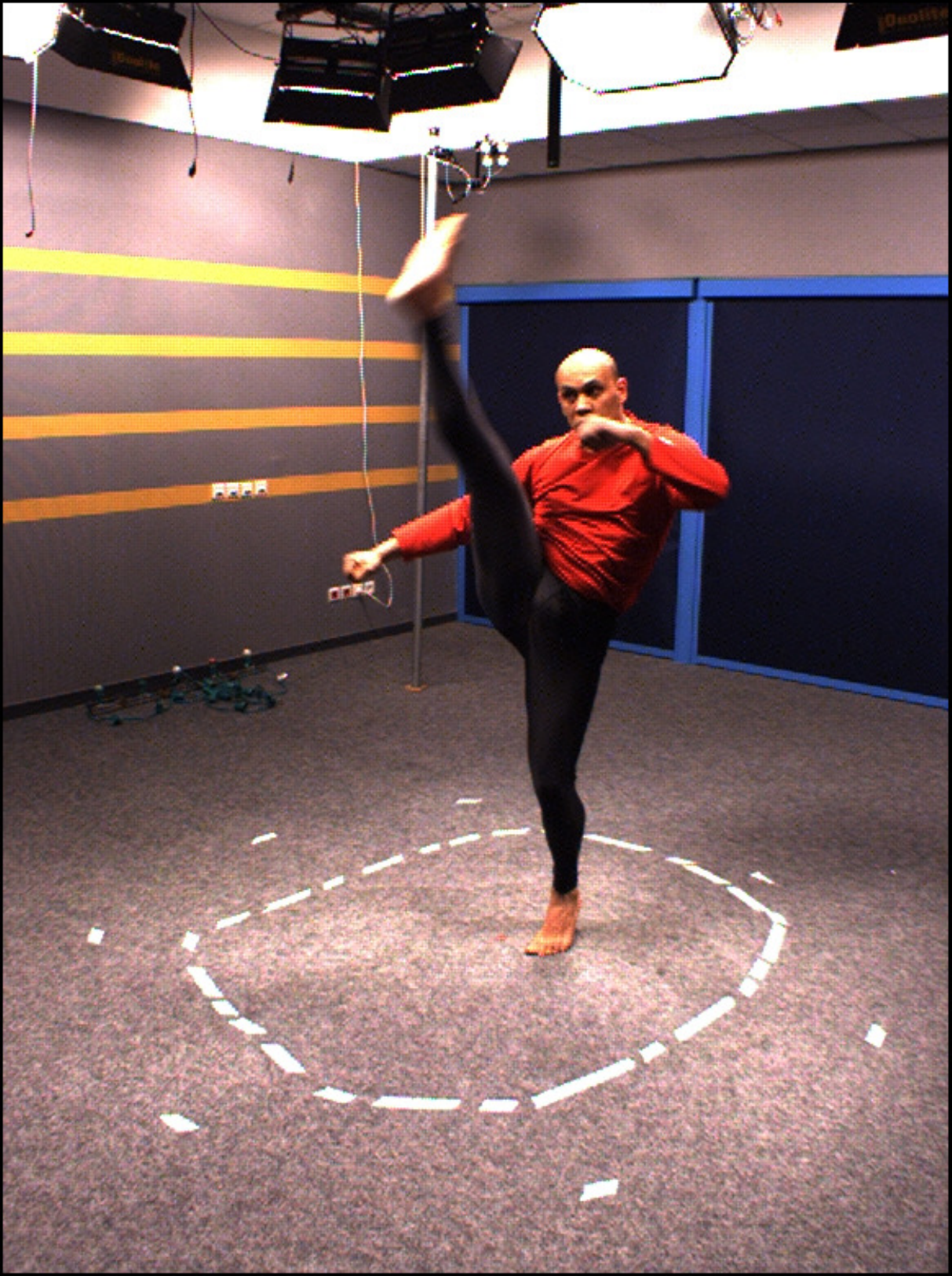} &
\includegraphics[width=0.12\textwidth]{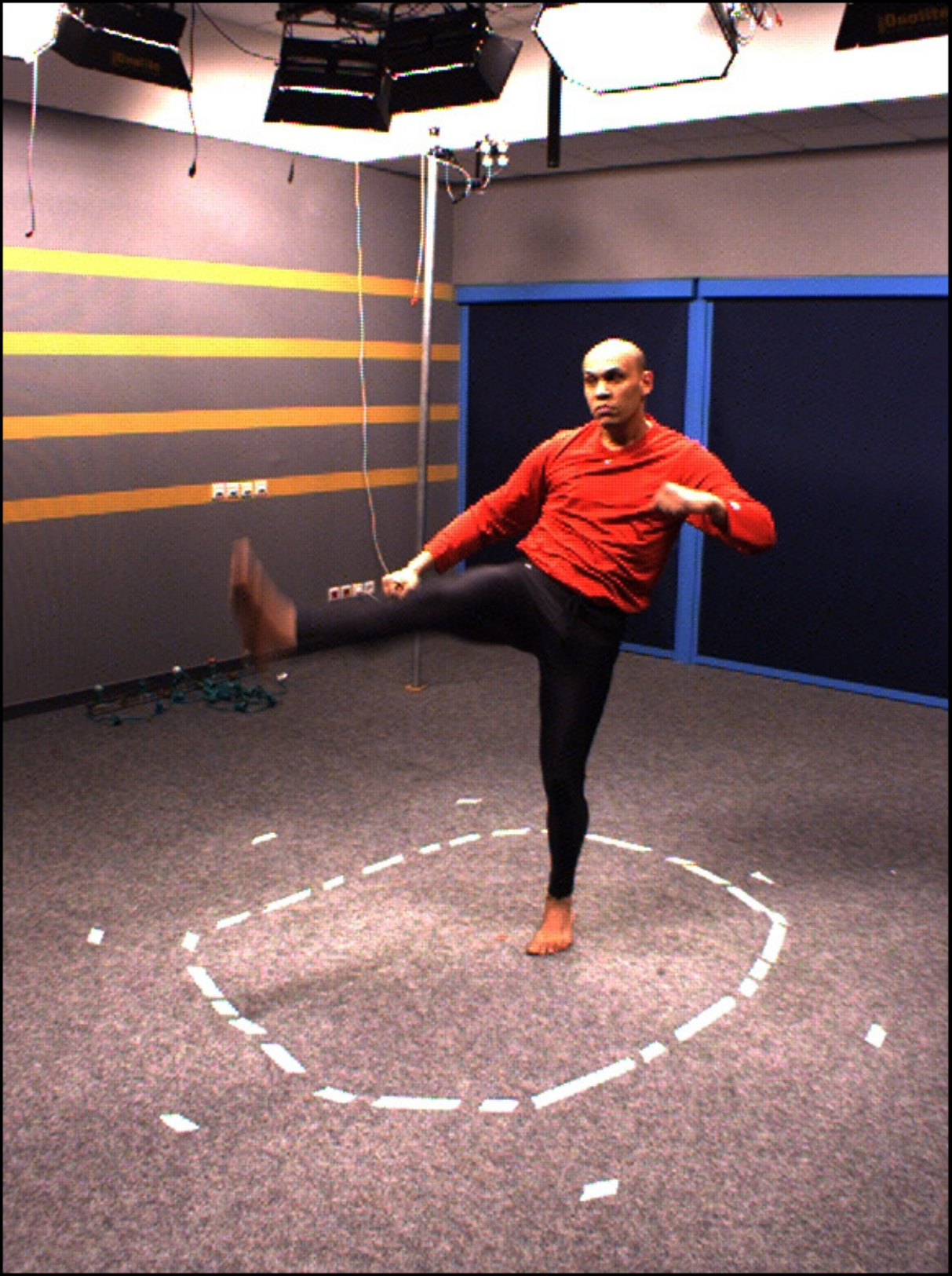}&
\includegraphics[width=0.12\textwidth]{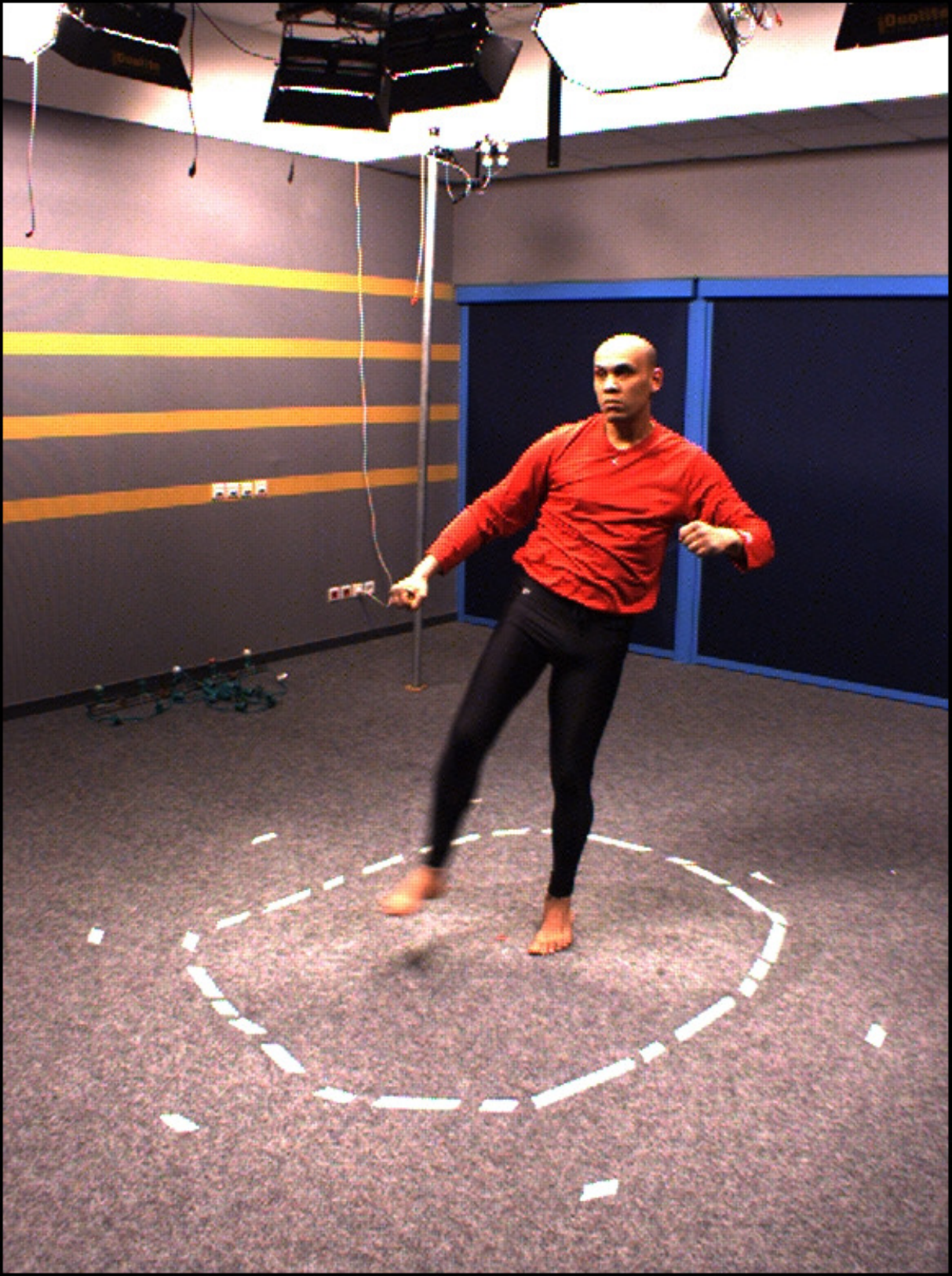}&
\includegraphics[width=0.12\textwidth]{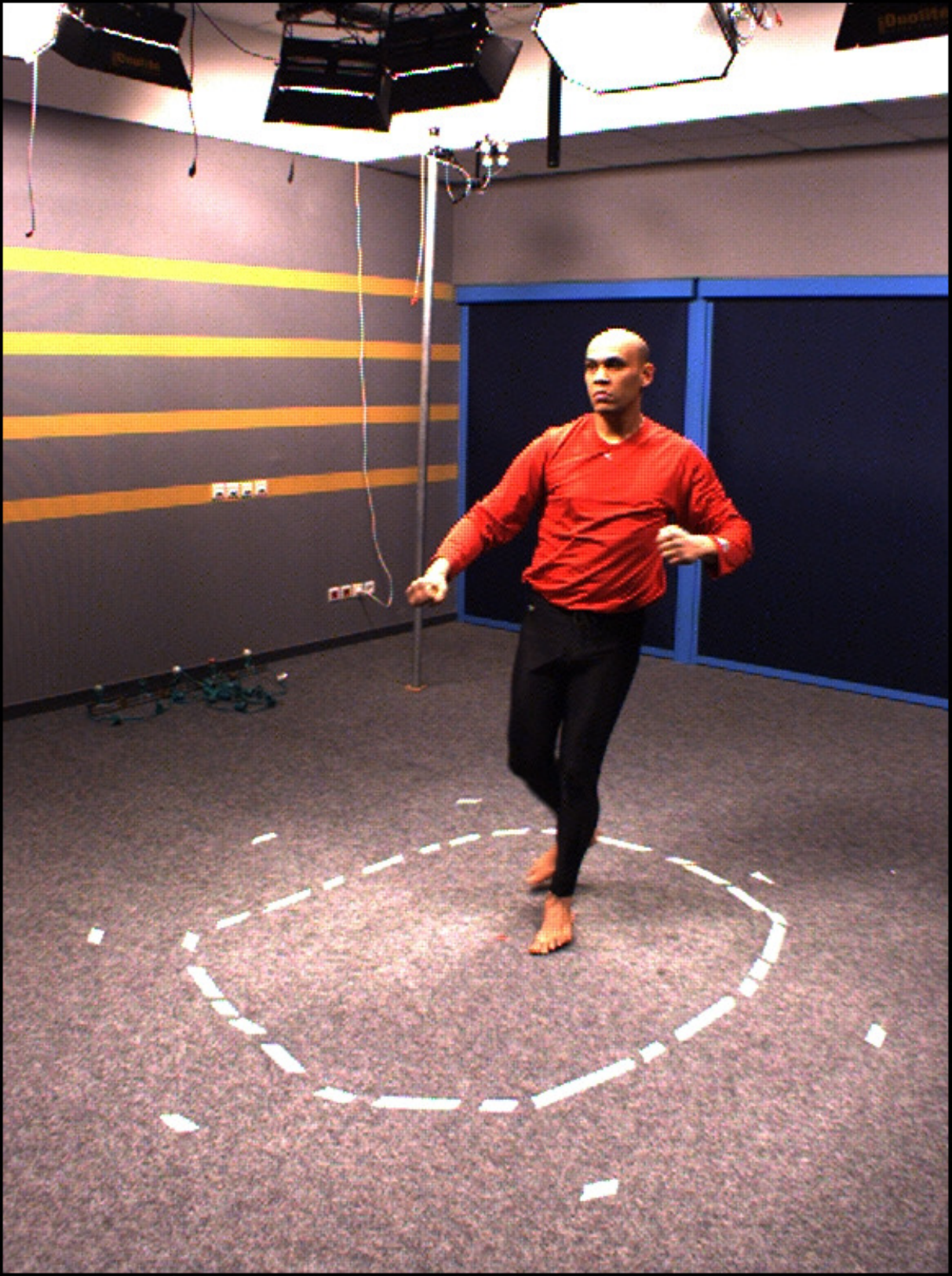}
\\
\includegraphics[width=0.12\textwidth]{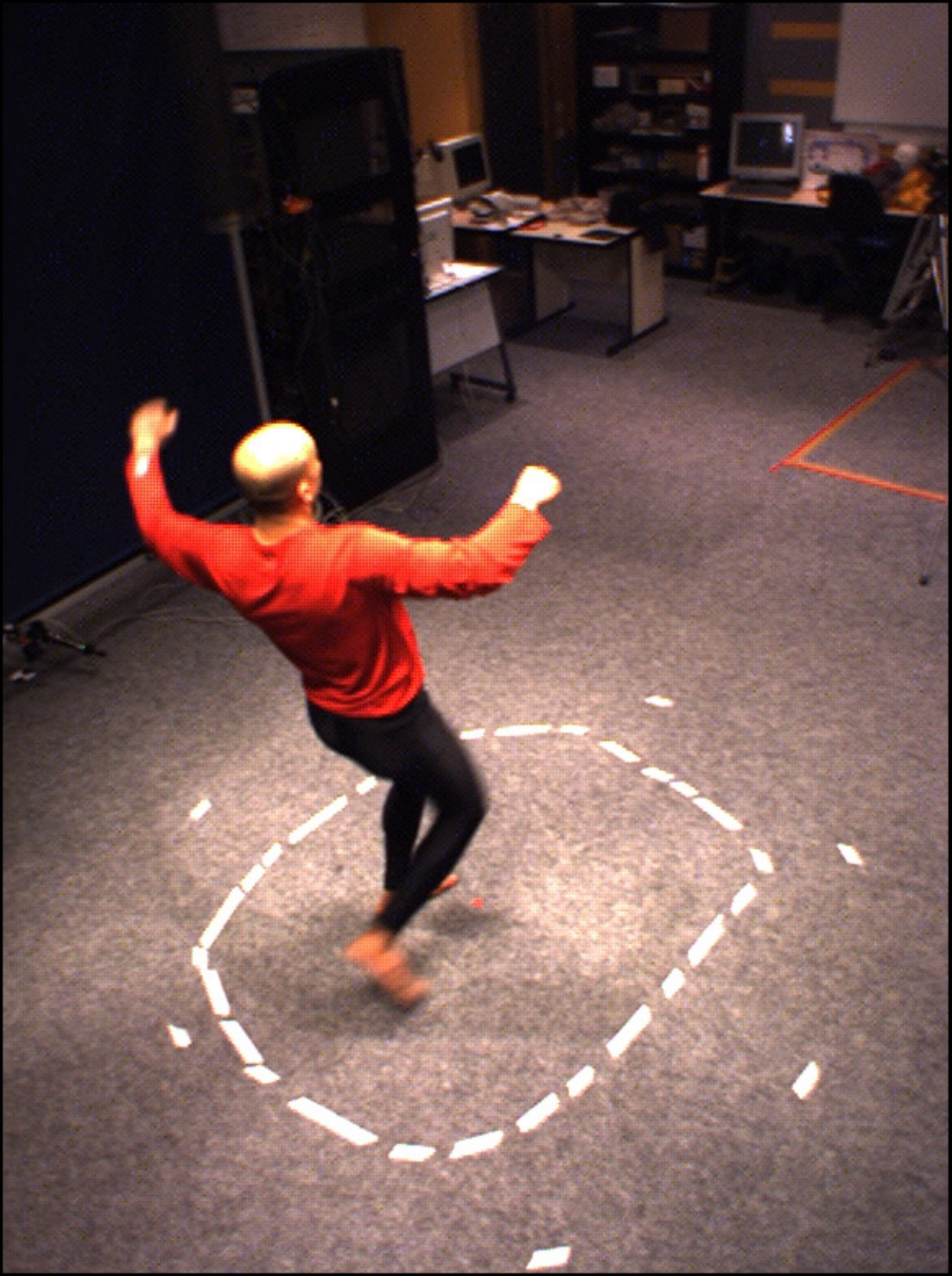} &
\includegraphics[width=0.12\textwidth]{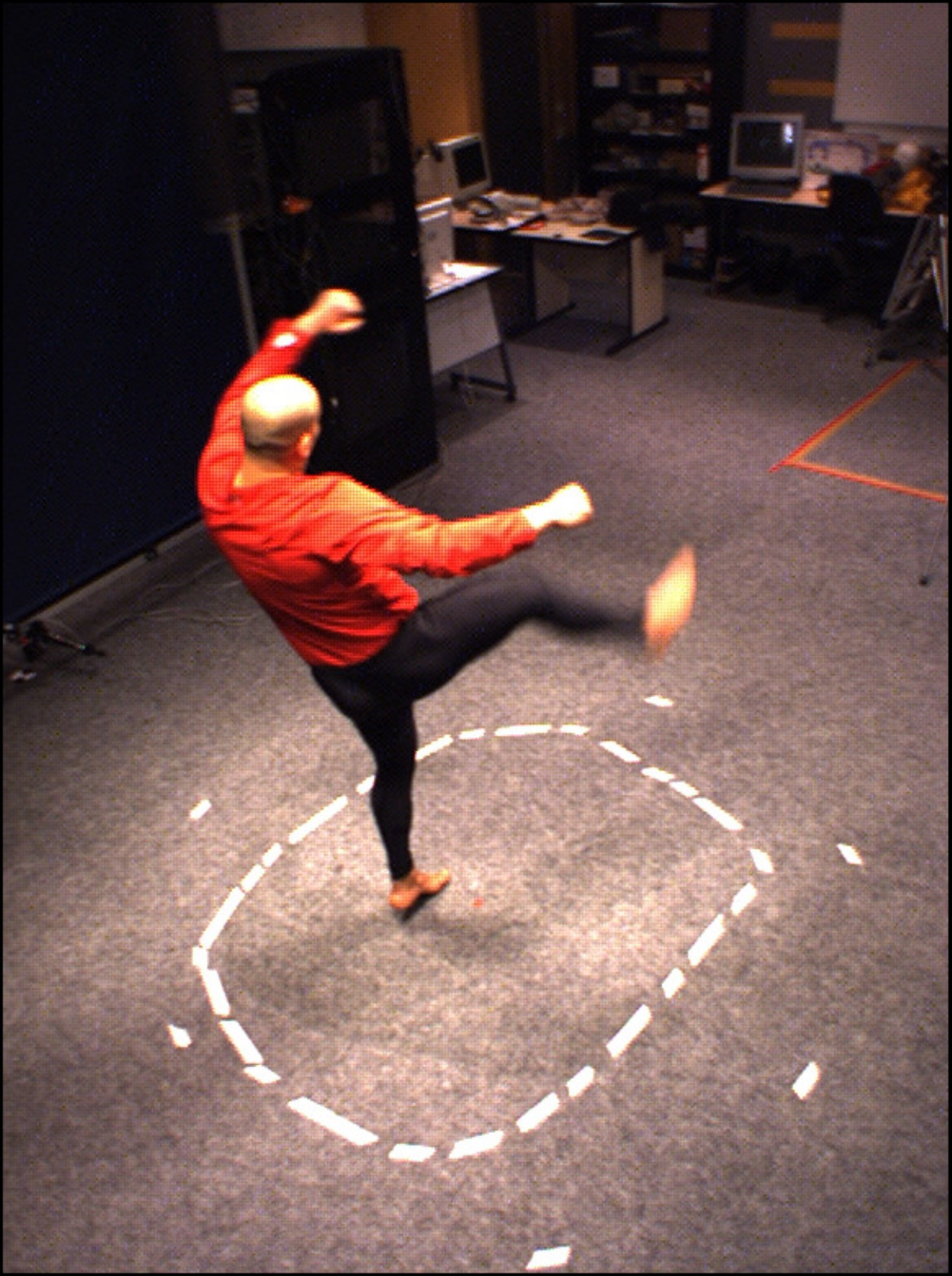} &
\includegraphics[width=0.12\textwidth]{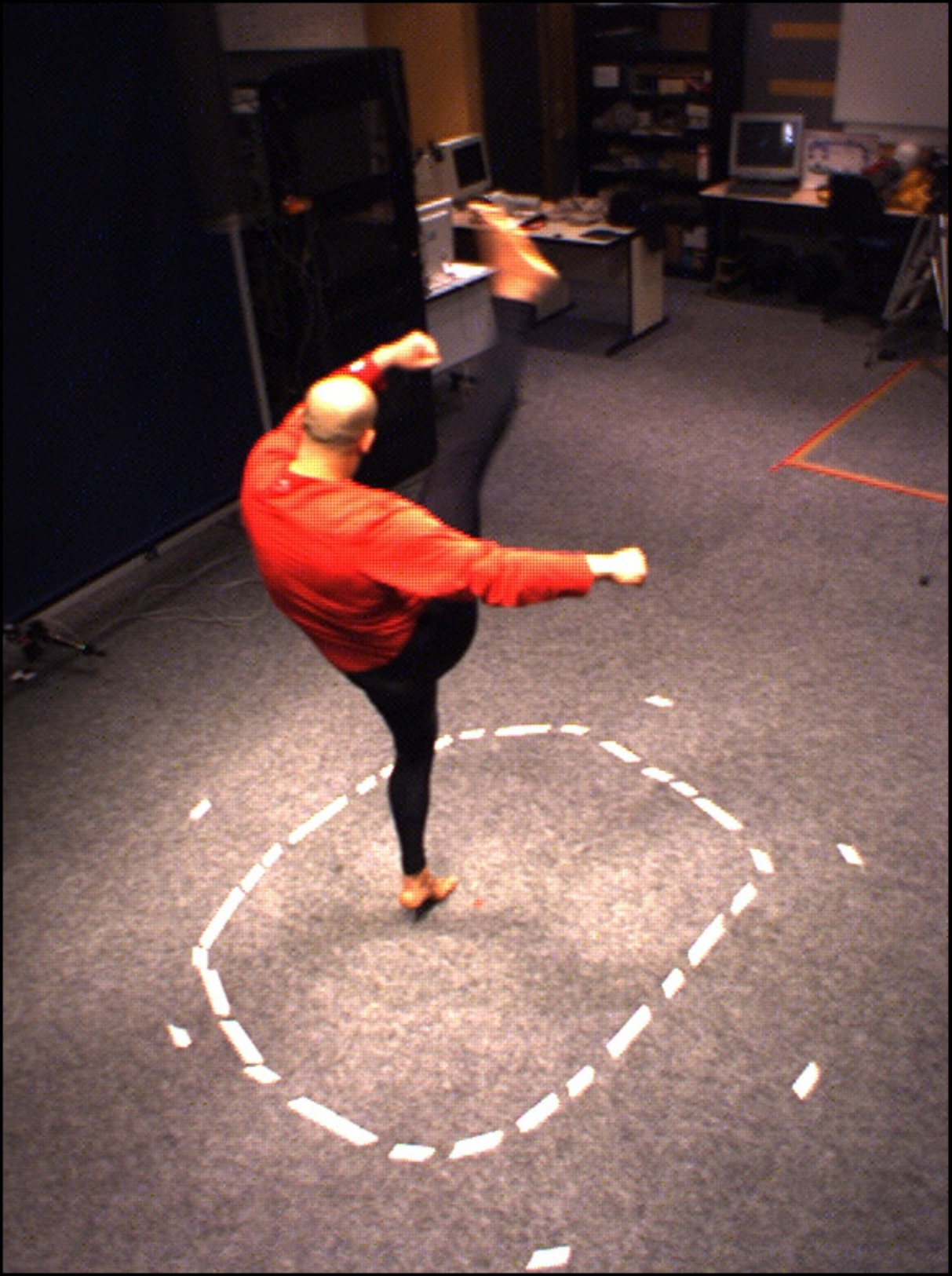} &
\includegraphics[width=0.12\textwidth]{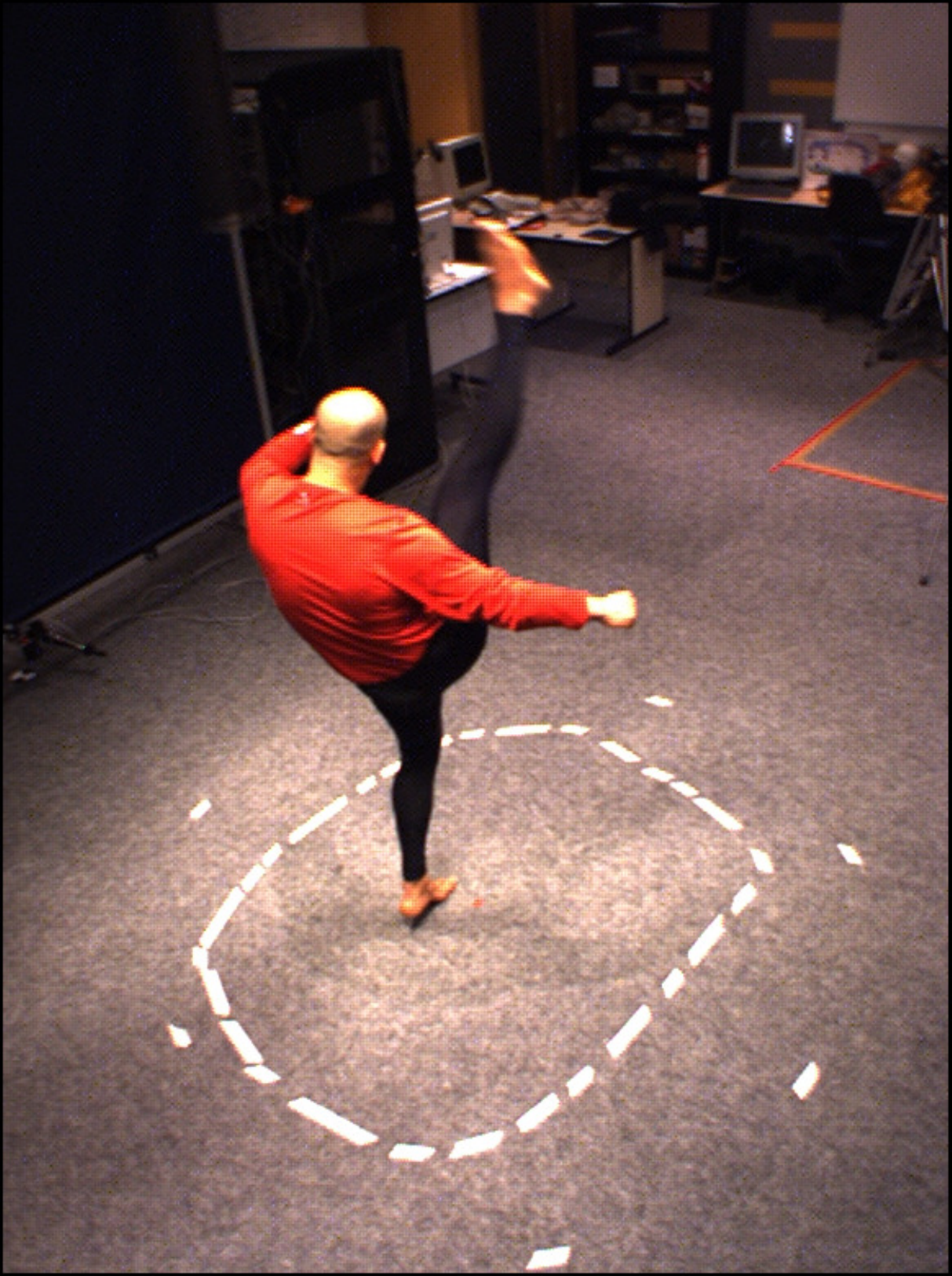} &
\includegraphics[width=0.12\textwidth]{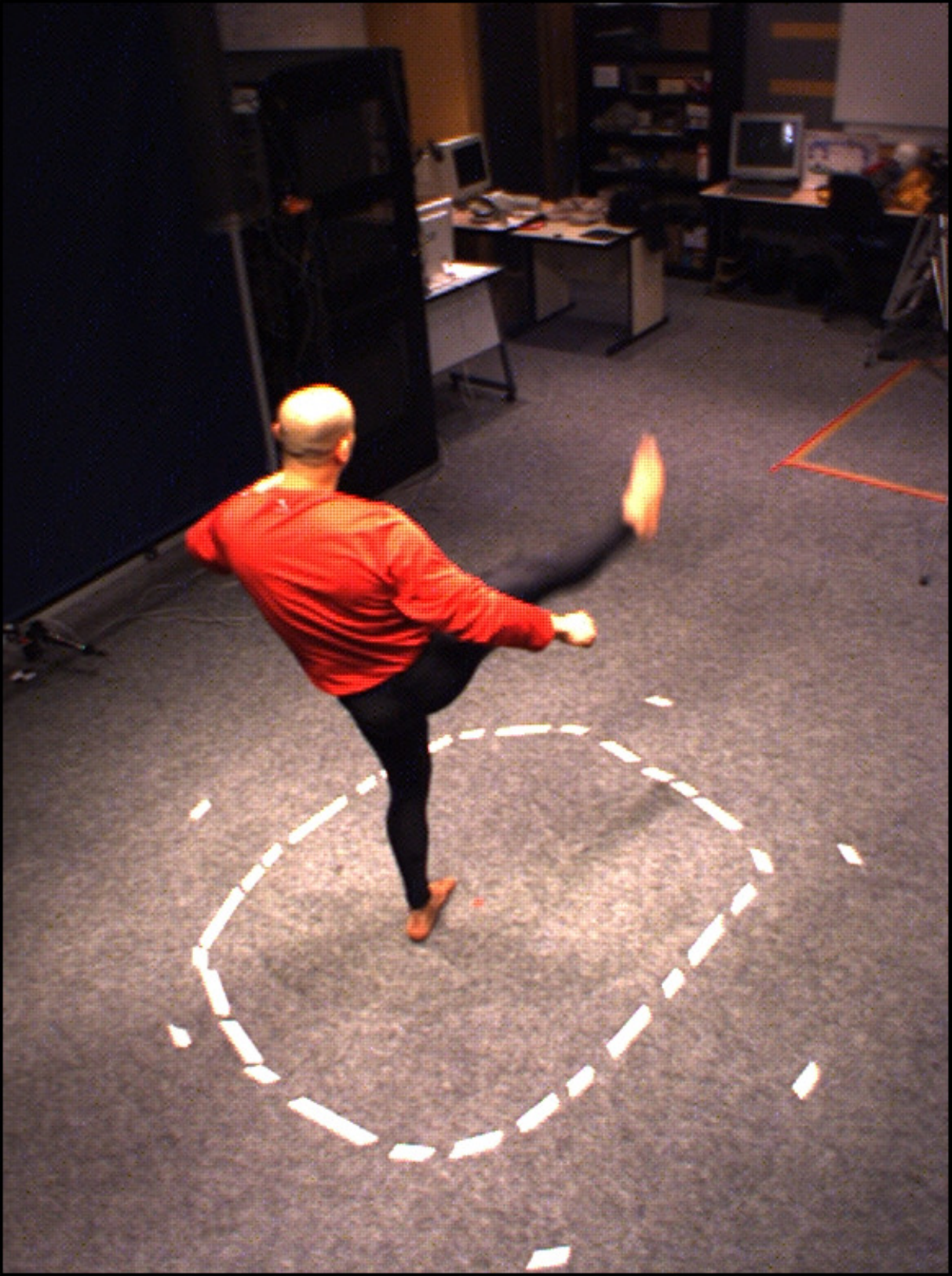}&
\includegraphics[width=0.12\textwidth]{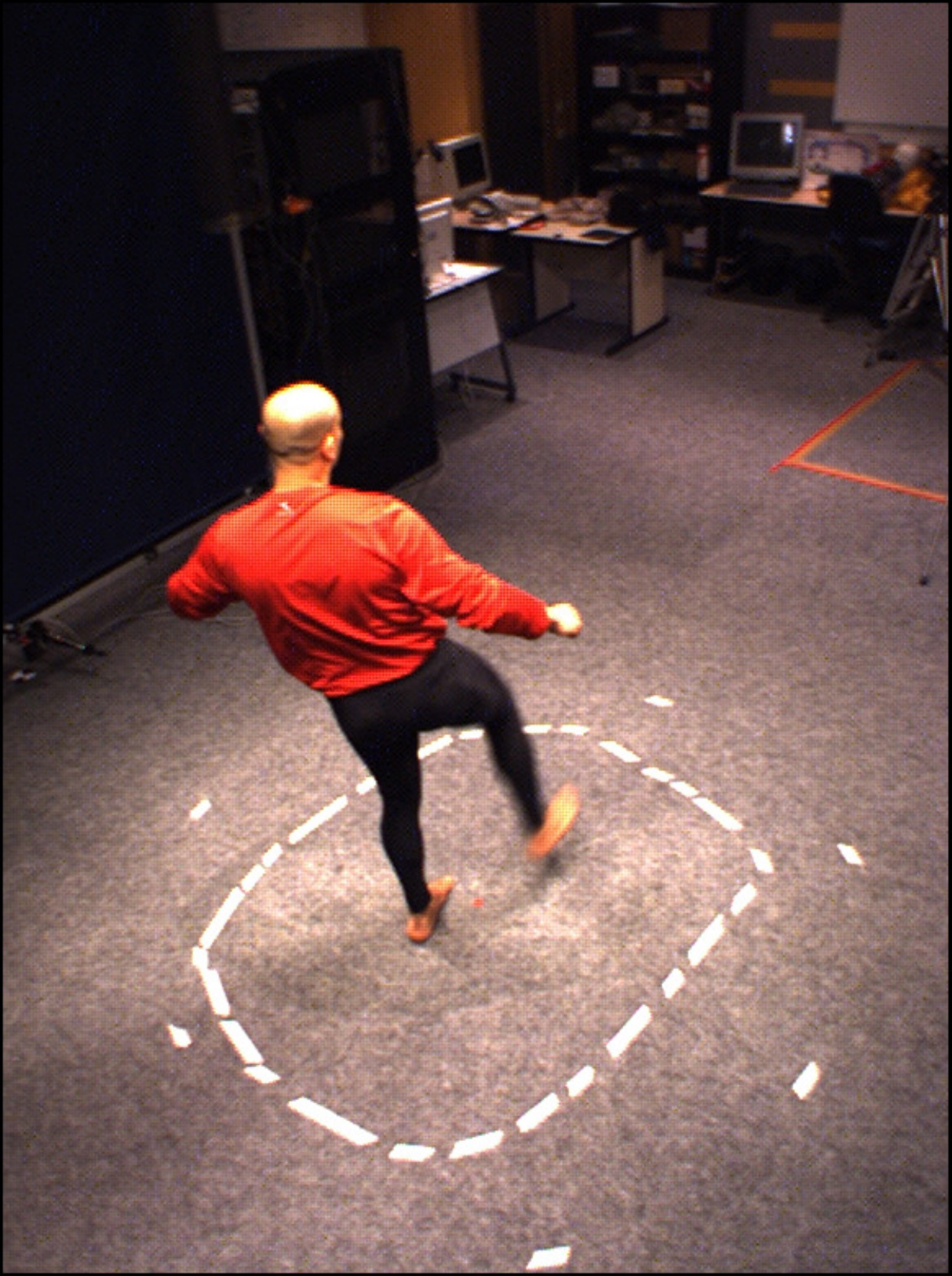}&
\includegraphics[width=0.12\textwidth]{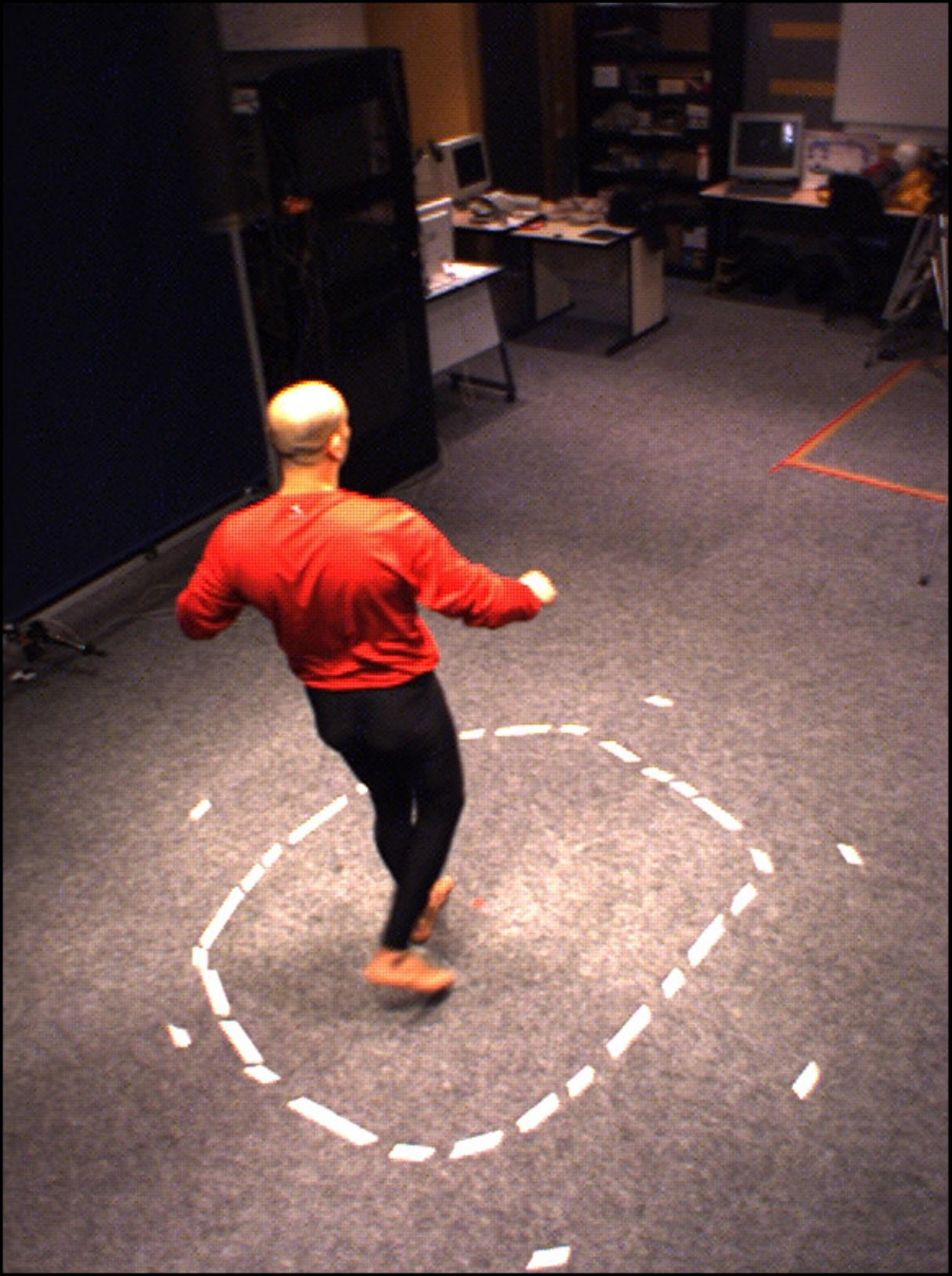}
\\
% first and fourth silhouettes
\includegraphics[width=0.12\textwidth]{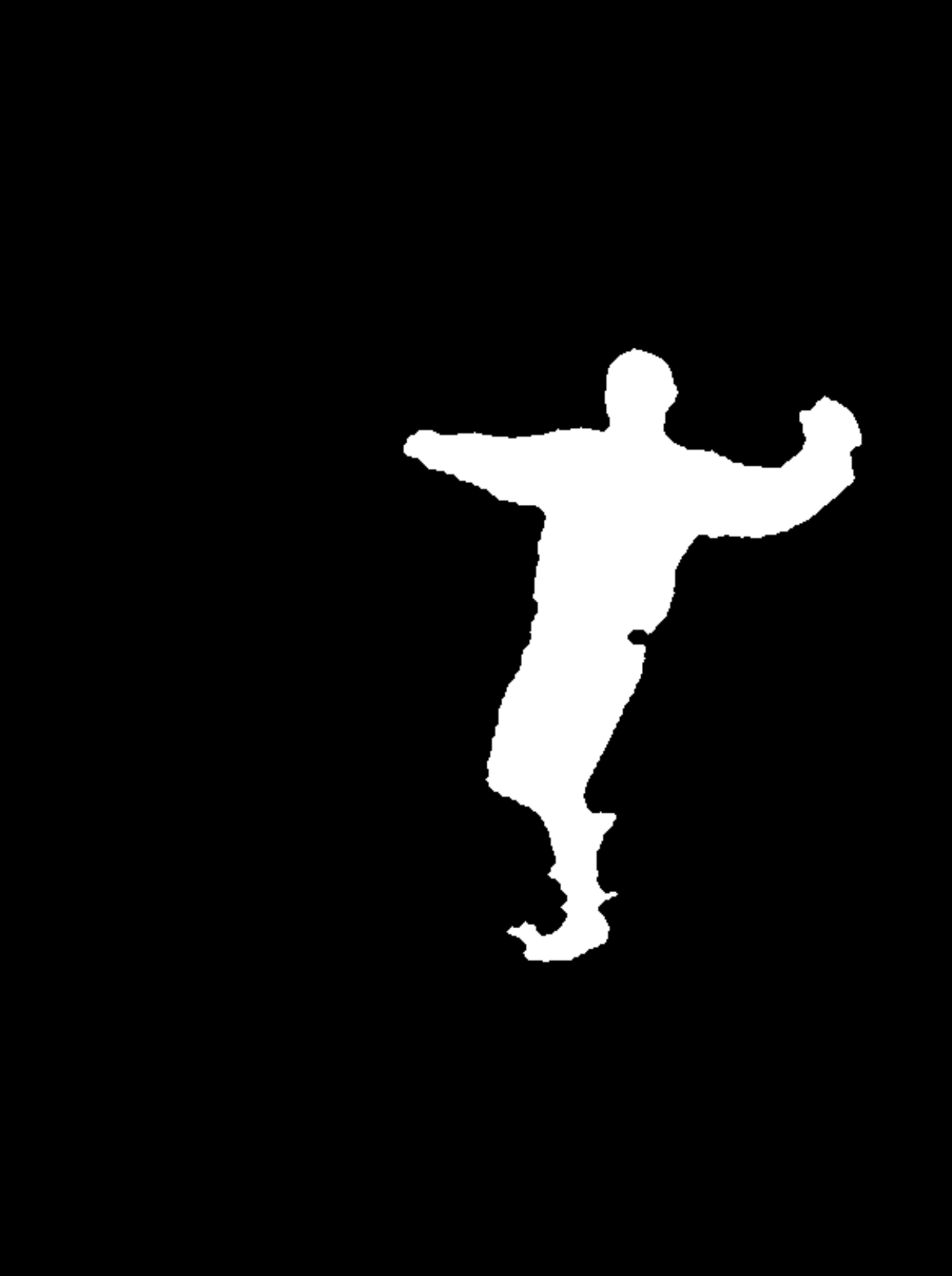} &
\includegraphics[width=0.12\textwidth]{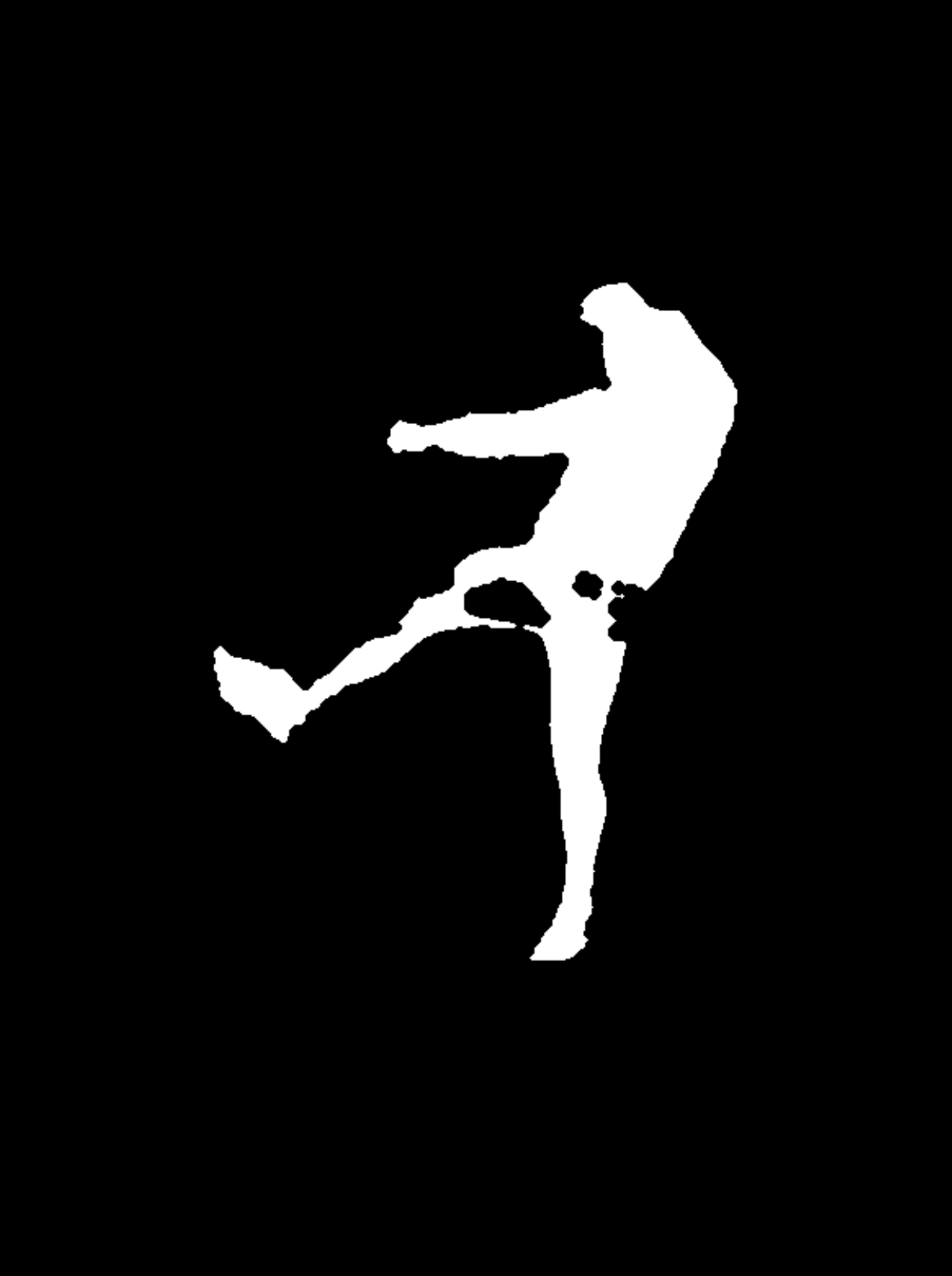} &
\includegraphics[width=0.12\textwidth]{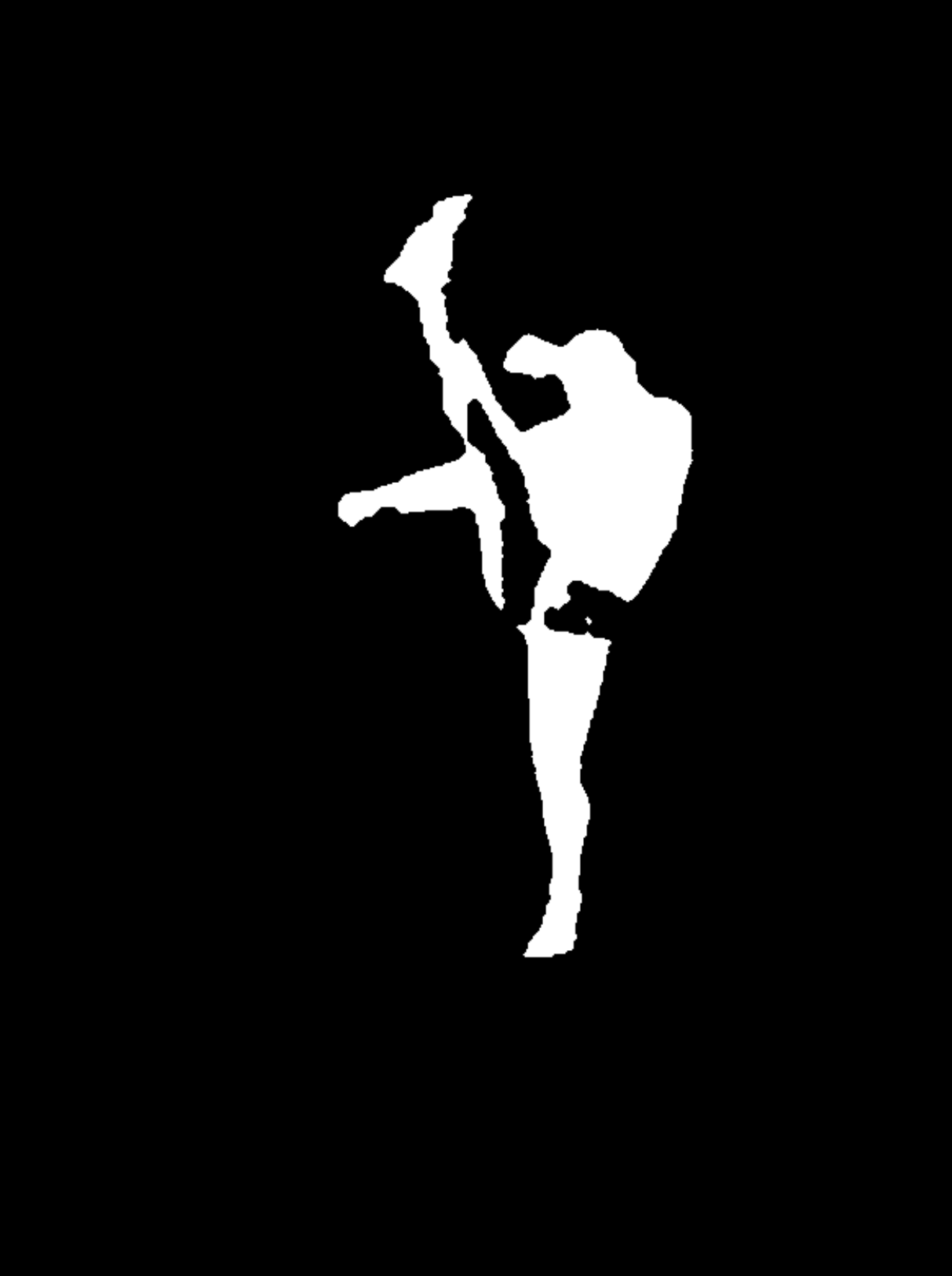} &
\includegraphics[width=0.12\textwidth]{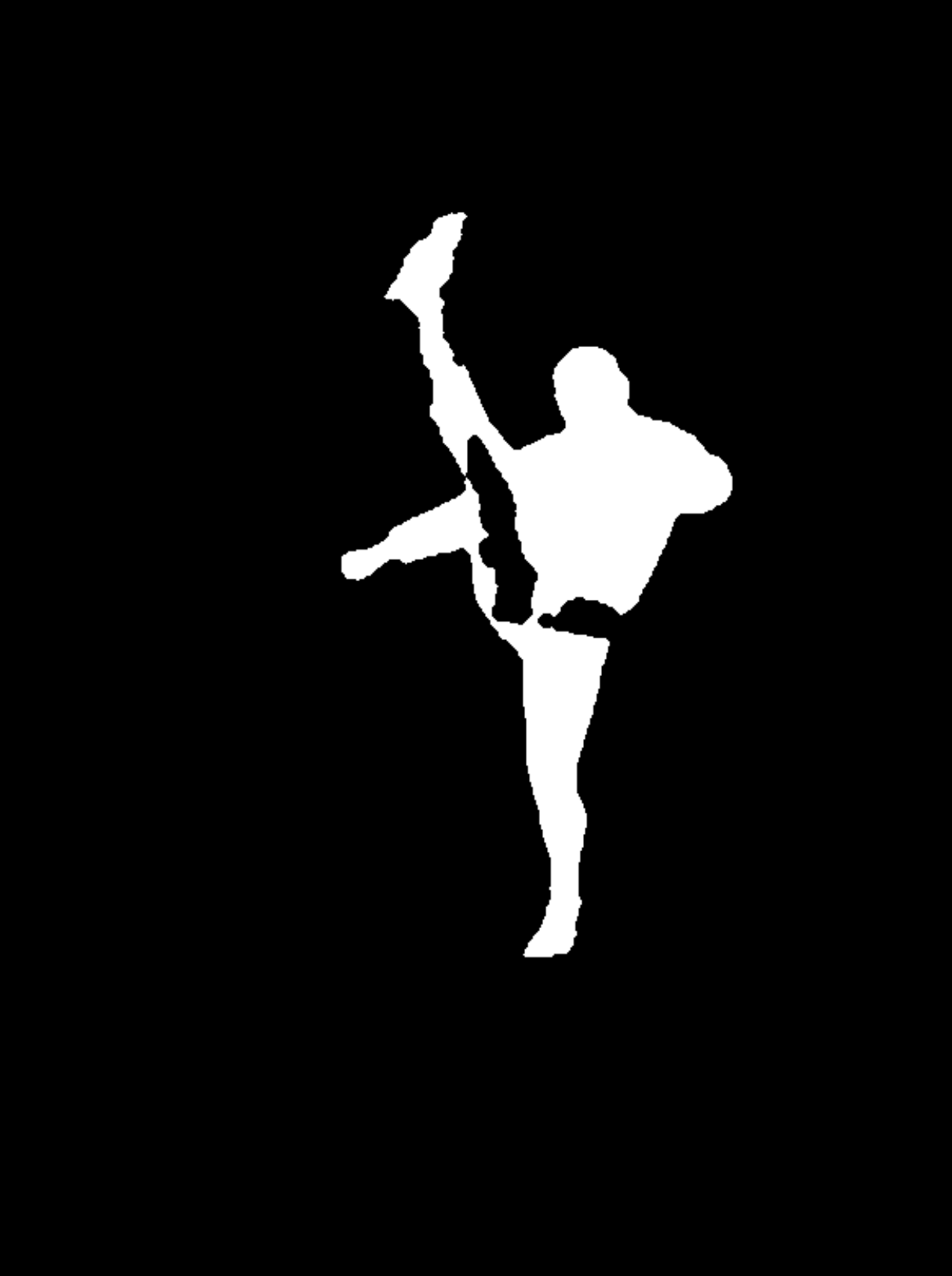} &
\includegraphics[width=0.12\textwidth]{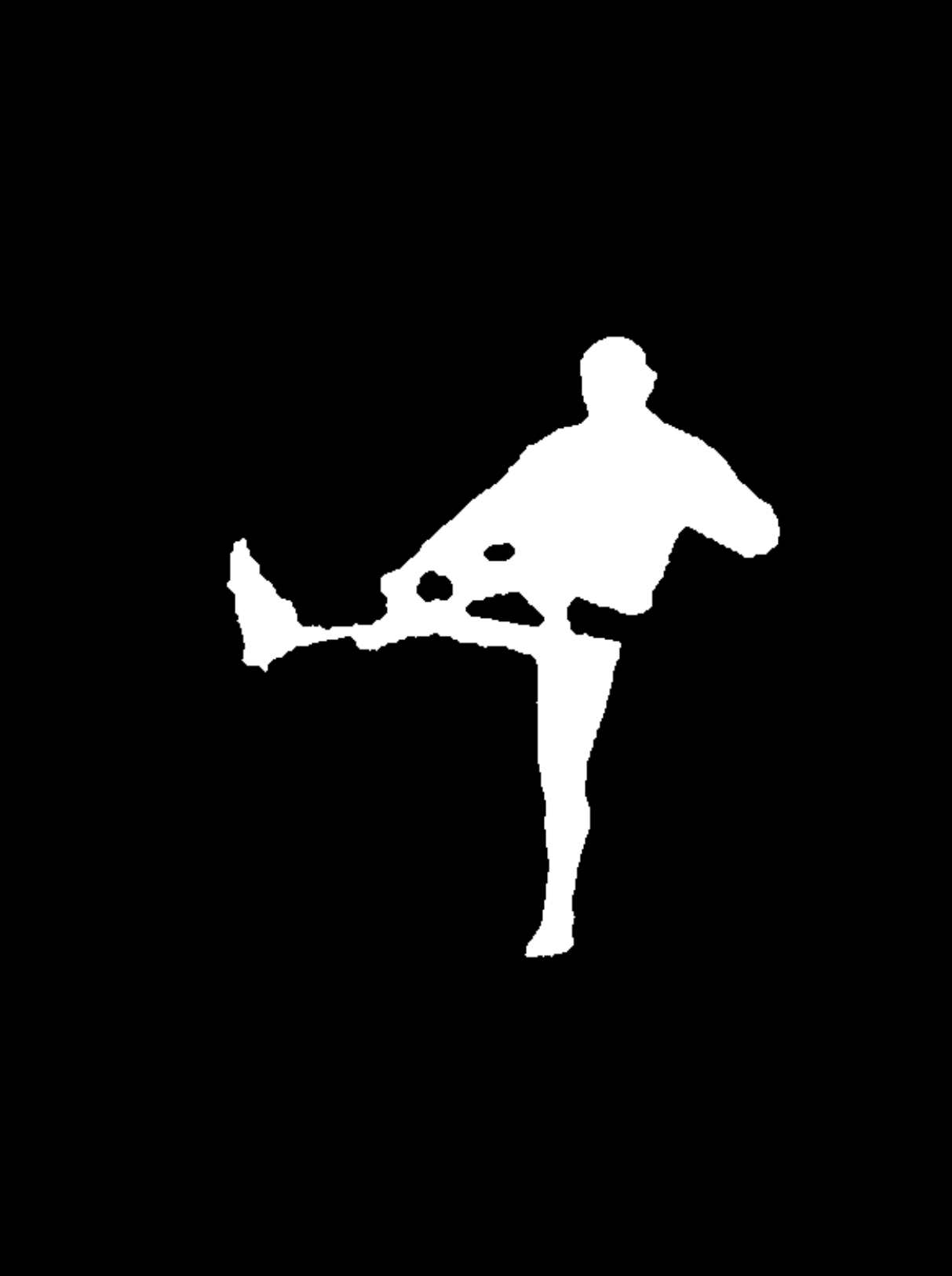}&
\includegraphics[width=0.12\textwidth]{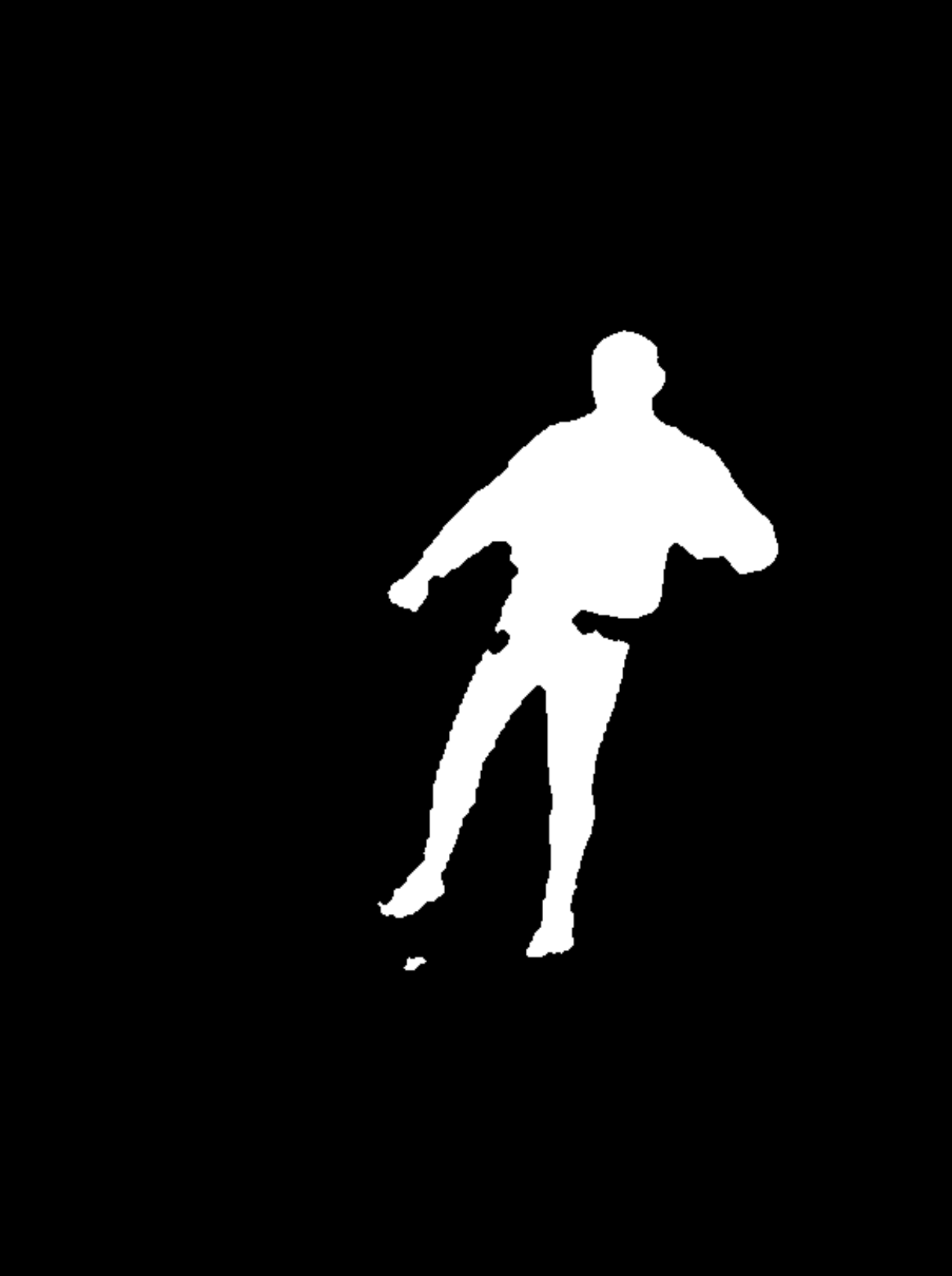}&
\includegraphics[width=0.12\textwidth]{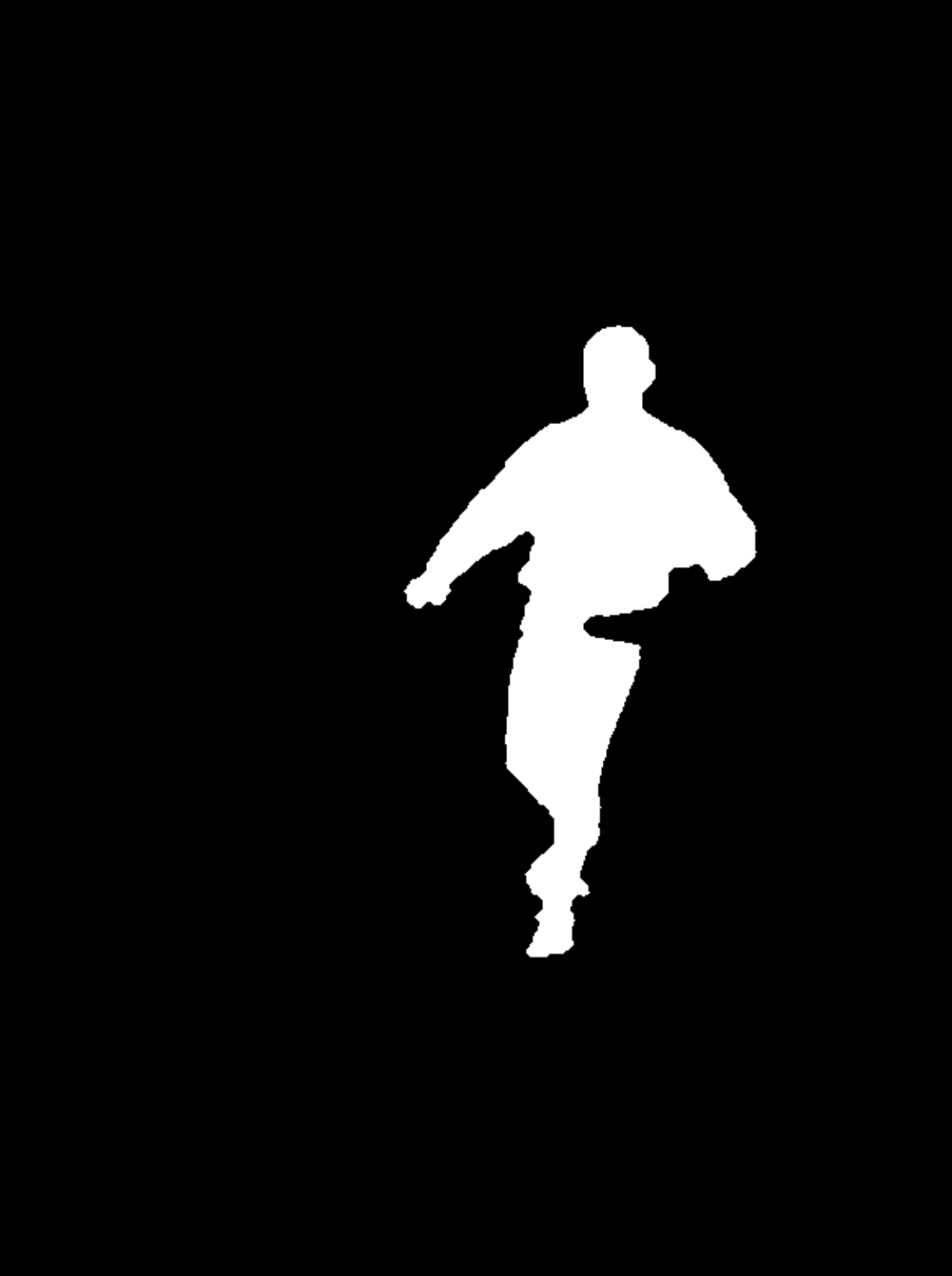}
\\
\includegraphics[width=0.12\textwidth]{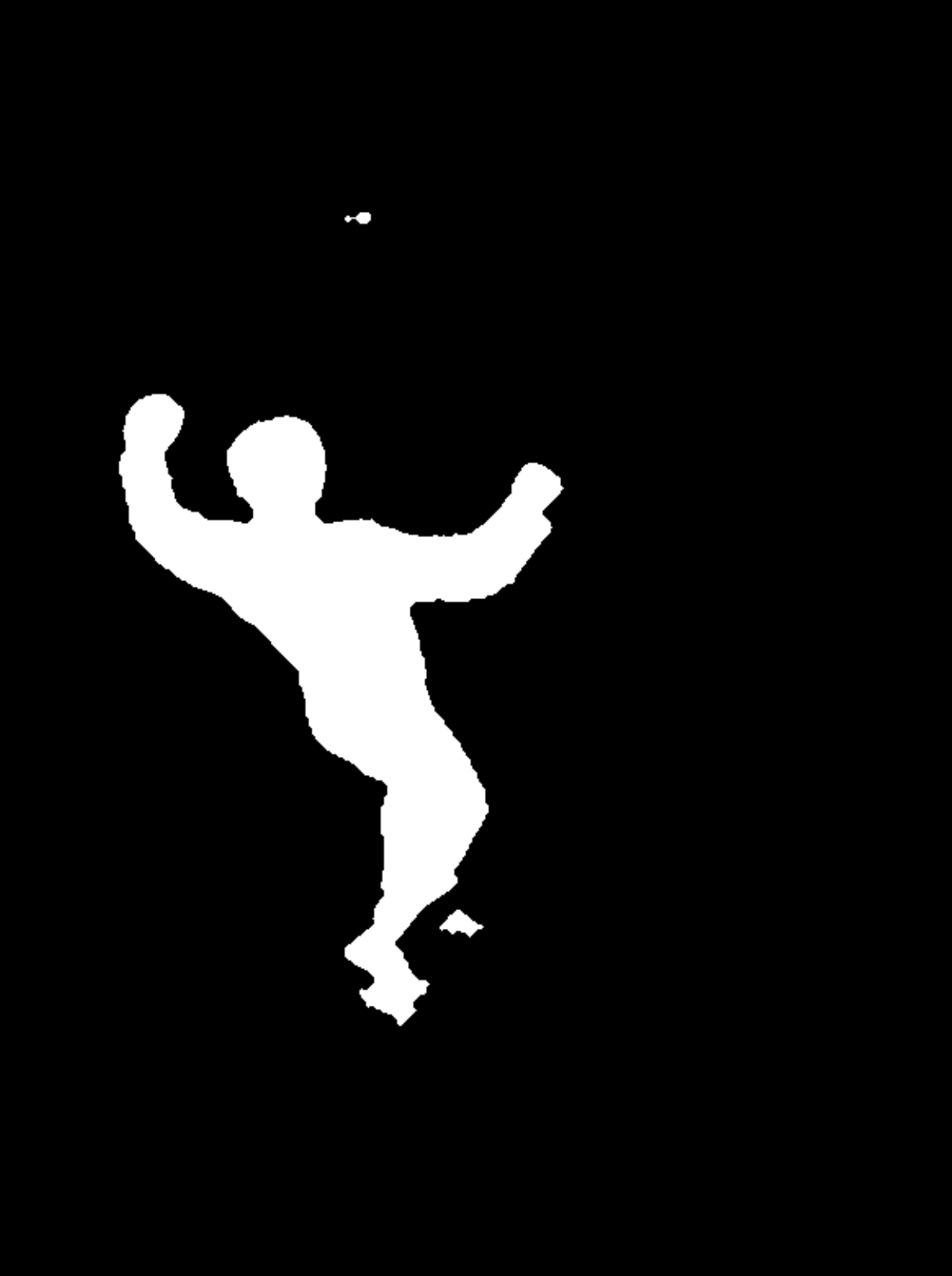} &
\includegraphics[width=0.12\textwidth]{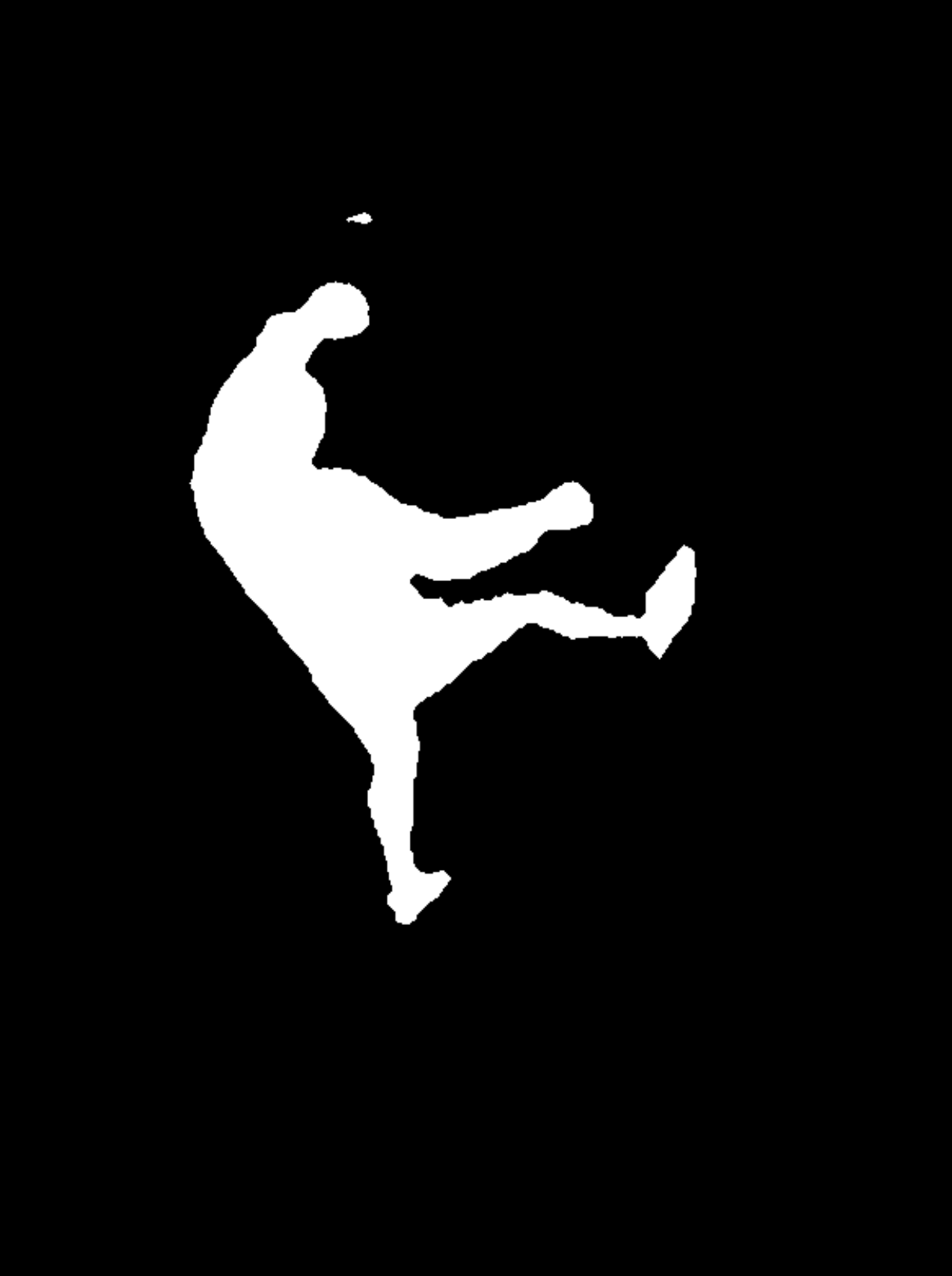} &
\includegraphics[width=0.12\textwidth]{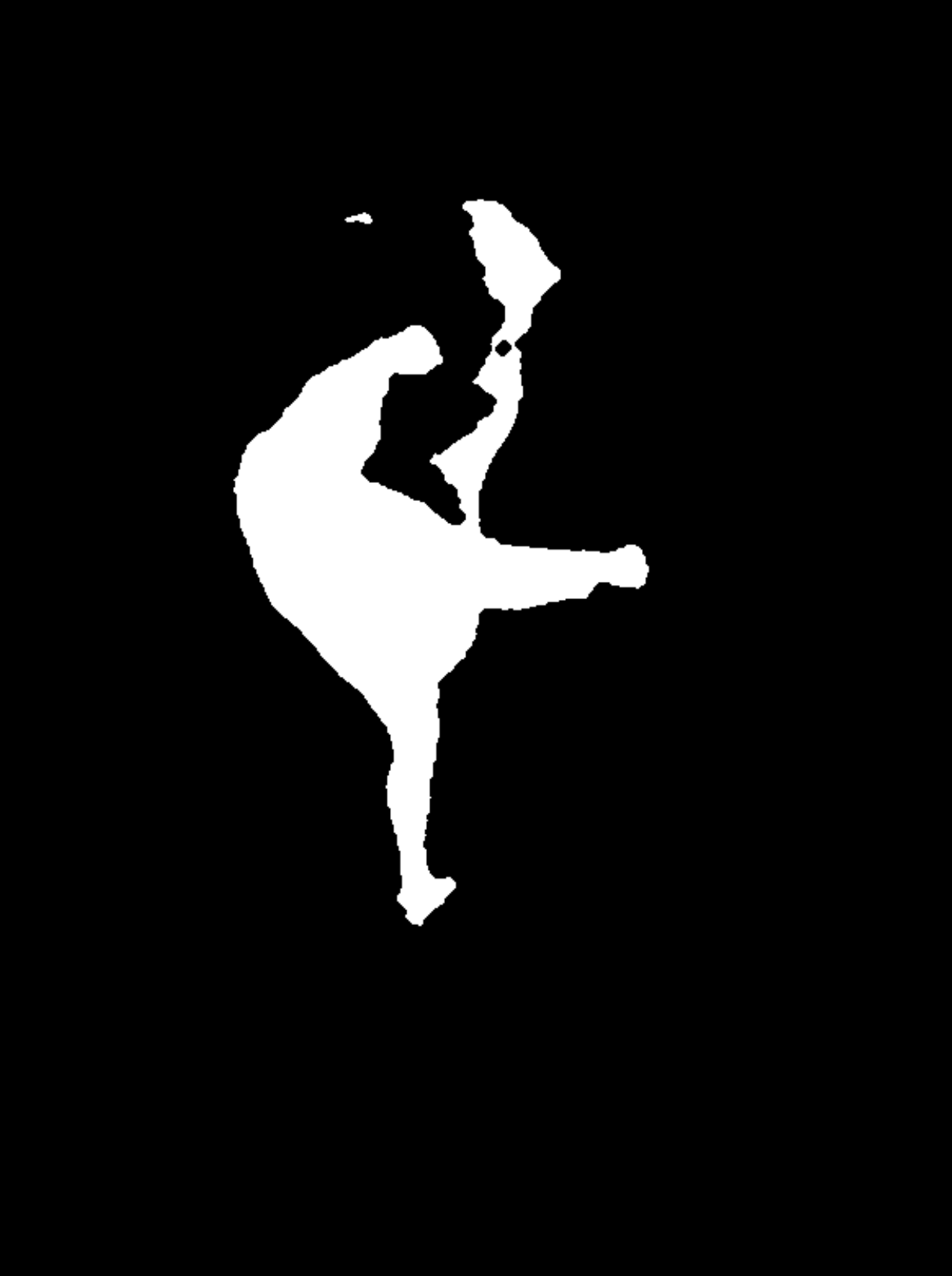} &
\includegraphics[width=0.12\textwidth]{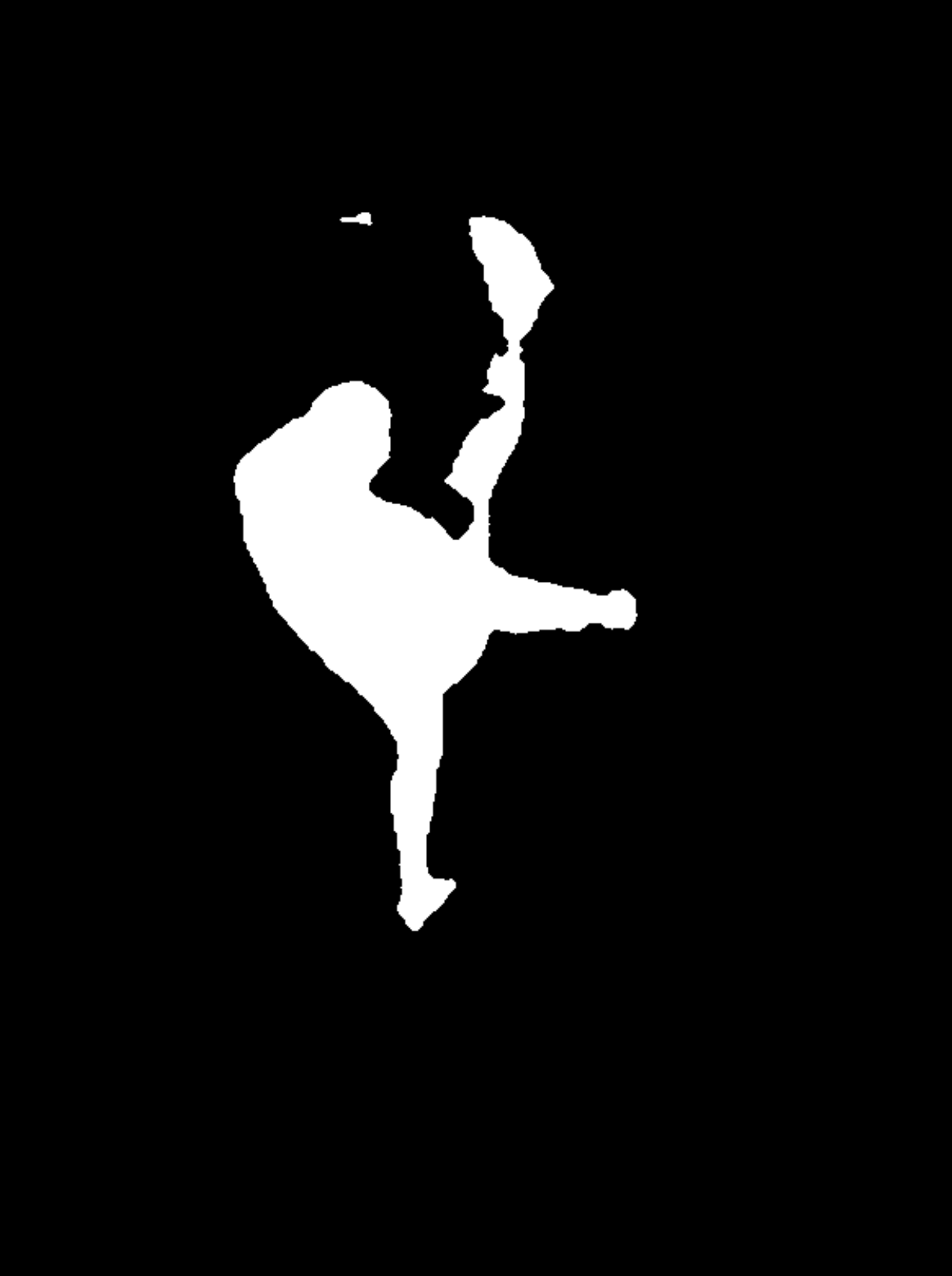} &
\includegraphics[width=0.12\textwidth]{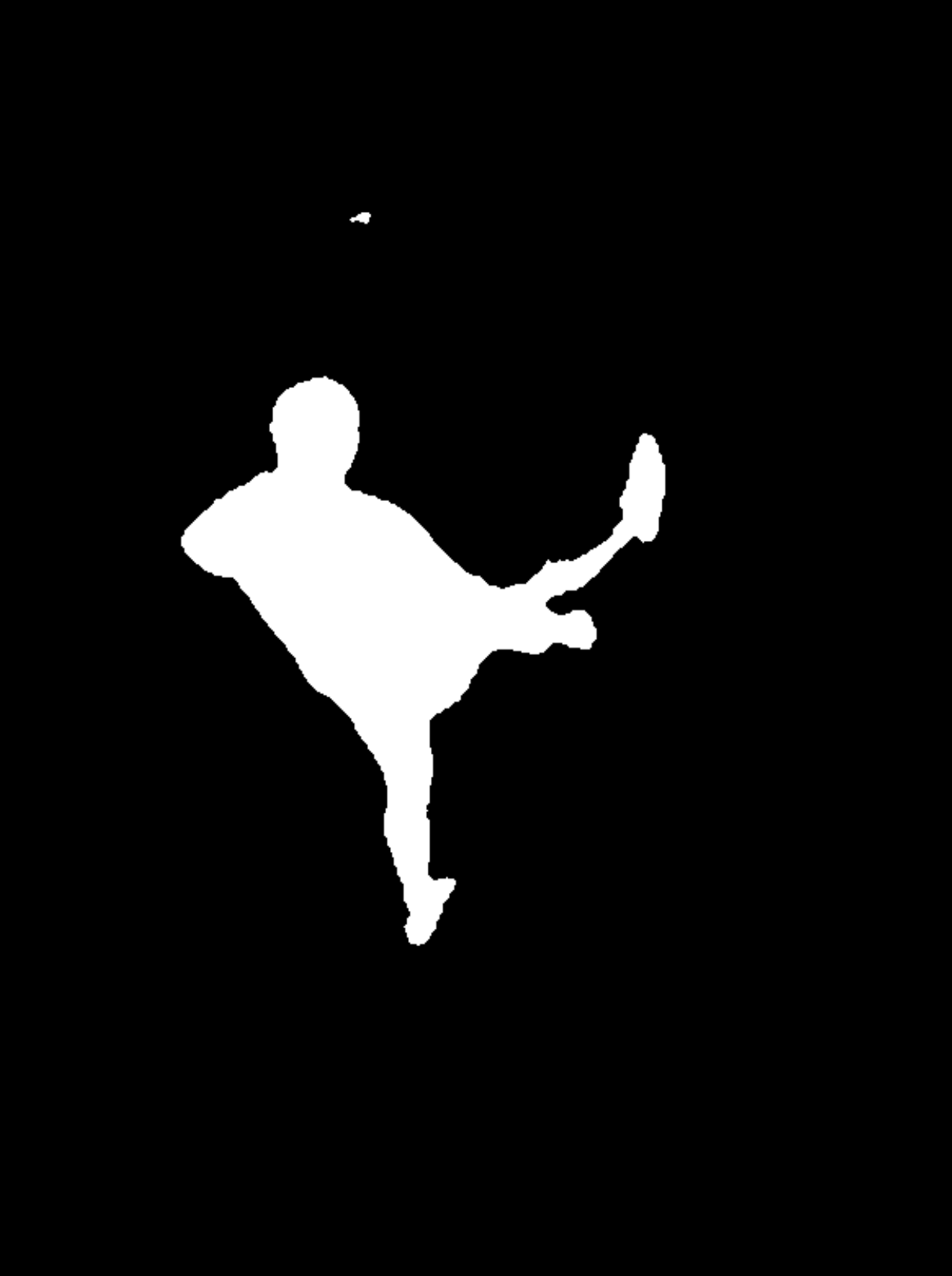}&
\includegraphics[width=0.12\textwidth]{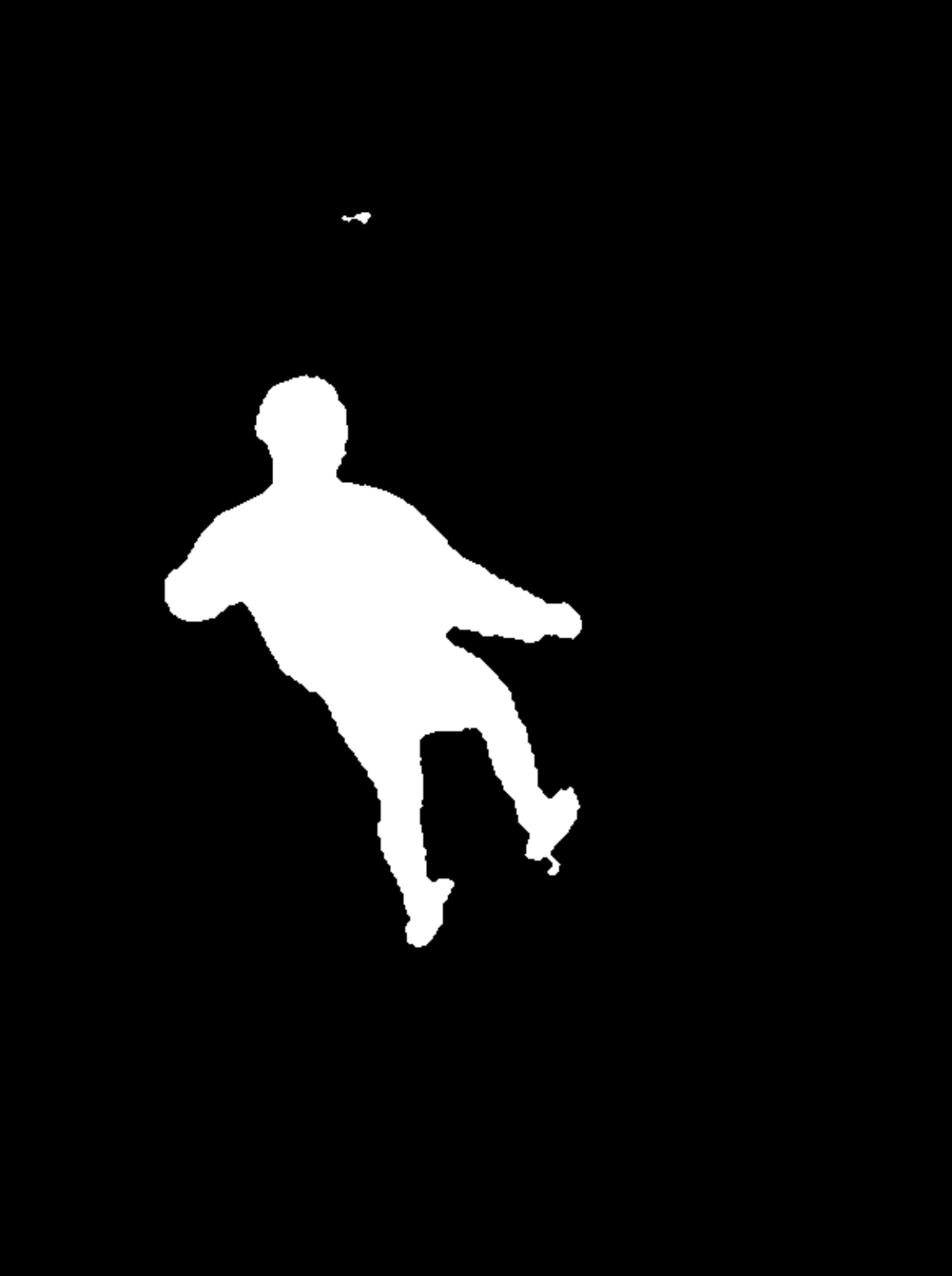}&
\includegraphics[width=0.12\textwidth]{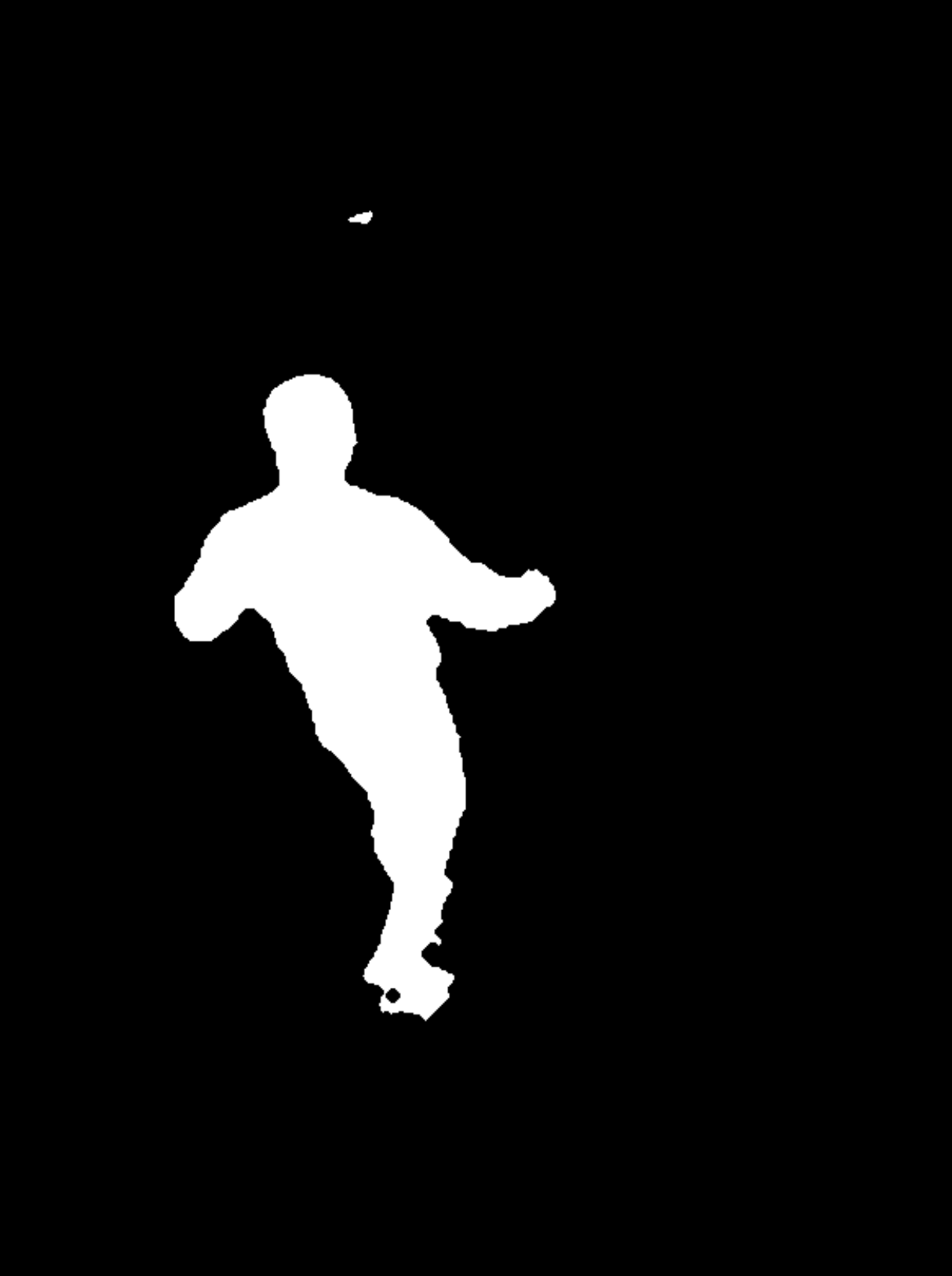}
\\
\end{tabular}
\caption{The taekwendo sequence. The images and the associated
  silhouettes are from the first and fourth cameras. Imperfect
  silhouettes generate outlier data which are properly handled by our method.} 
\label{fig:Ben-taekwendo-rawdata}
\end{center}
\end{figure*}

We simulated a running sequence that involves both
kinematic and free-motion parameters.  
The graphs of Figure~\ref{fig:graphs} illustrate the average error 
between the simulated angle values and their estimated values
(measured in radians) for the
kinematic parameters but not for the free-motion parameters. 
The first
graph compares the ground-truth (simulated) joint trajectories of the left
knee and of the right elbow (dashed curve) with the trajectories estimated with
our method (solid curve) over 100 frames. The second graph
illustrates the behavior of the method in the presence of
silhouette noise. 
The
results of using both 3-D points and normals (dashed curve) are plotted
against the results obtained using 3-D points and the algebraic
distance (solid line). 
The relative large error corresponds to the fact that the shape model
used by the animation package is not the same as our shape
model. Hence, there is a systematic offset between the ground-truth
kinematic parameters and the estimated parameters.

The third graph shows the average angle error as a
function of the number of frames per second. 
The last graph shows the influence of
the number of observations, where the latter varies from 50 to 550. The
average angle error drops as the number of observations increases and 
our method (dashed) performs better than 
using 3-D points alone (solid). 
From all these experiments
one may conclude that tracking is improved when
both points and normals are used instead of just points.

%The geometry of the cameras together with an example of
%recovered 3-D data are shown on Figure~\ref{fig:illustration}.
%\begin{figure}[ht]
%\begin{center}
%\includegraphics[width=0.60\textwidth]{modele-eps-converted-to.pdf}
%\caption{The six calibrated and finely synchronized cameras overlook a
%  moving character and a reconstruction
%method estimates 3-D points (connected to form a mesh for the
%purpose of the display) as well as 3-D unit vectors (shown as a needle
%field) normal to a smooth surface approximating the shape.} 
%\label{fig:illustration}
%\end{center}
%\end{figure}

\begin{figure*}[h!t!b]
\begin{center}
\begin{tabular}{ccccccc}
% 3-D observations seen from first and fourht viewpoints
\frame{\includegraphics[width=0.12\textwidth]{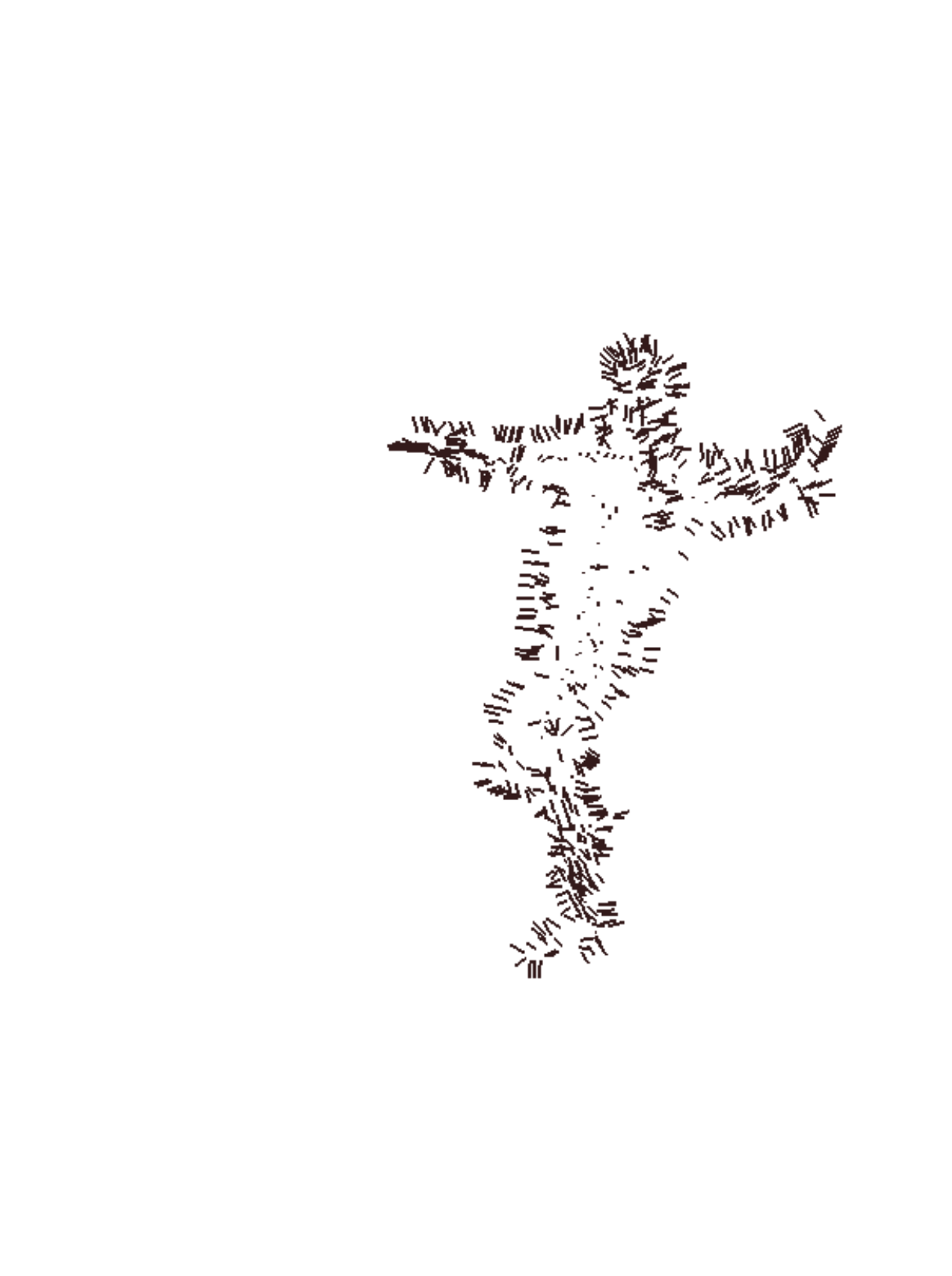}} &
\frame{\includegraphics[width=0.12\textwidth]{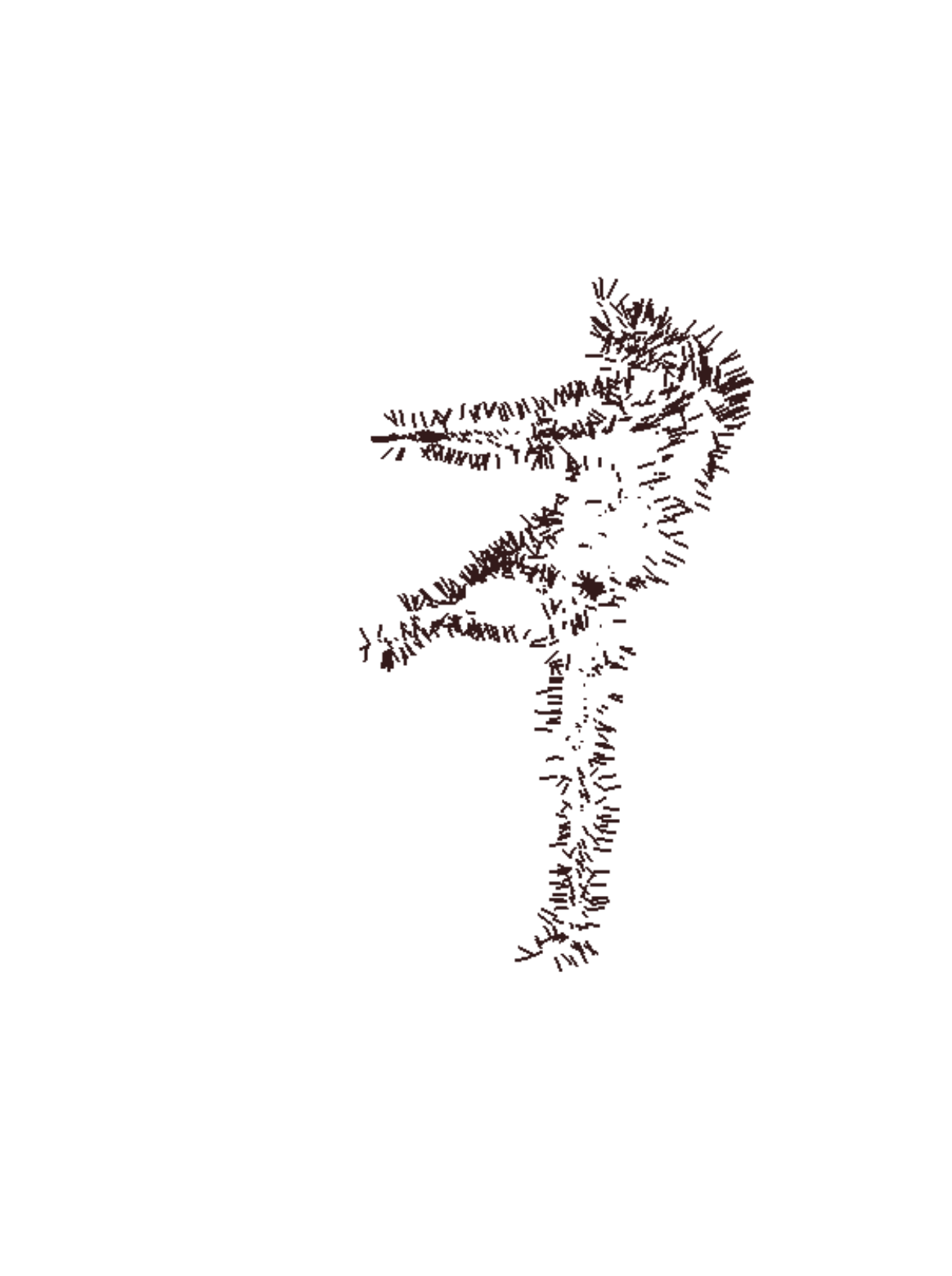}} &
\frame{\includegraphics[width=0.12\textwidth]{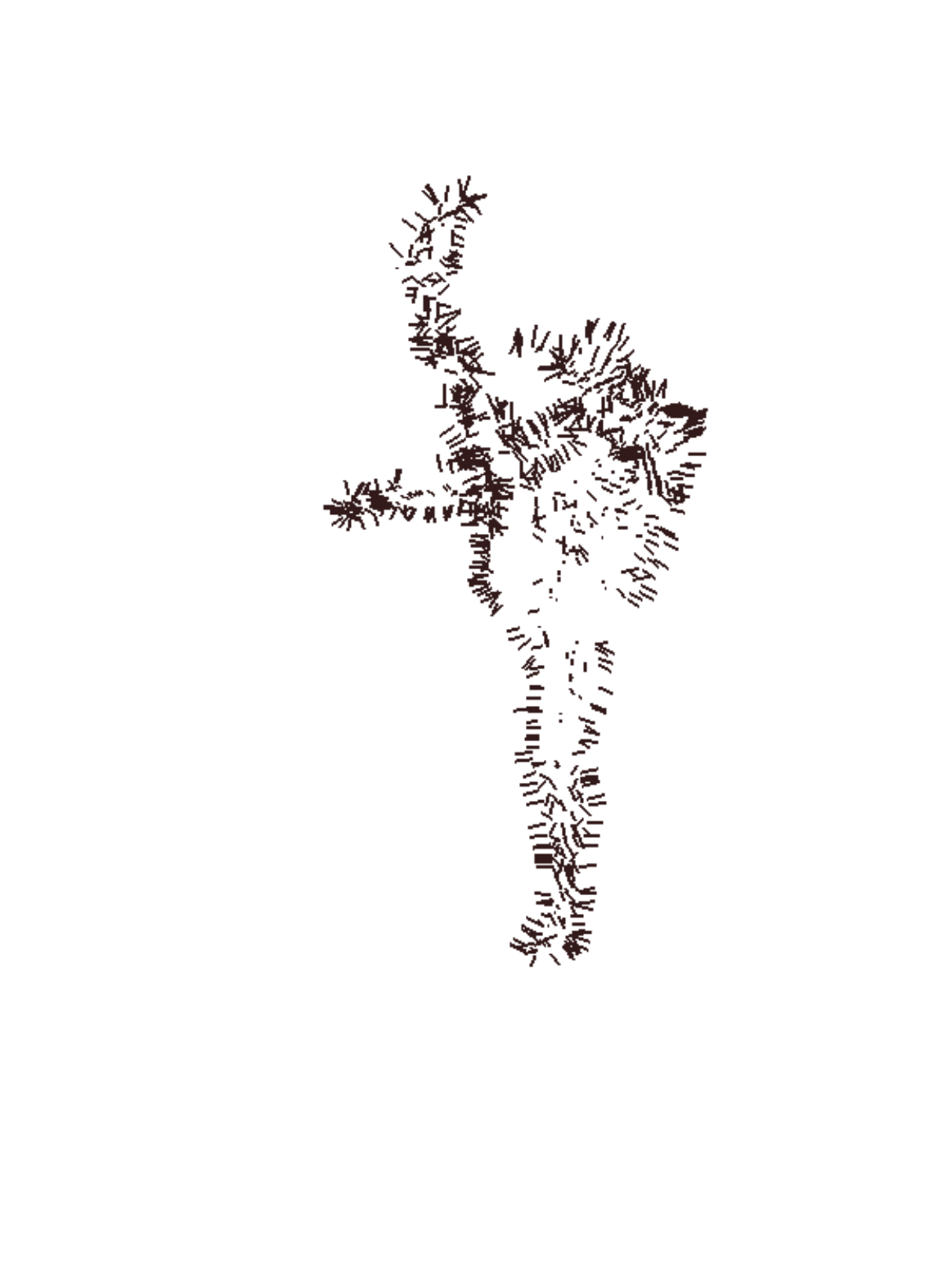}} &
\frame{\includegraphics[width=0.12\textwidth]{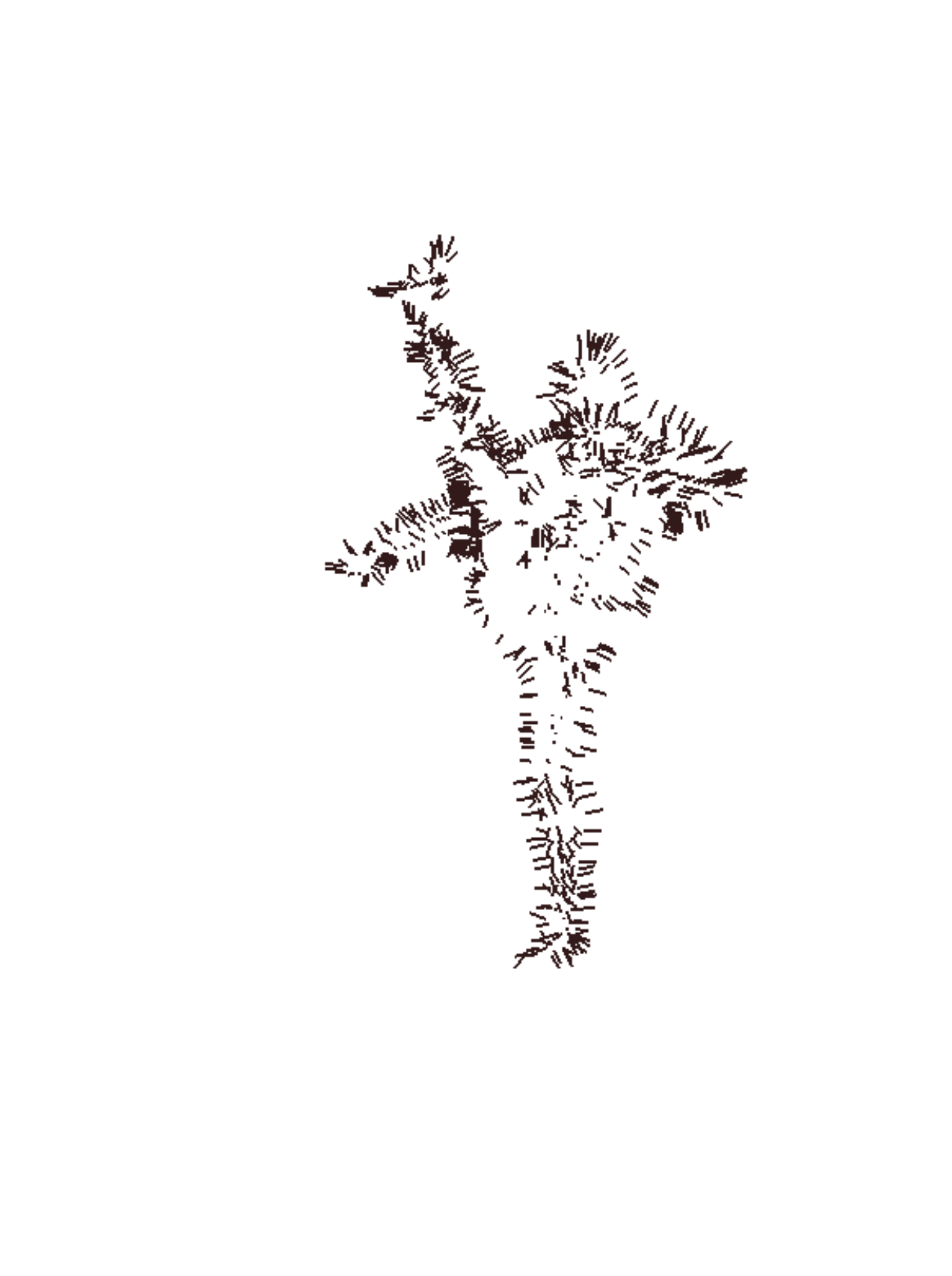}} &
\frame{\includegraphics[width=0.12\textwidth]{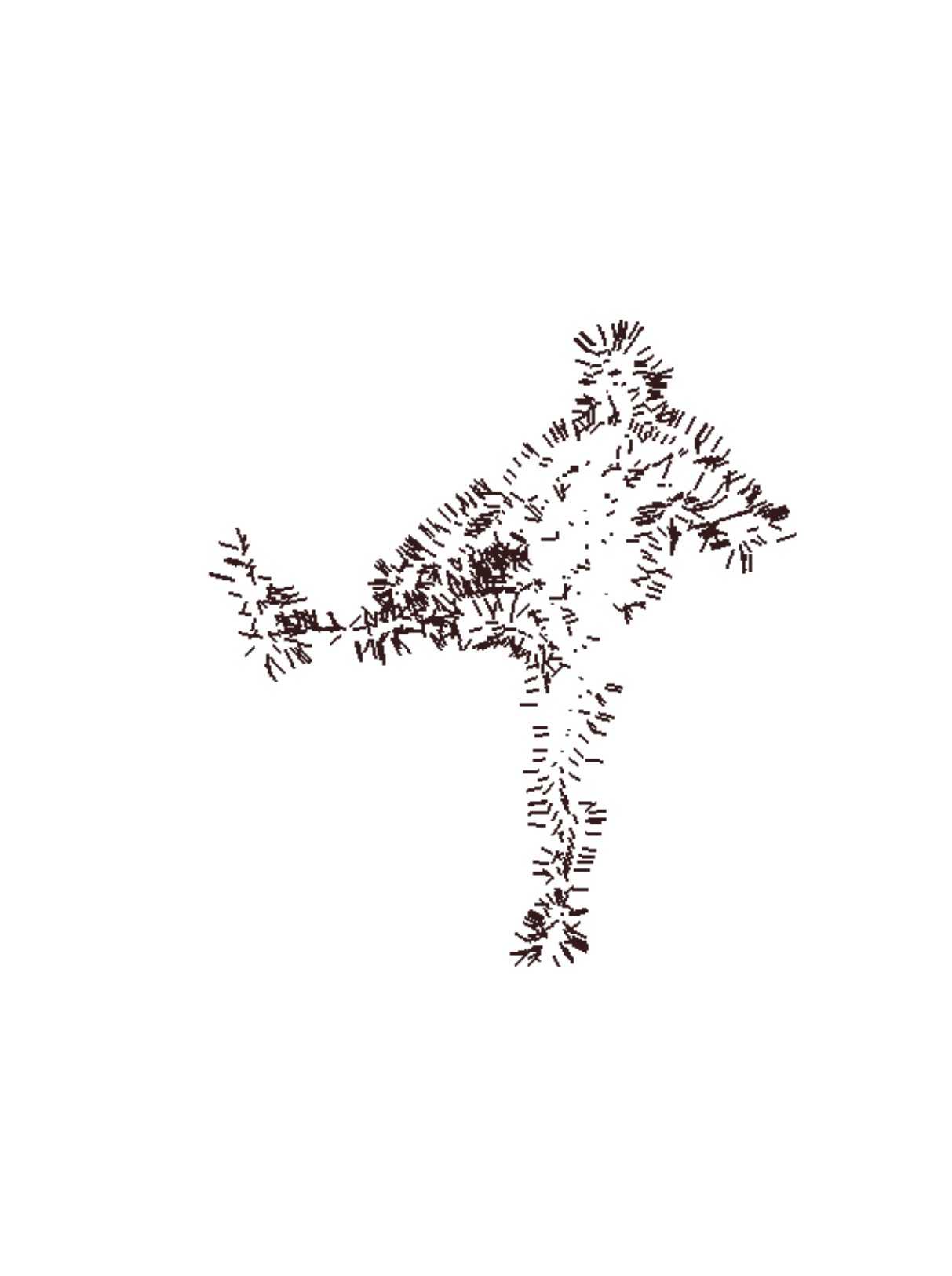}}&
\frame{\includegraphics[width=0.12\textwidth]{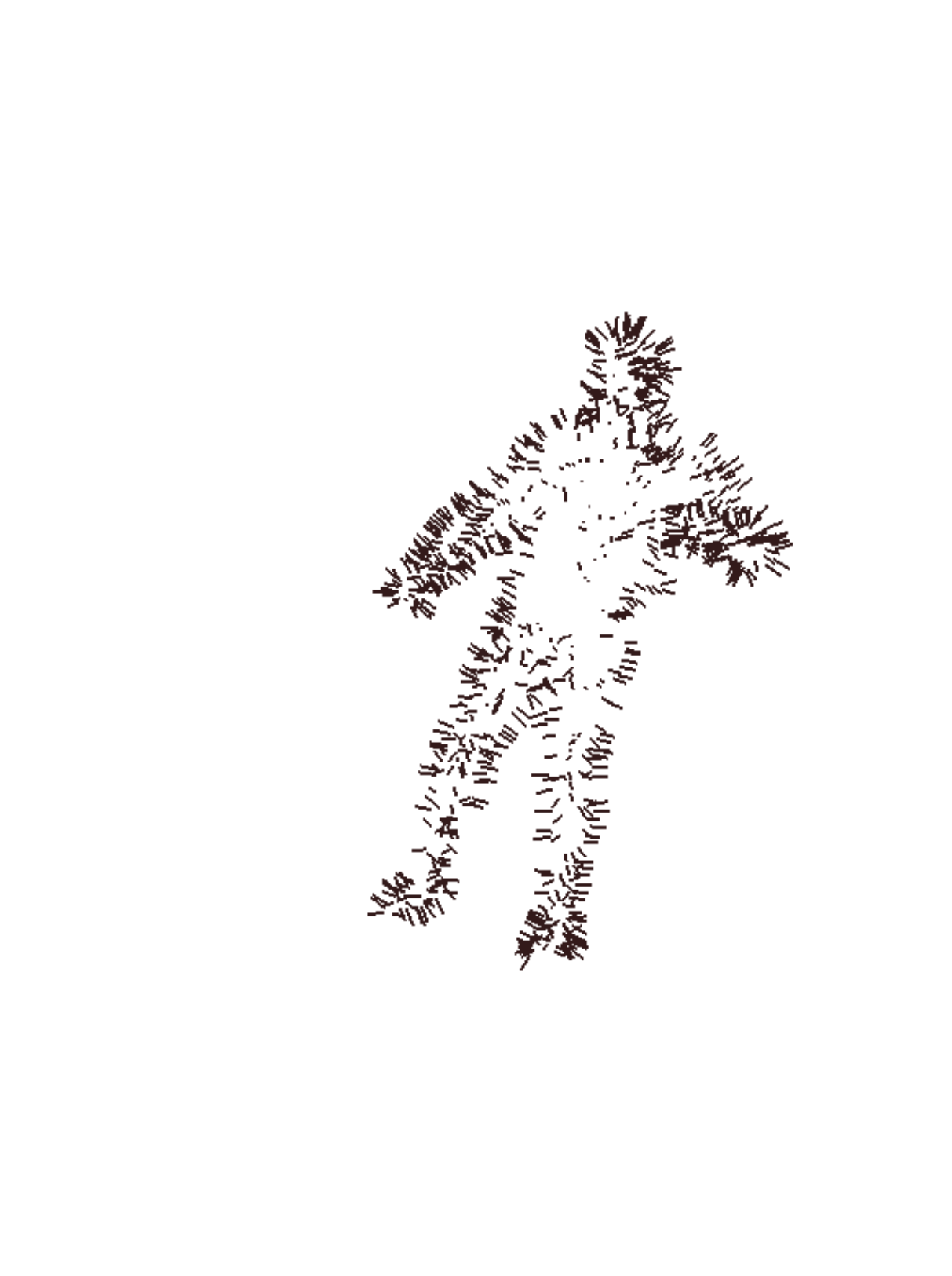}}&
\frame{\includegraphics[width=0.12\textwidth]{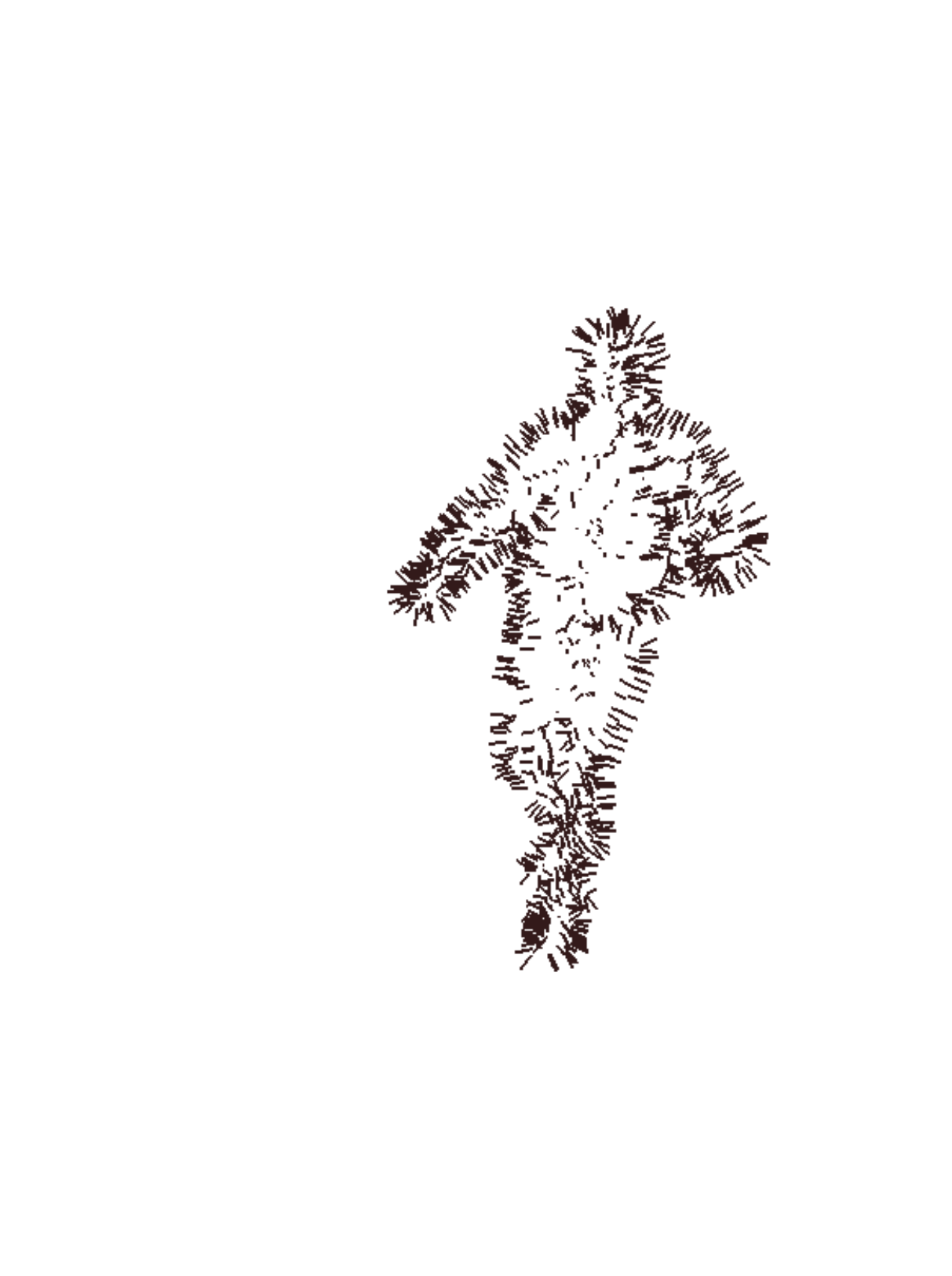}}
\\
\frame{\includegraphics[width=0.12\textwidth]{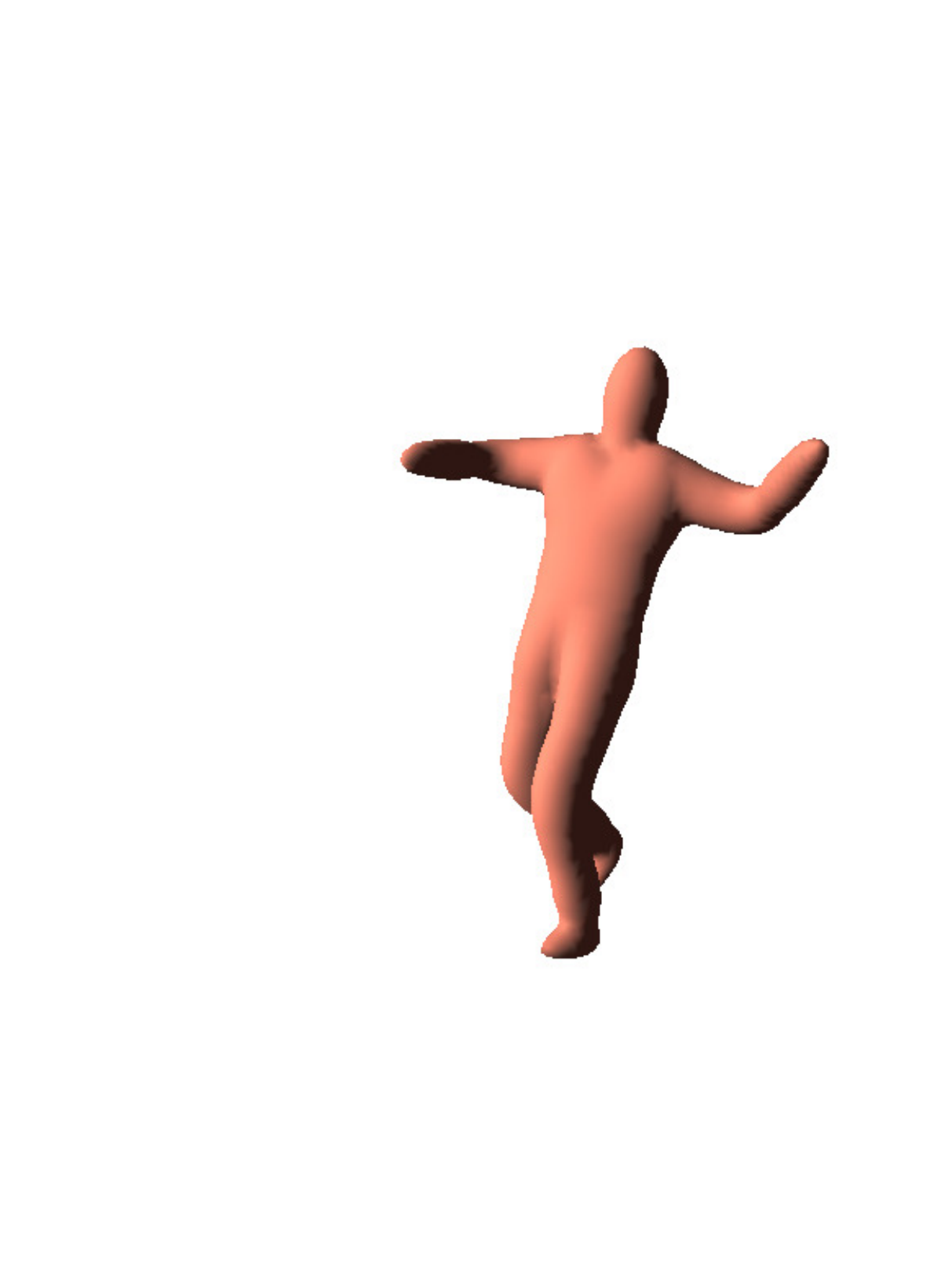}} &
\frame{\includegraphics[width=0.12\textwidth]{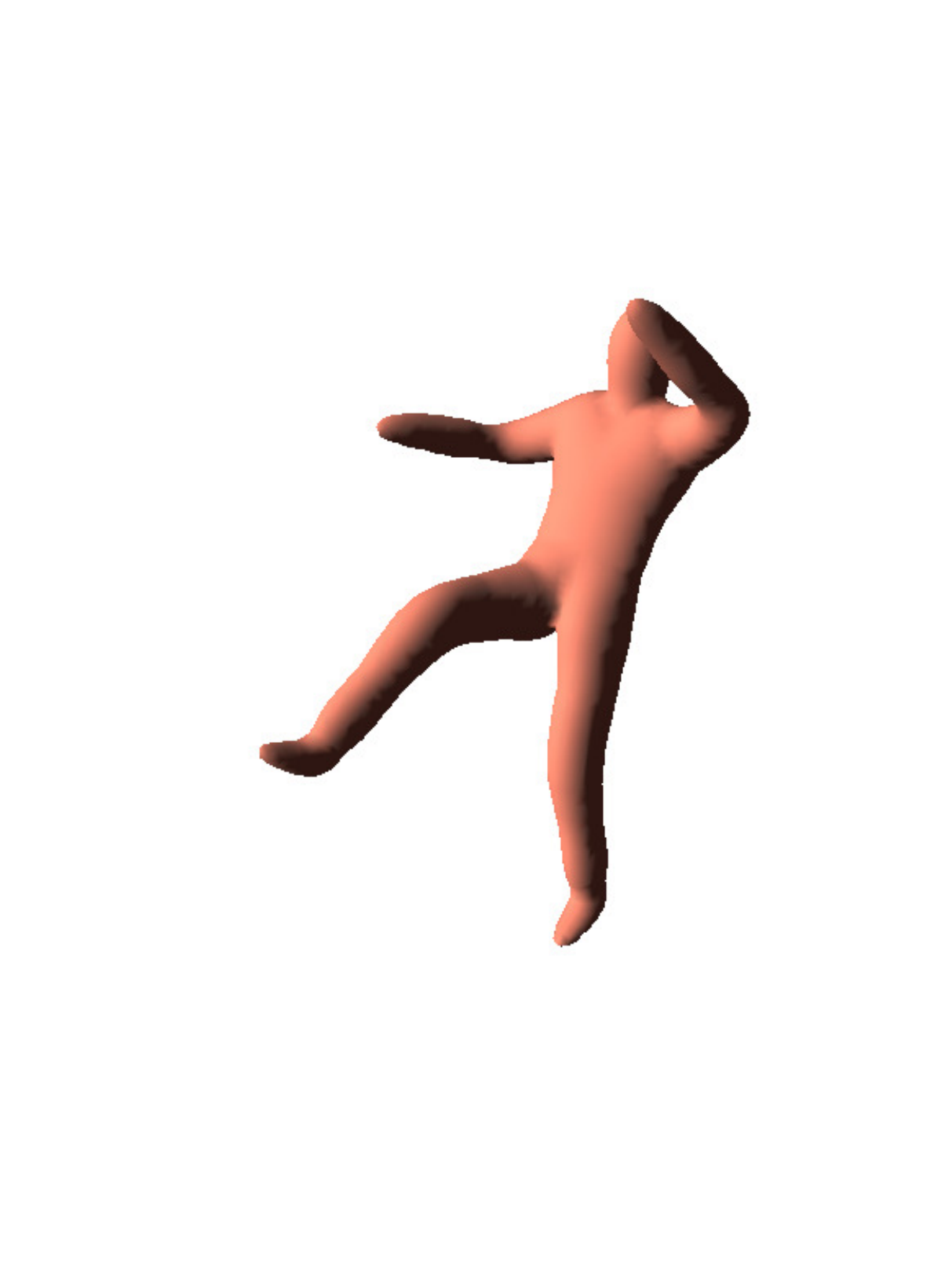}} &
\frame{\includegraphics[width=0.12\textwidth]{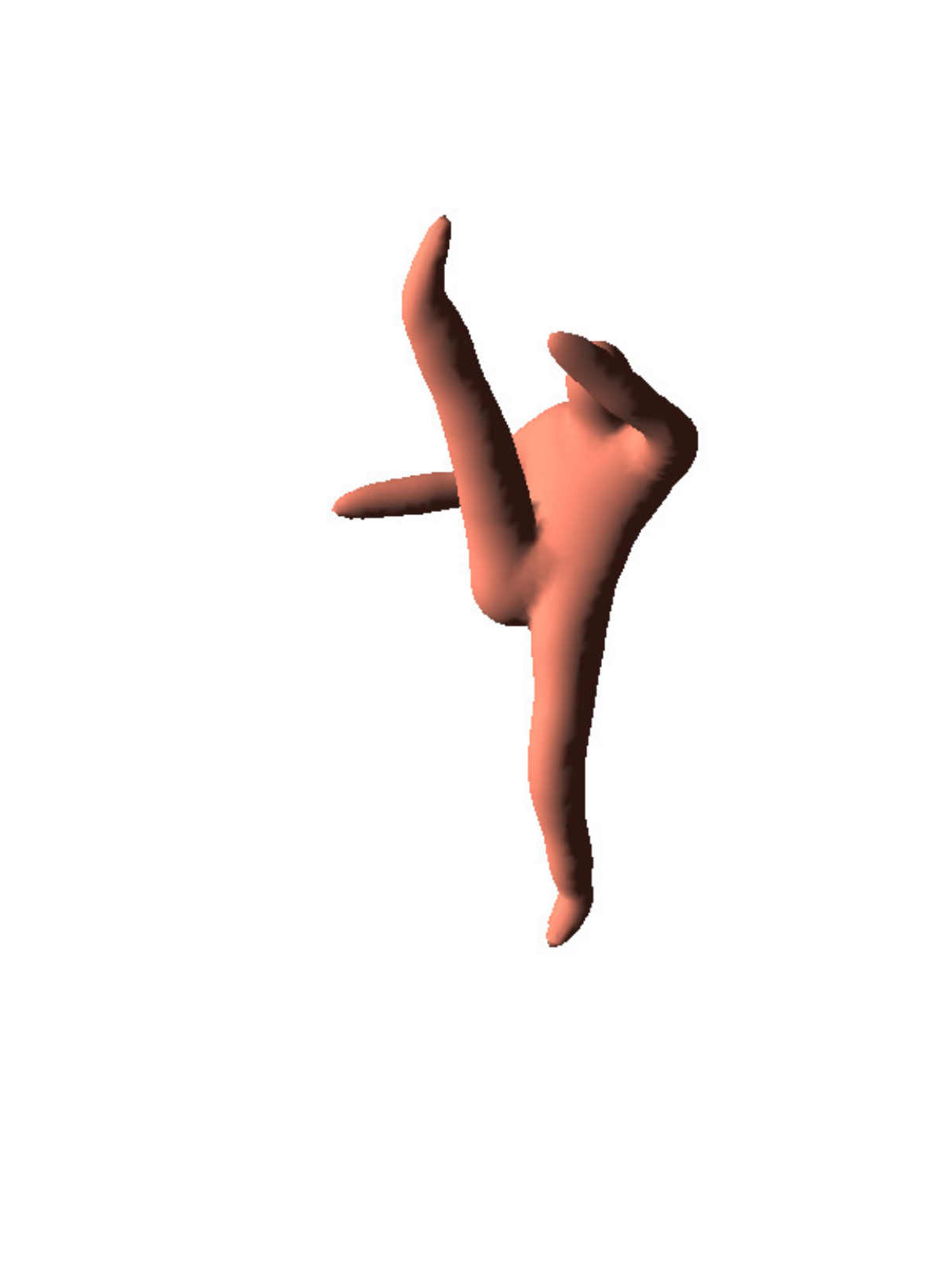}} &
\frame{\includegraphics[width=0.12\textwidth]{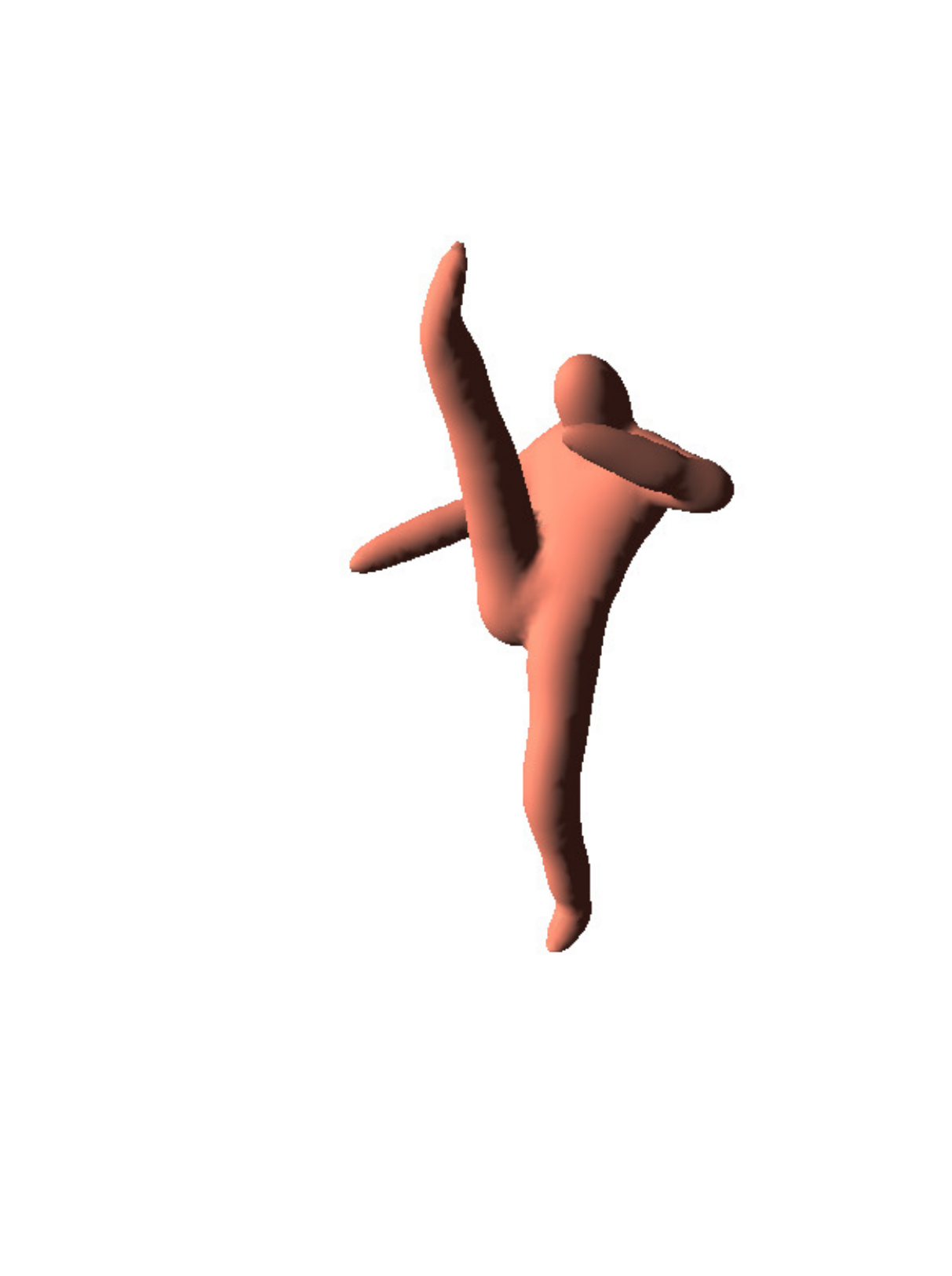}} &
\frame{\includegraphics[width=0.12\textwidth]{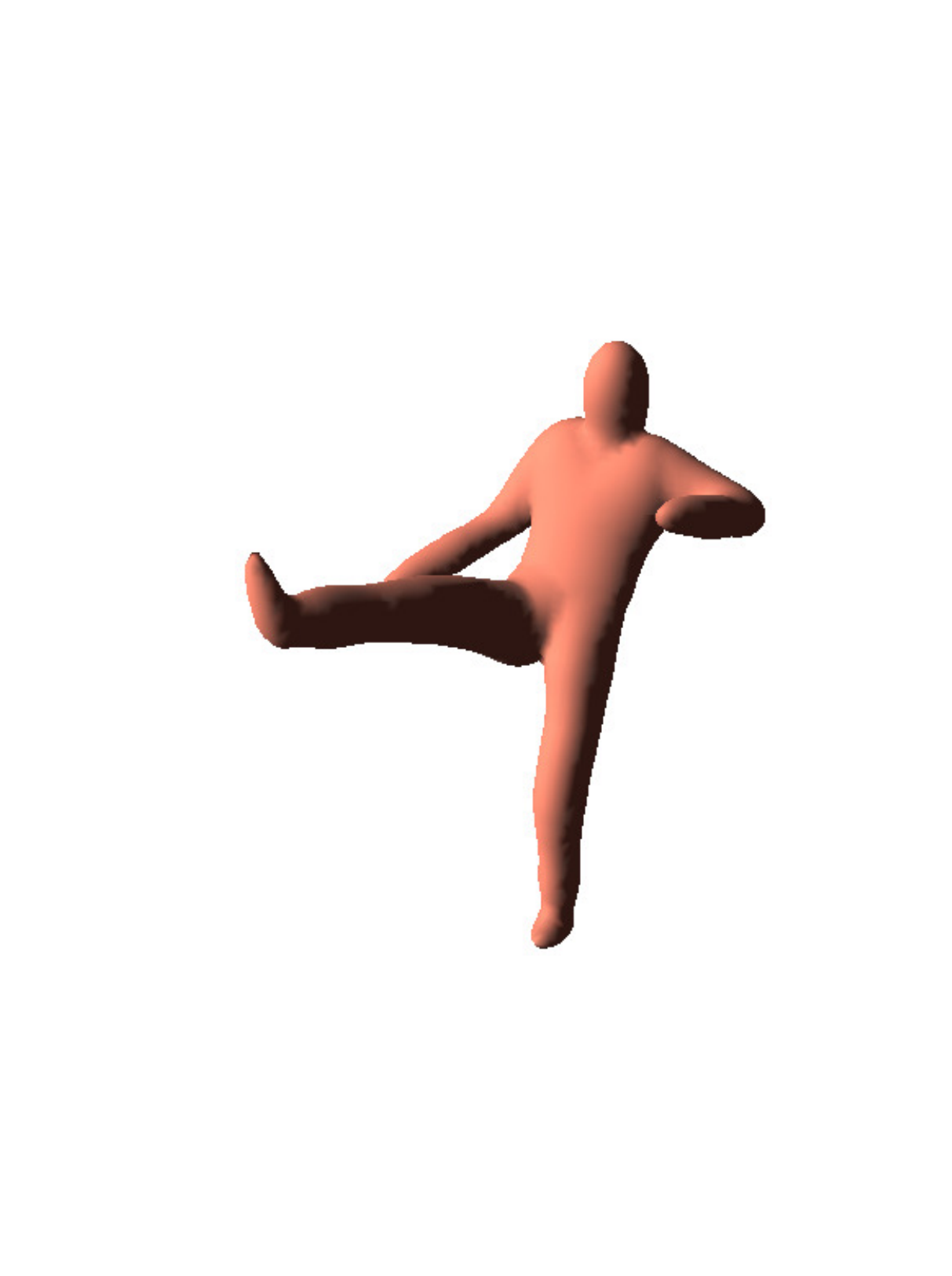}}&
\frame{\includegraphics[width=0.12\textwidth]{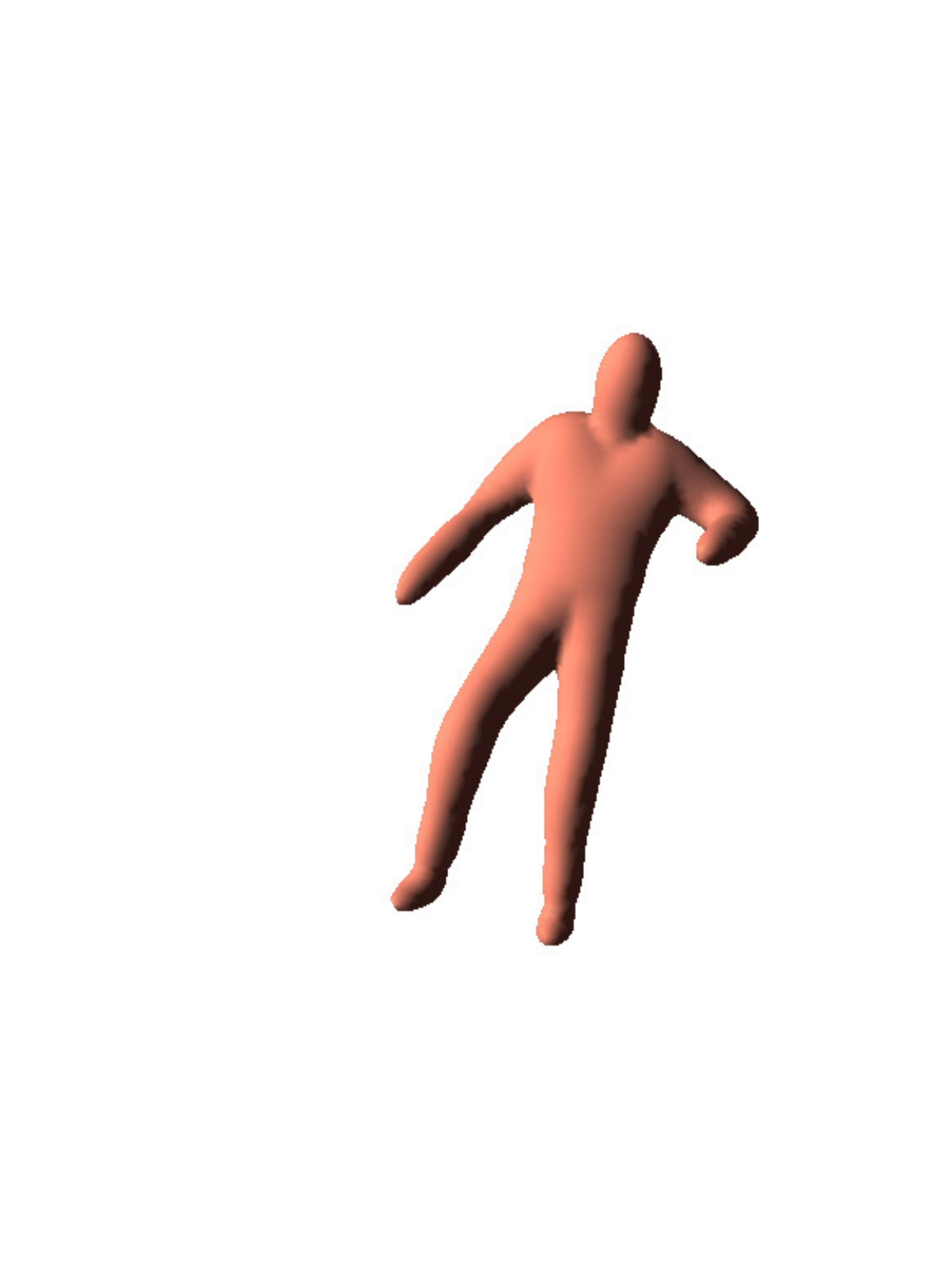}}&
\frame{\includegraphics[width=0.12\textwidth]{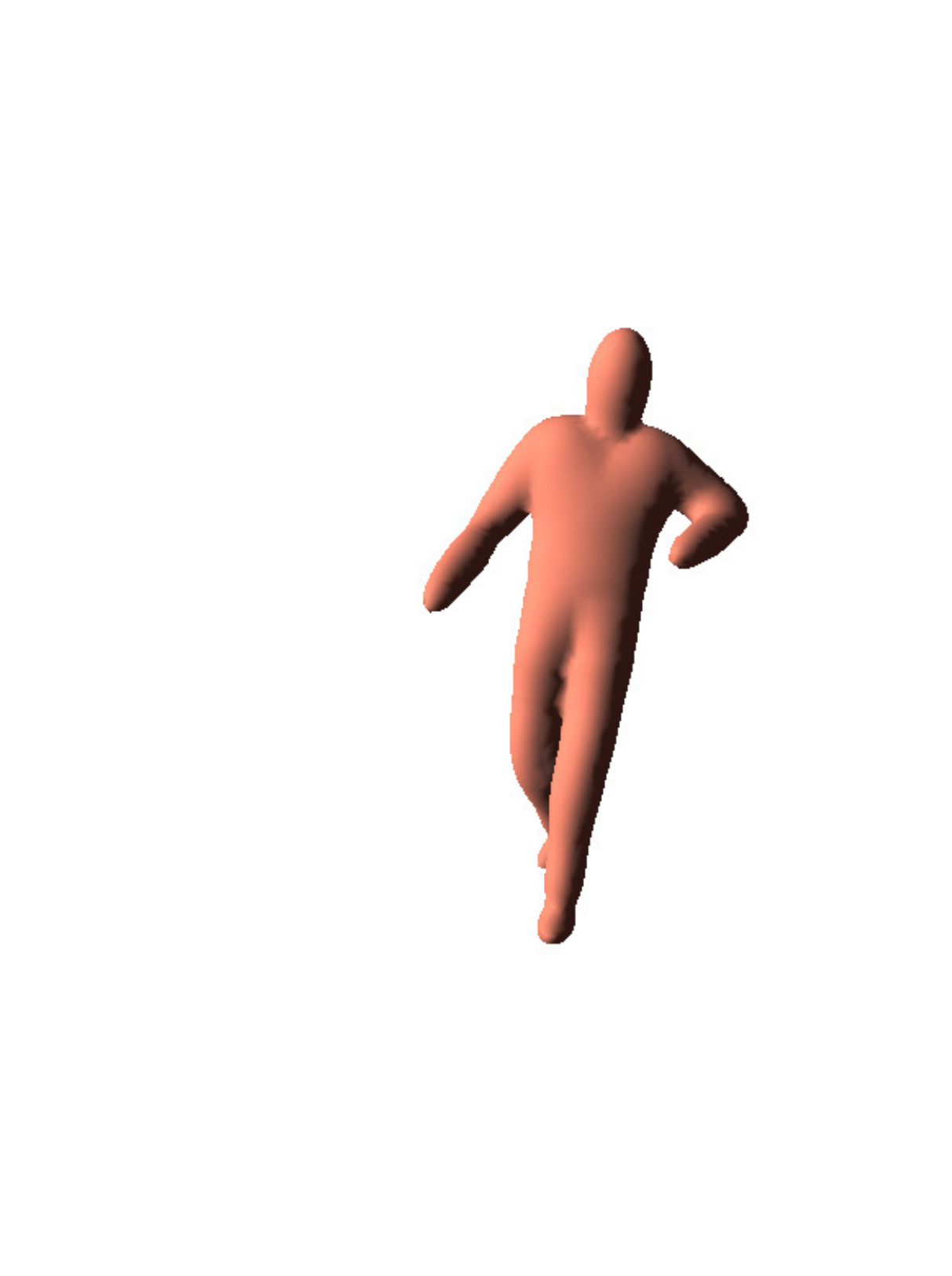}}
\\
\frame{\includegraphics[width=0.12\textwidth]{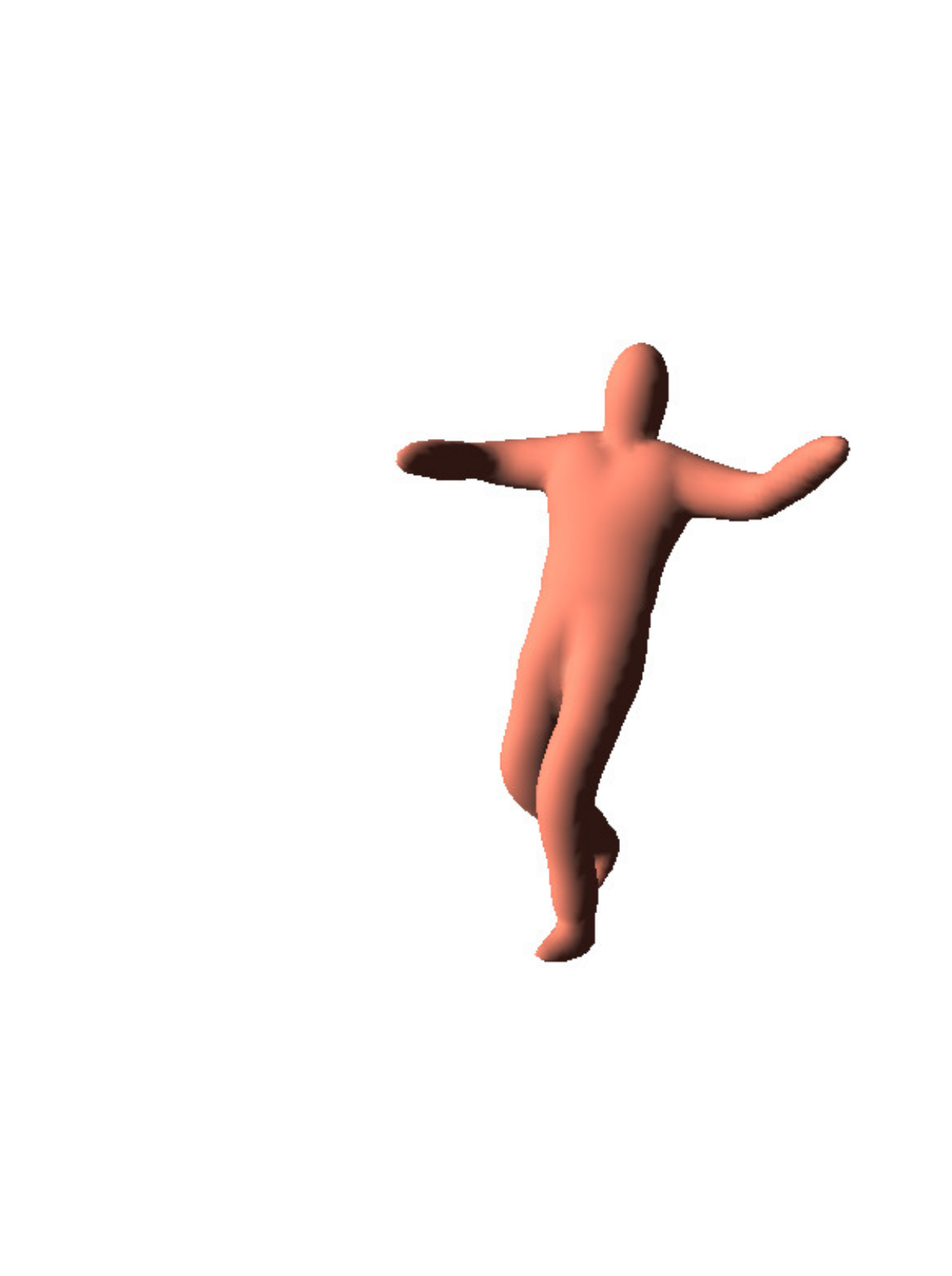}} &
\frame{\includegraphics[width=0.12\textwidth]{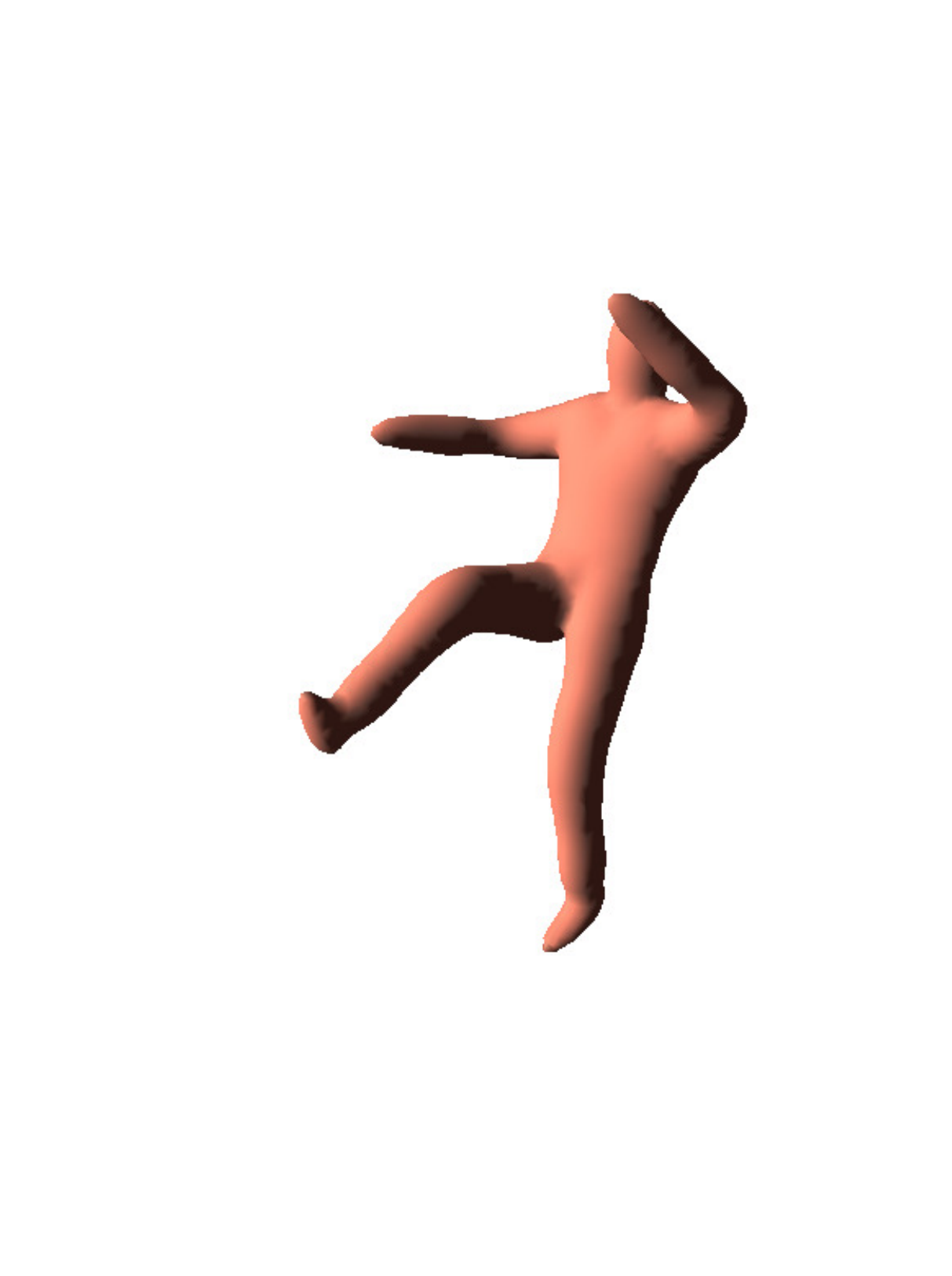}} &
\frame{\includegraphics[width=0.12\textwidth]{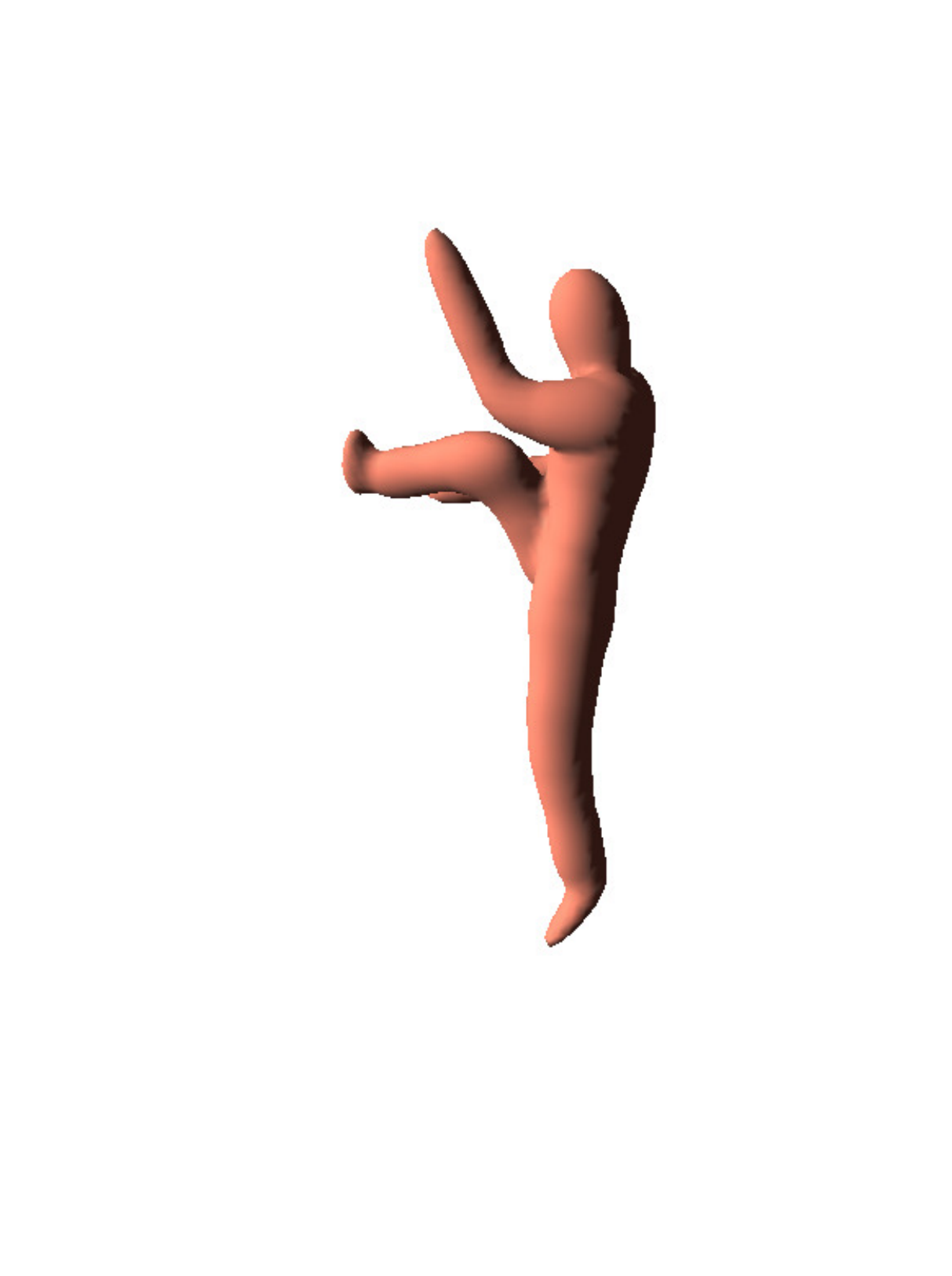}} &
\frame{\includegraphics[width=0.12\textwidth]{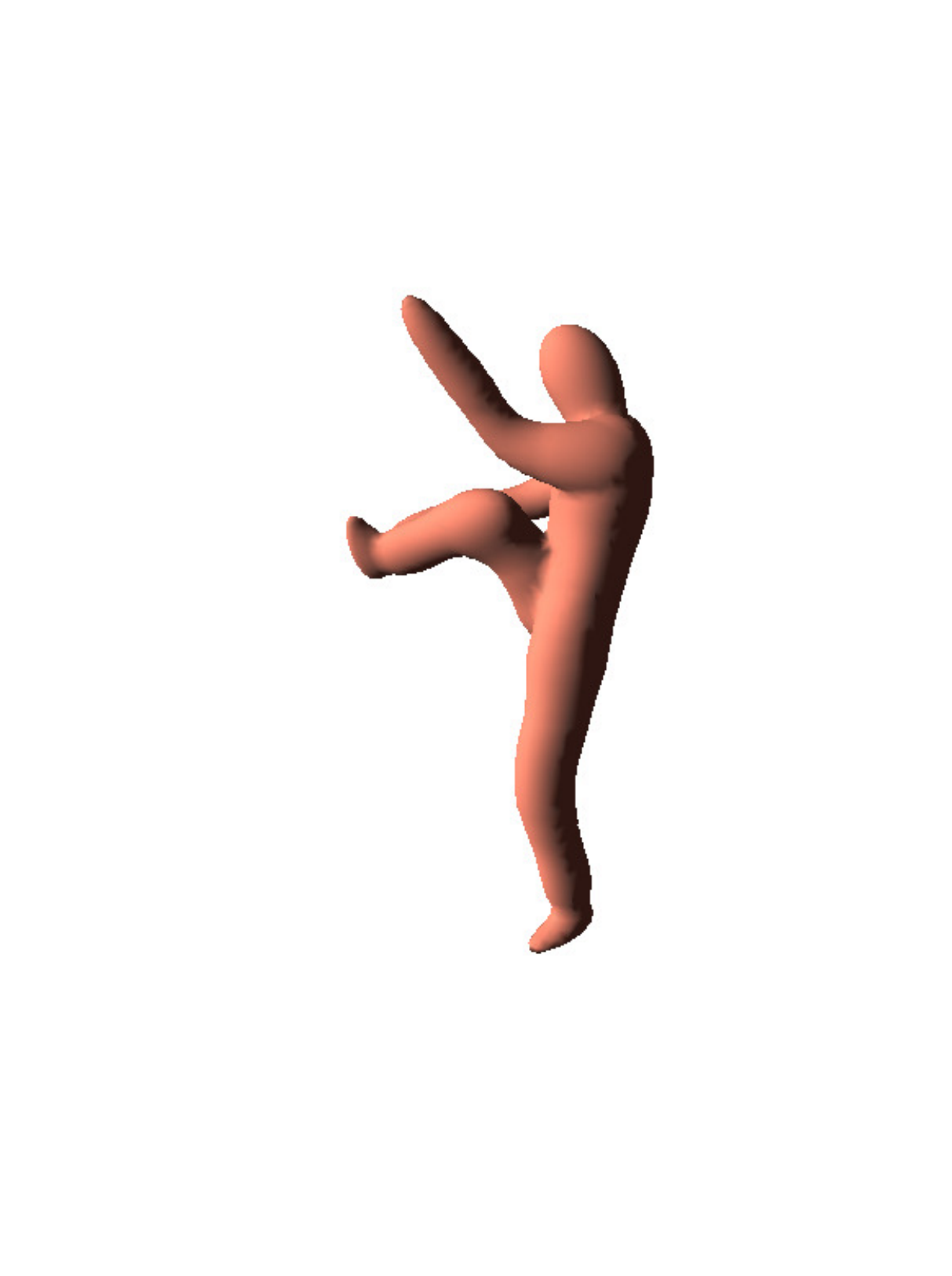}} &
\frame{\includegraphics[width=0.12\textwidth]{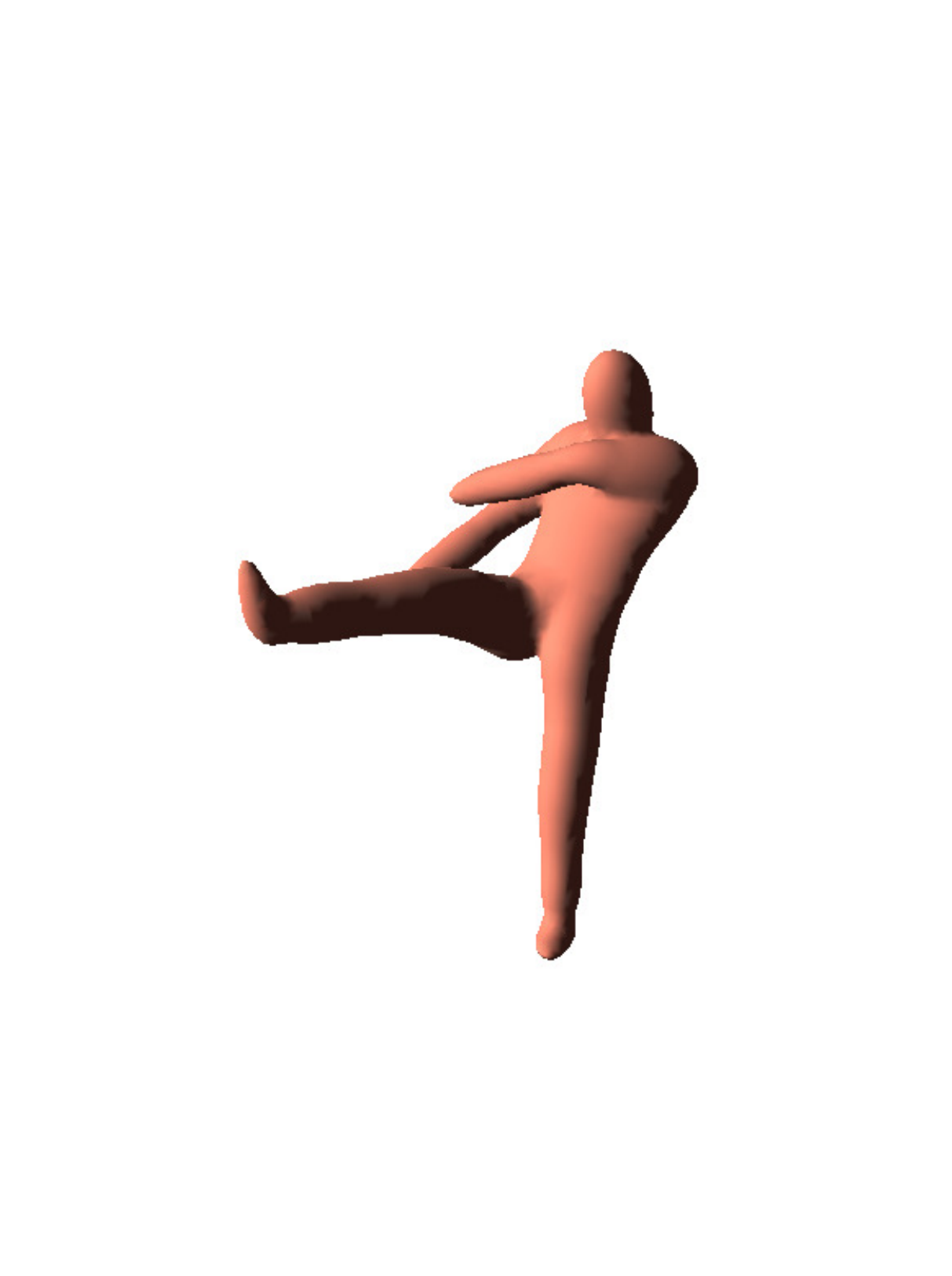}}&
\frame{\includegraphics[width=0.12\textwidth]{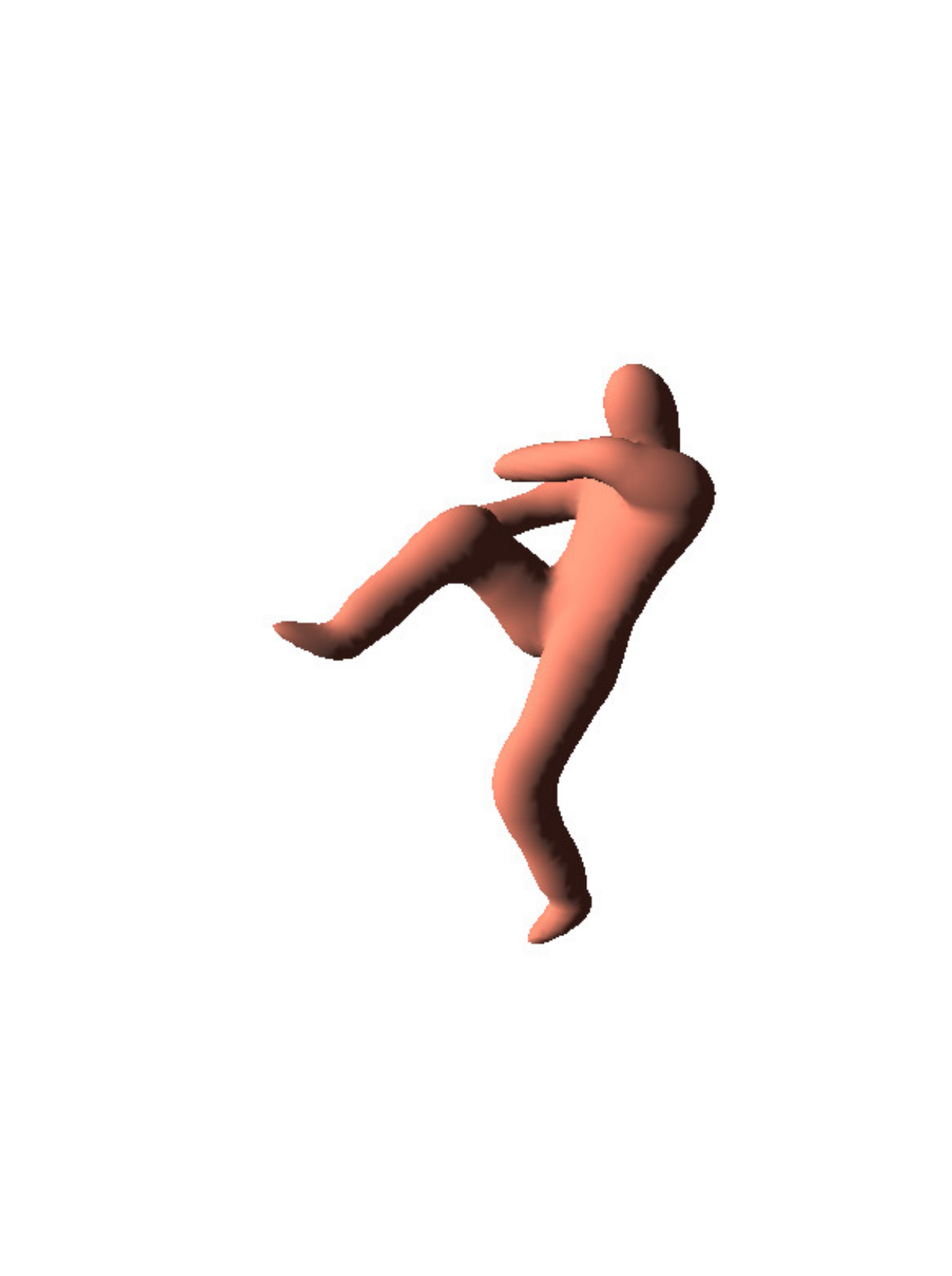}}&
\frame{\includegraphics[width=0.12\textwidth]{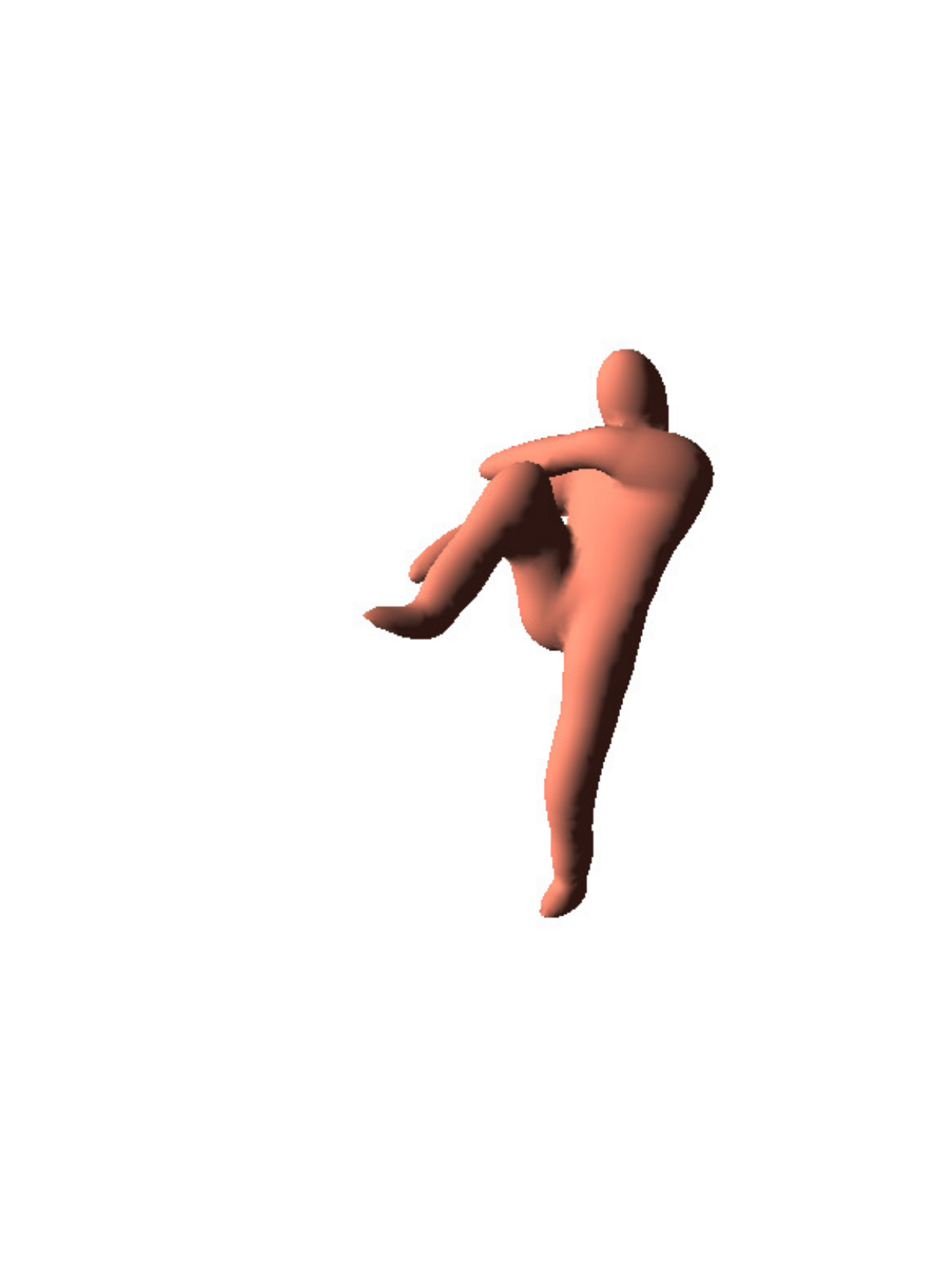}}
\\
\end{tabular}
\caption{The 3-D visual hull (top), the implicit surface fitted to the taekwendo
  sequence using points and normals (middle), and using
  points alone (bottom).} 
\label{fig:Ben-taekwendo-implicit-model}
\end{center}
\end{figure*}
\begin{figure*}[h!tb!]
\begin{center}
\begin{tabular}{ccccccc}
\frame{\includegraphics[width=0.12\textwidth]{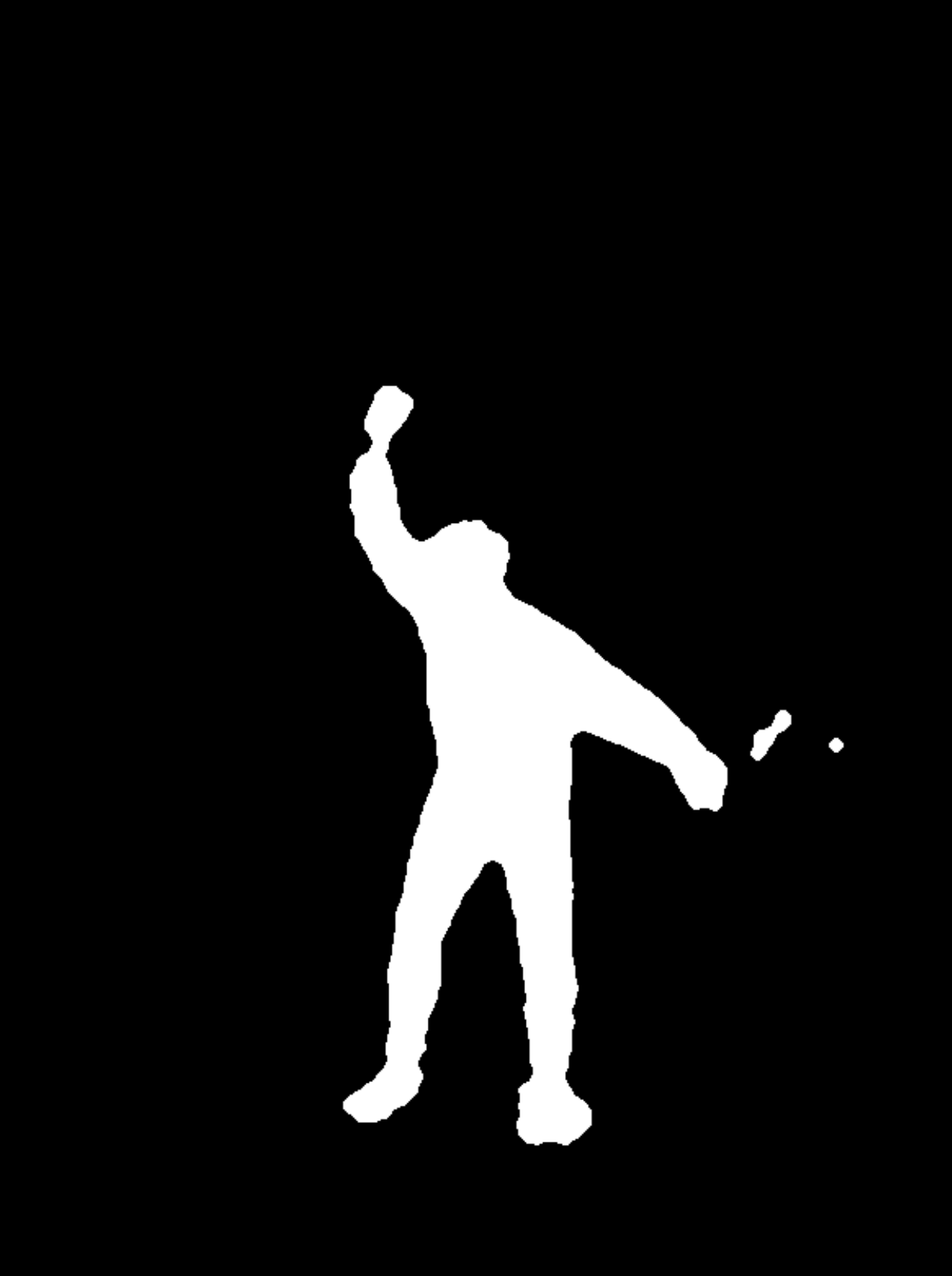}} &
\frame{\includegraphics[width=0.12\textwidth]{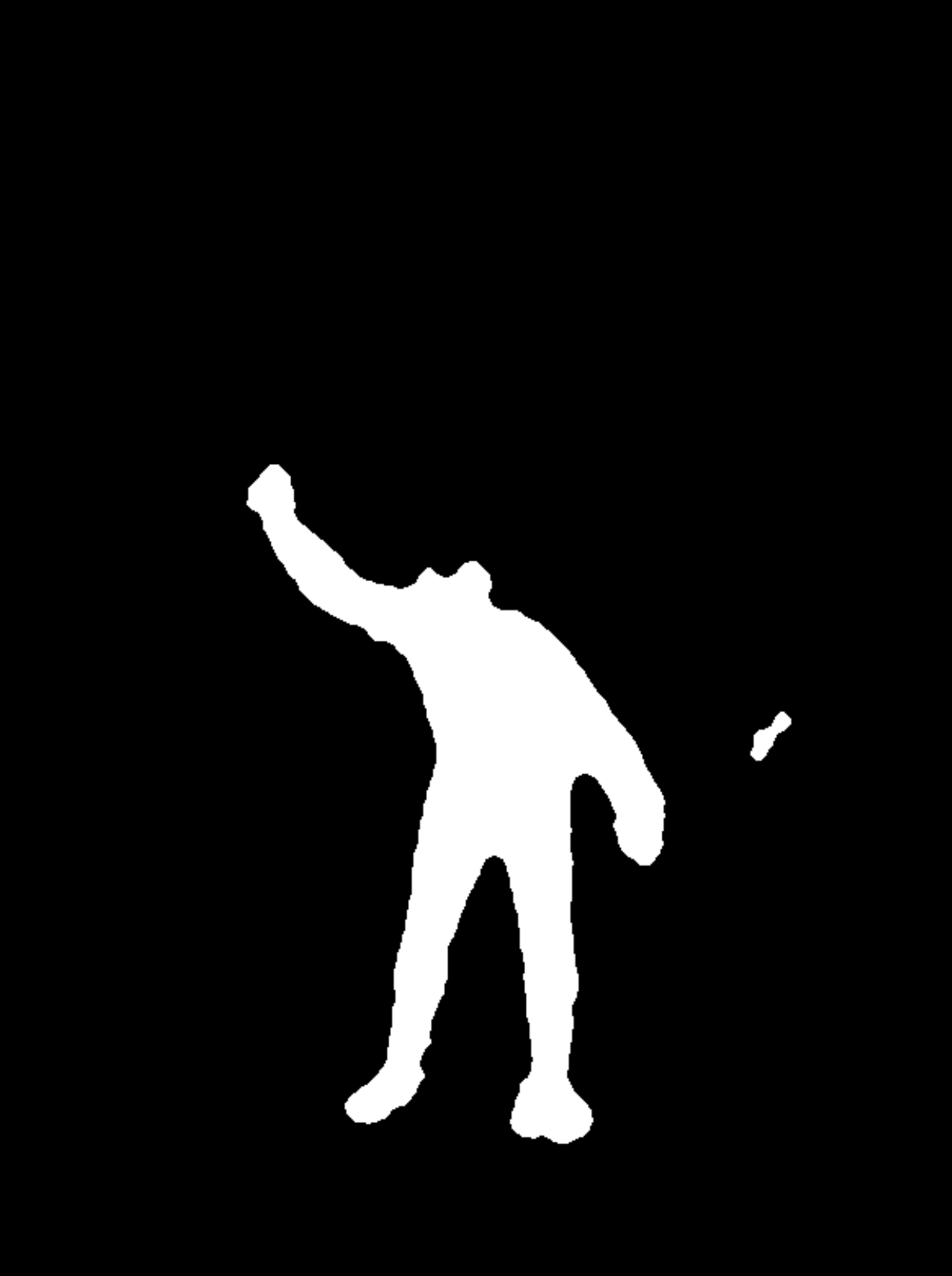}} &
\frame{\includegraphics[width=0.12\textwidth]{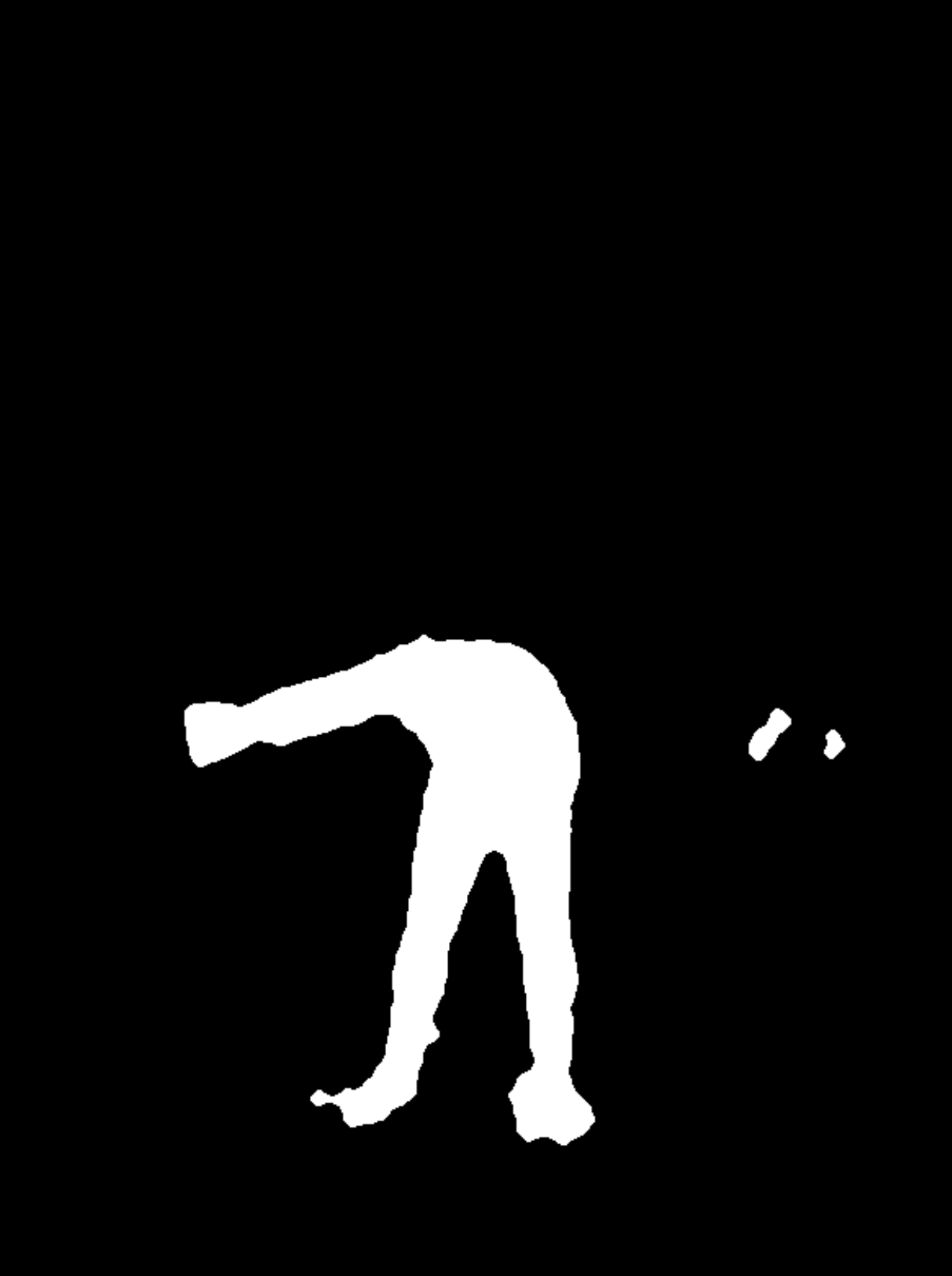}} &
\frame{\includegraphics[width=0.12\textwidth]{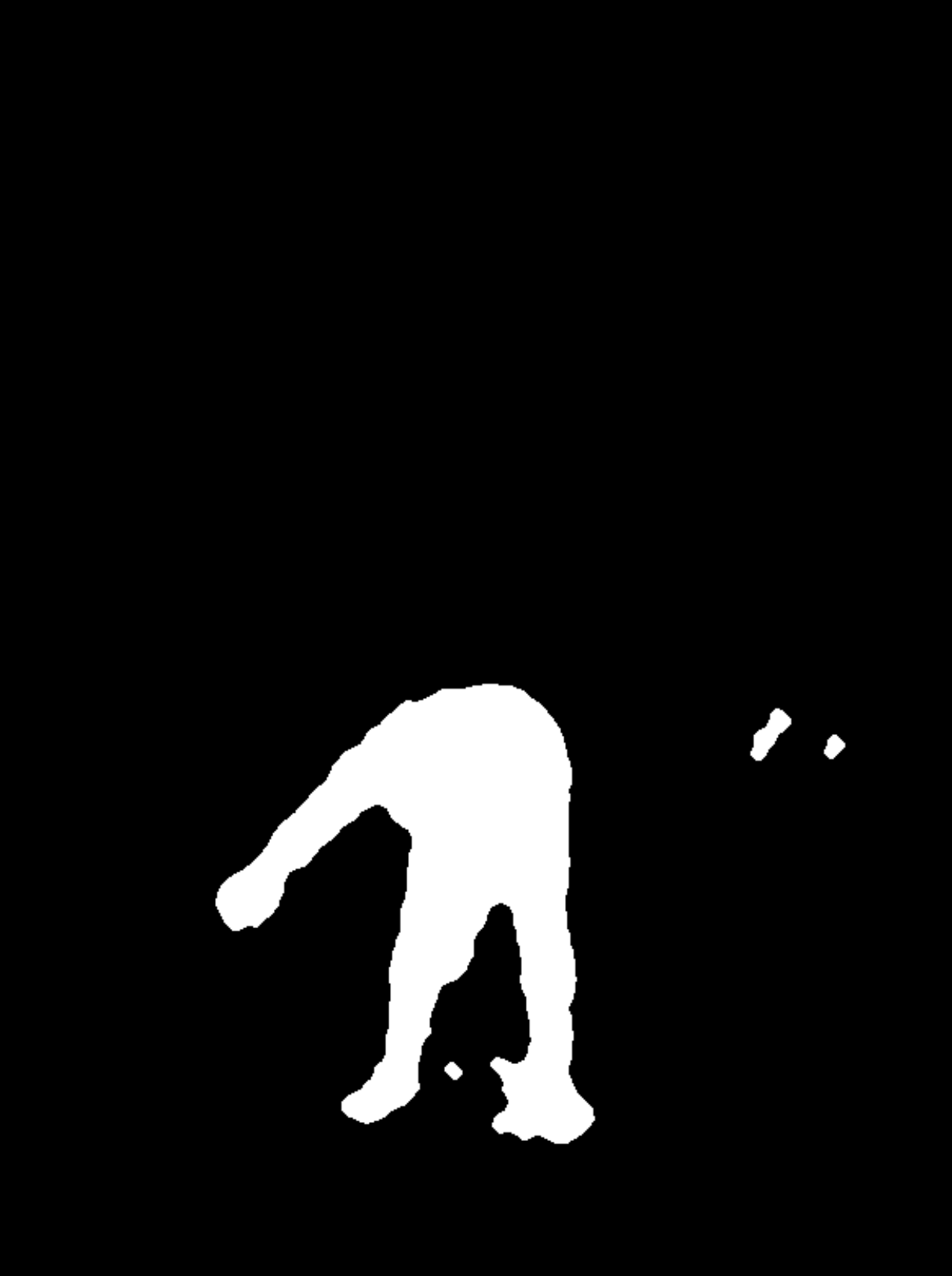}} &
\frame{\includegraphics[width=0.12\textwidth]{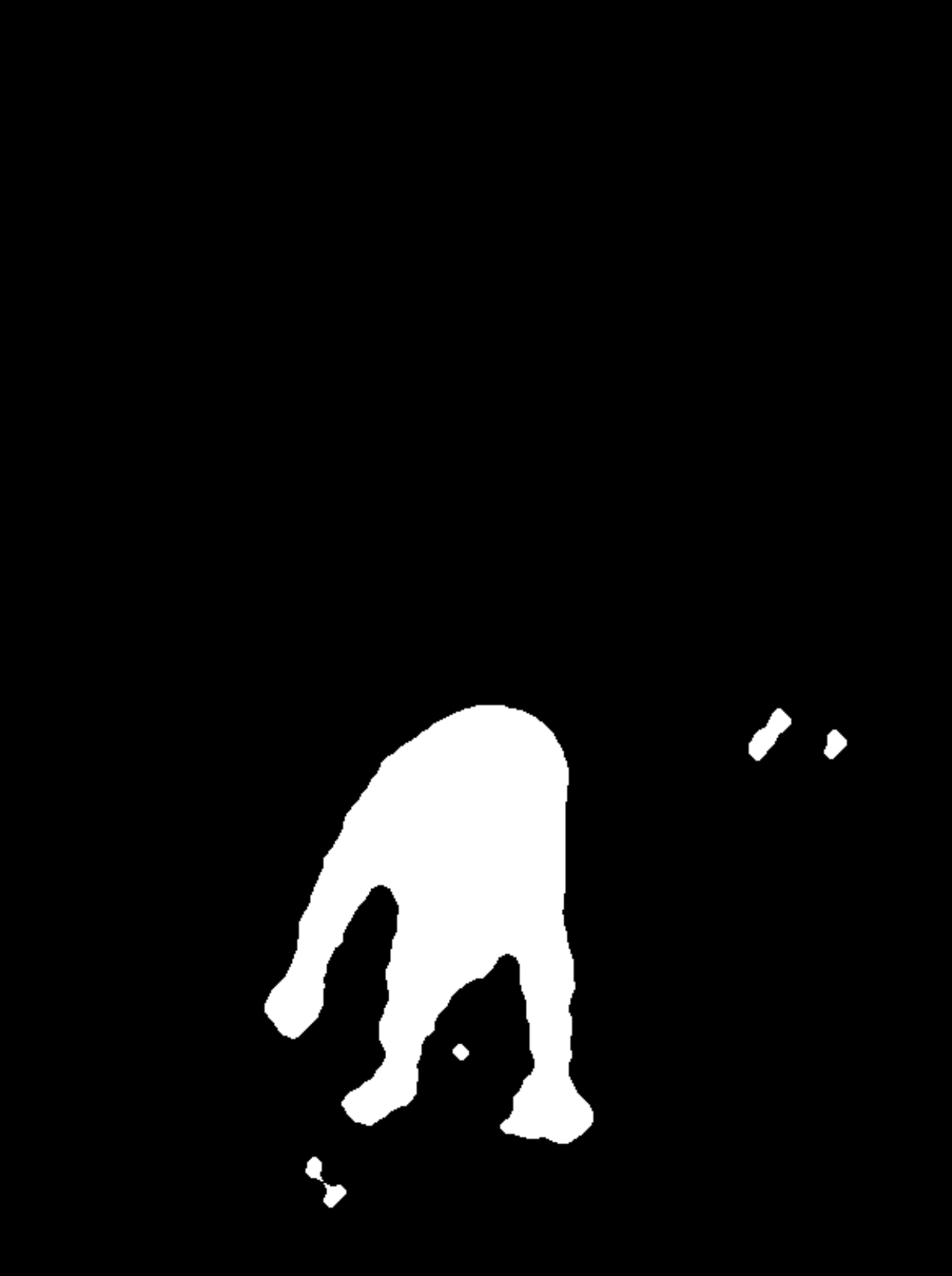}}&
\frame{\includegraphics[width=0.12\textwidth]{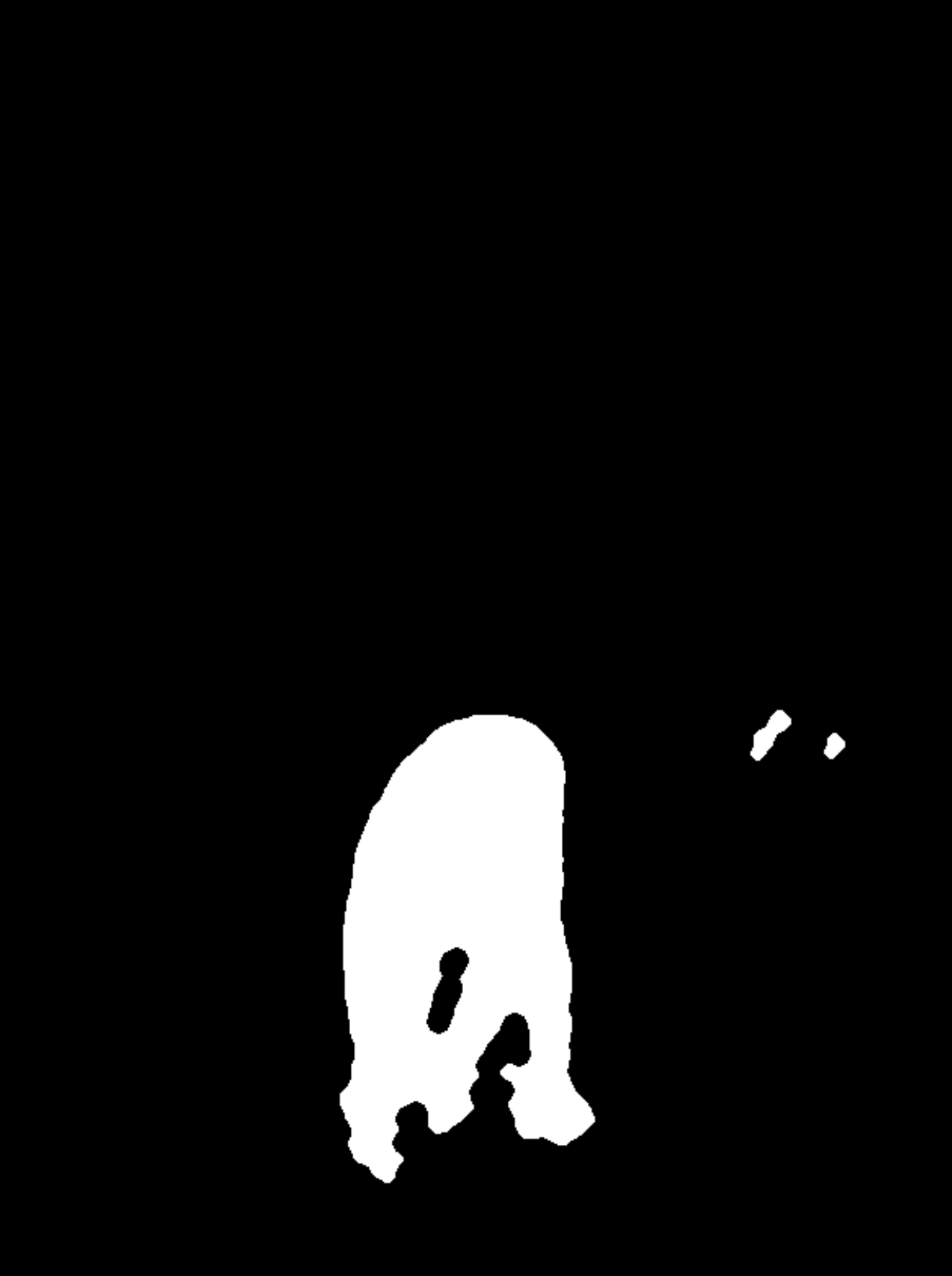}}&
\frame{\includegraphics[width=0.12\textwidth]{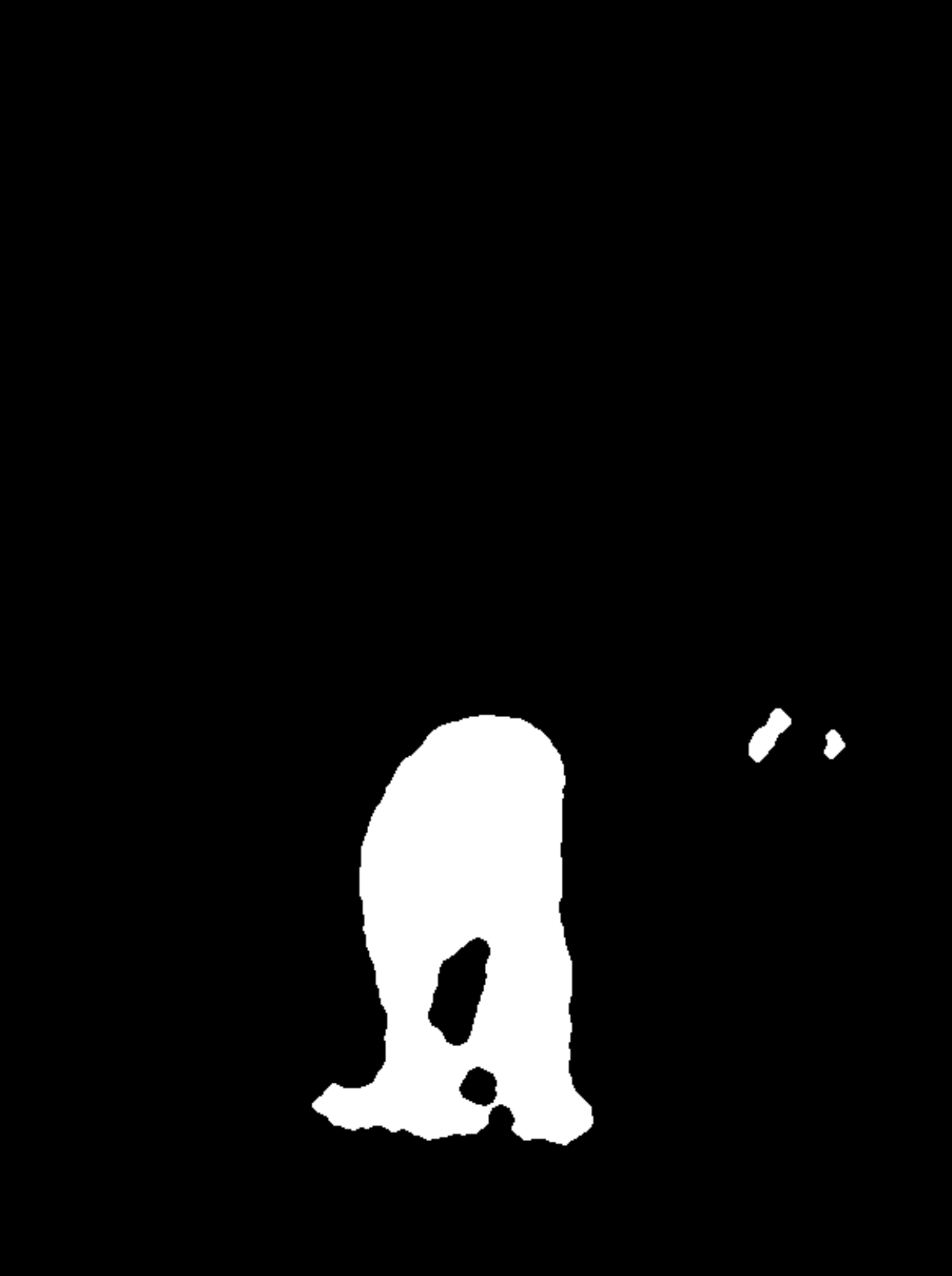}}
\\
\frame{\includegraphics[width=0.12\textwidth]{cam1_078-eps-converted-to.pdf}} &
\frame{\includegraphics[width=0.12\textwidth]{cam1_080-eps-converted-to.pdf}} &
\frame{\includegraphics[width=0.12\textwidth]{cam1_082-eps-converted-to.pdf}} &
\frame{\includegraphics[width=0.12\textwidth]{cam1_084-eps-converted-to.pdf}} &
\frame{\includegraphics[width=0.12\textwidth]{cam1_086-eps-converted-to.pdf}}&
\frame{\includegraphics[width=0.12\textwidth]{cam1_088-eps-converted-to.pdf}}&
\frame{\includegraphics[width=0.12\textwidth]{cam1_090-eps-converted-to.pdf}}
\\
\frame{\includegraphics[width=0.12\textwidth]{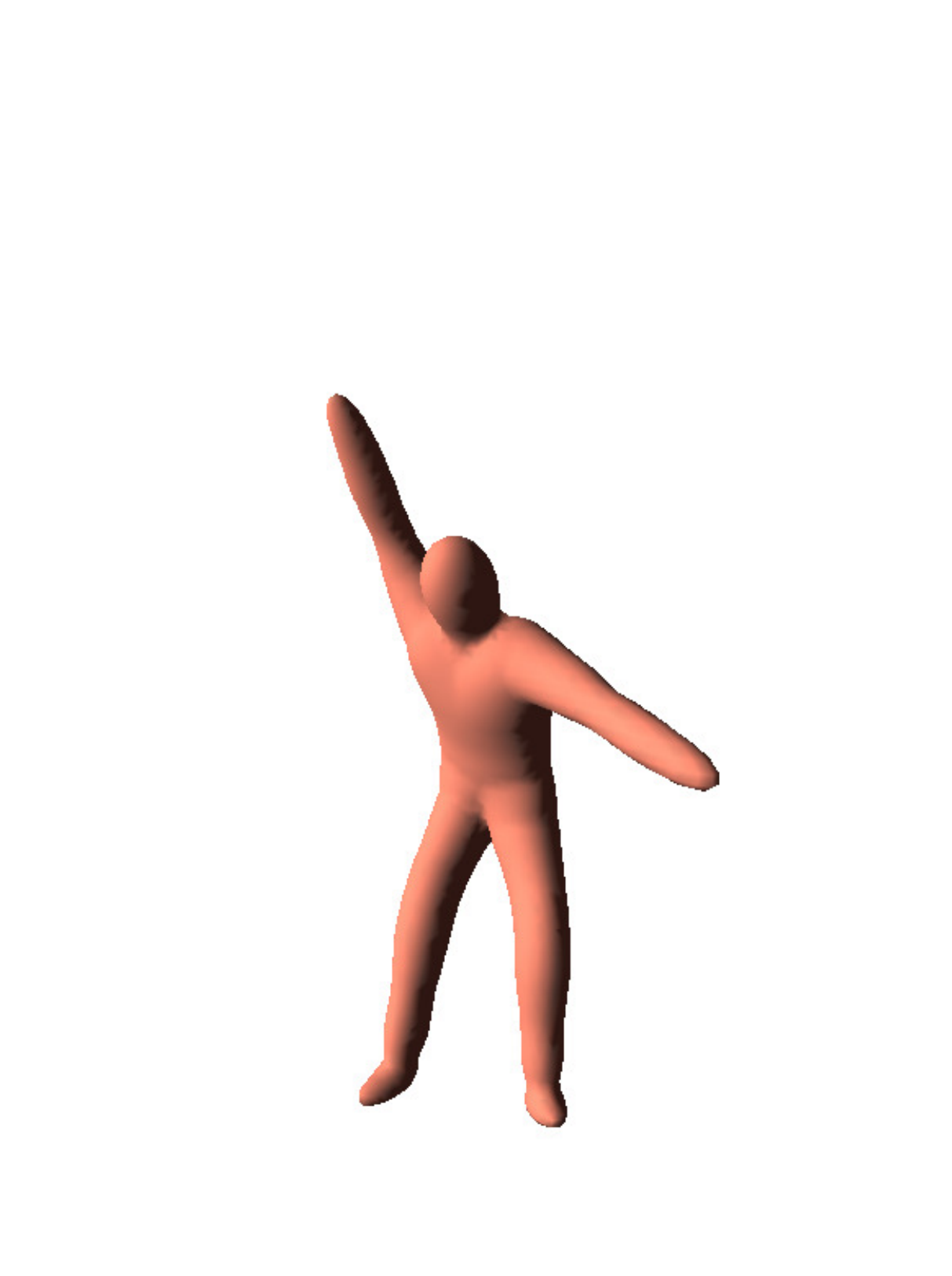}} &
\frame{\includegraphics[width=0.12\textwidth]{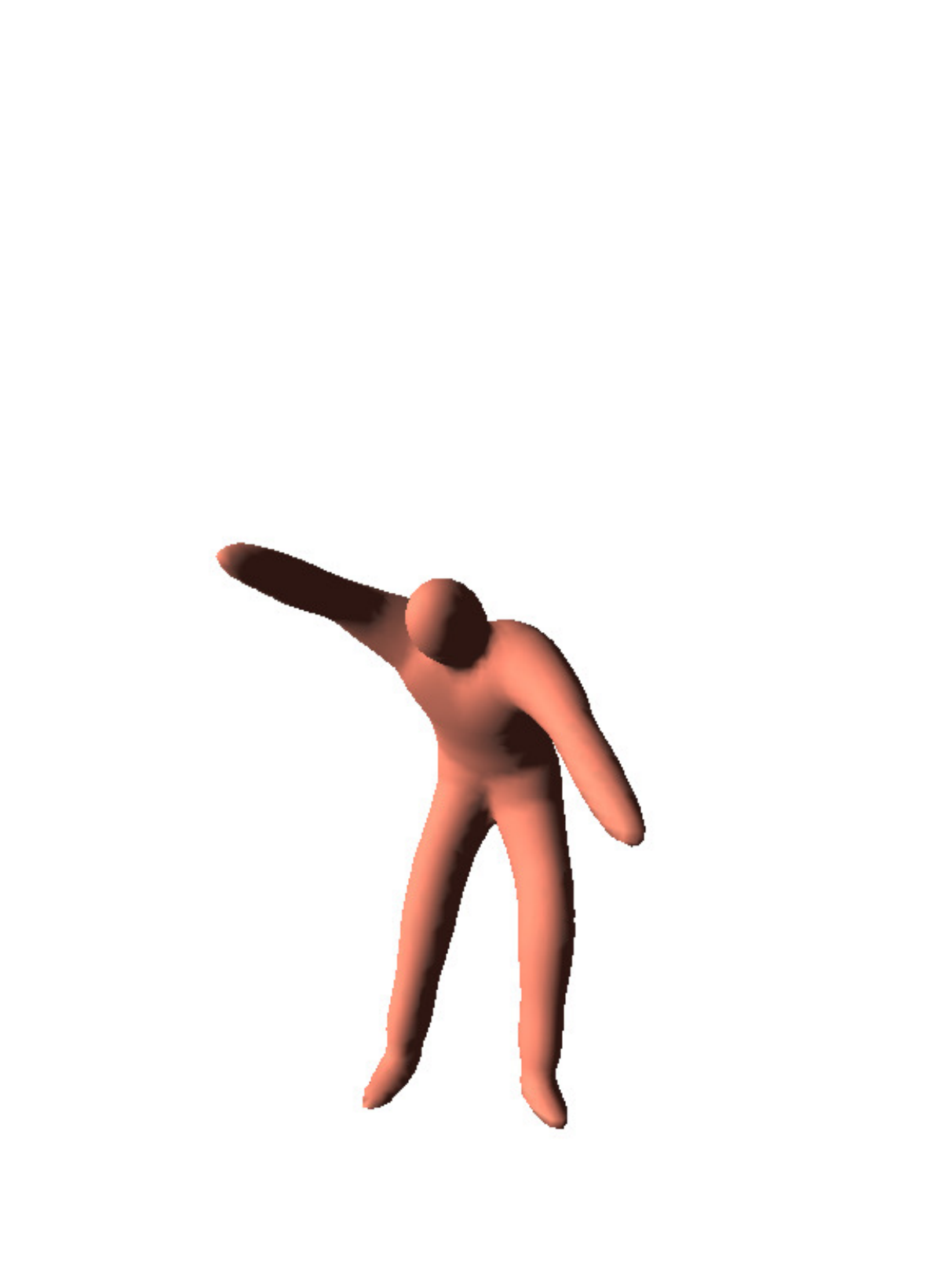}} &
\frame{\includegraphics[width=0.12\textwidth]{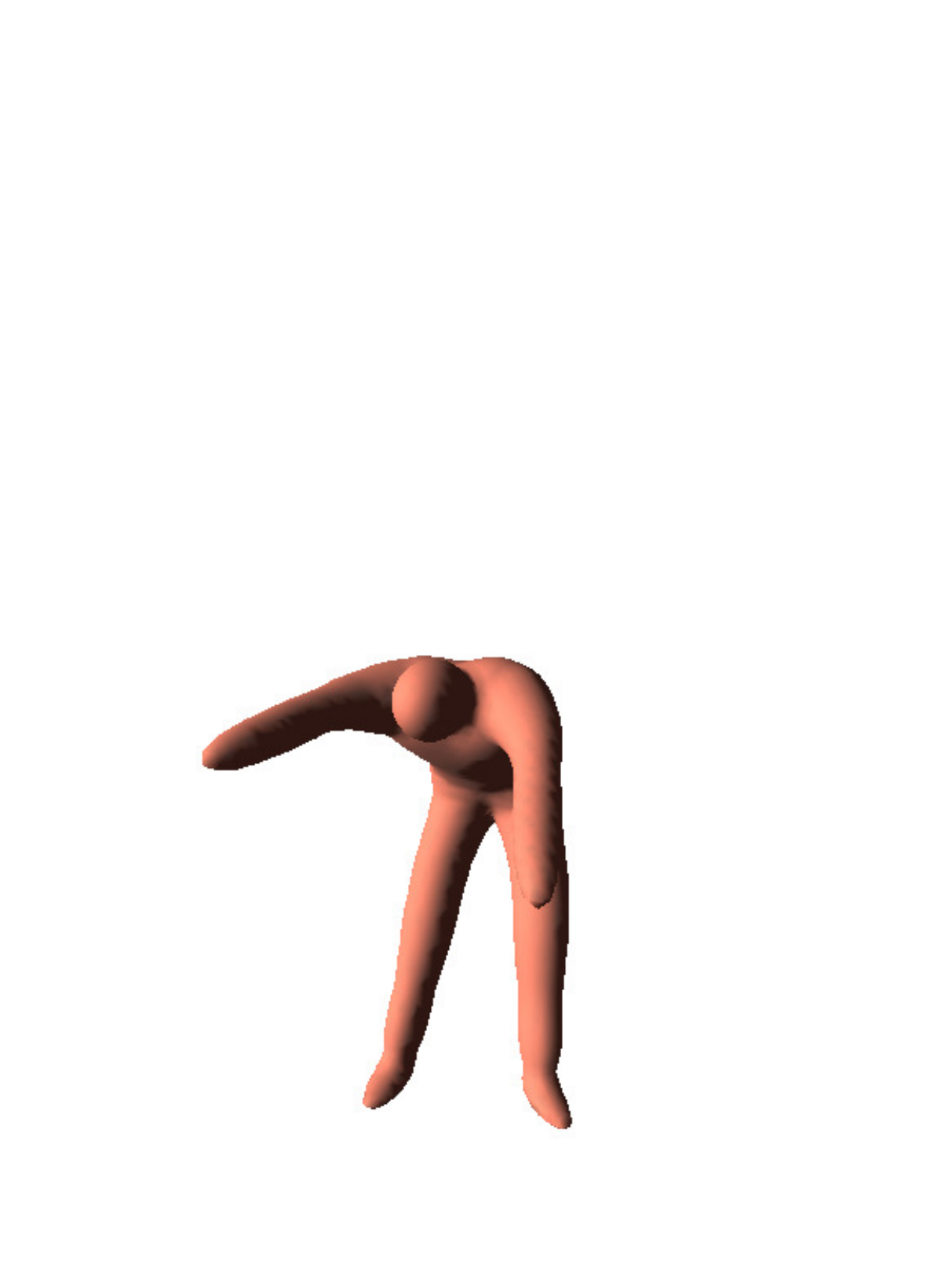}} &
\frame{\includegraphics[width=0.12\textwidth]{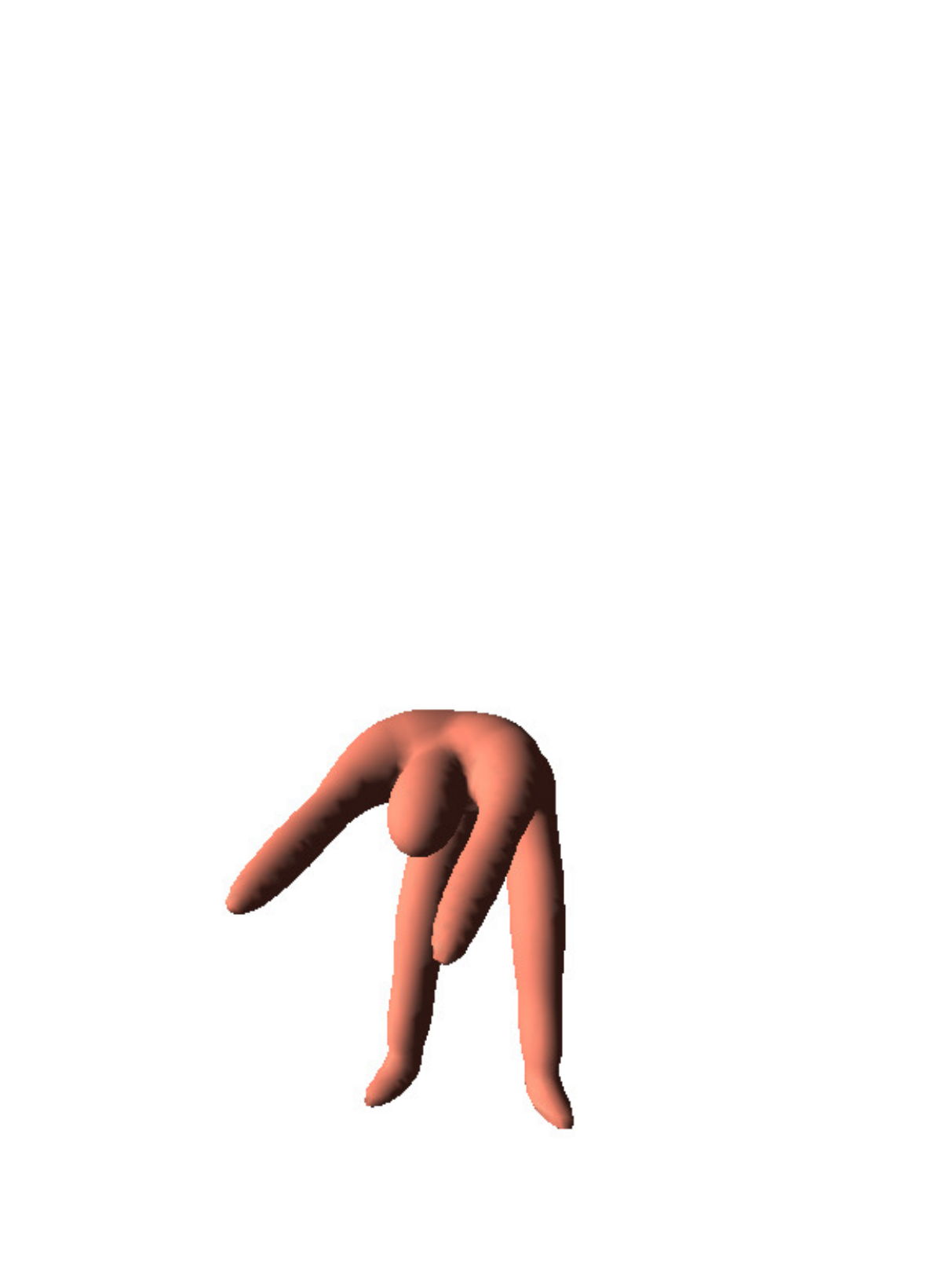}} &
\frame{\includegraphics[width=0.12\textwidth]{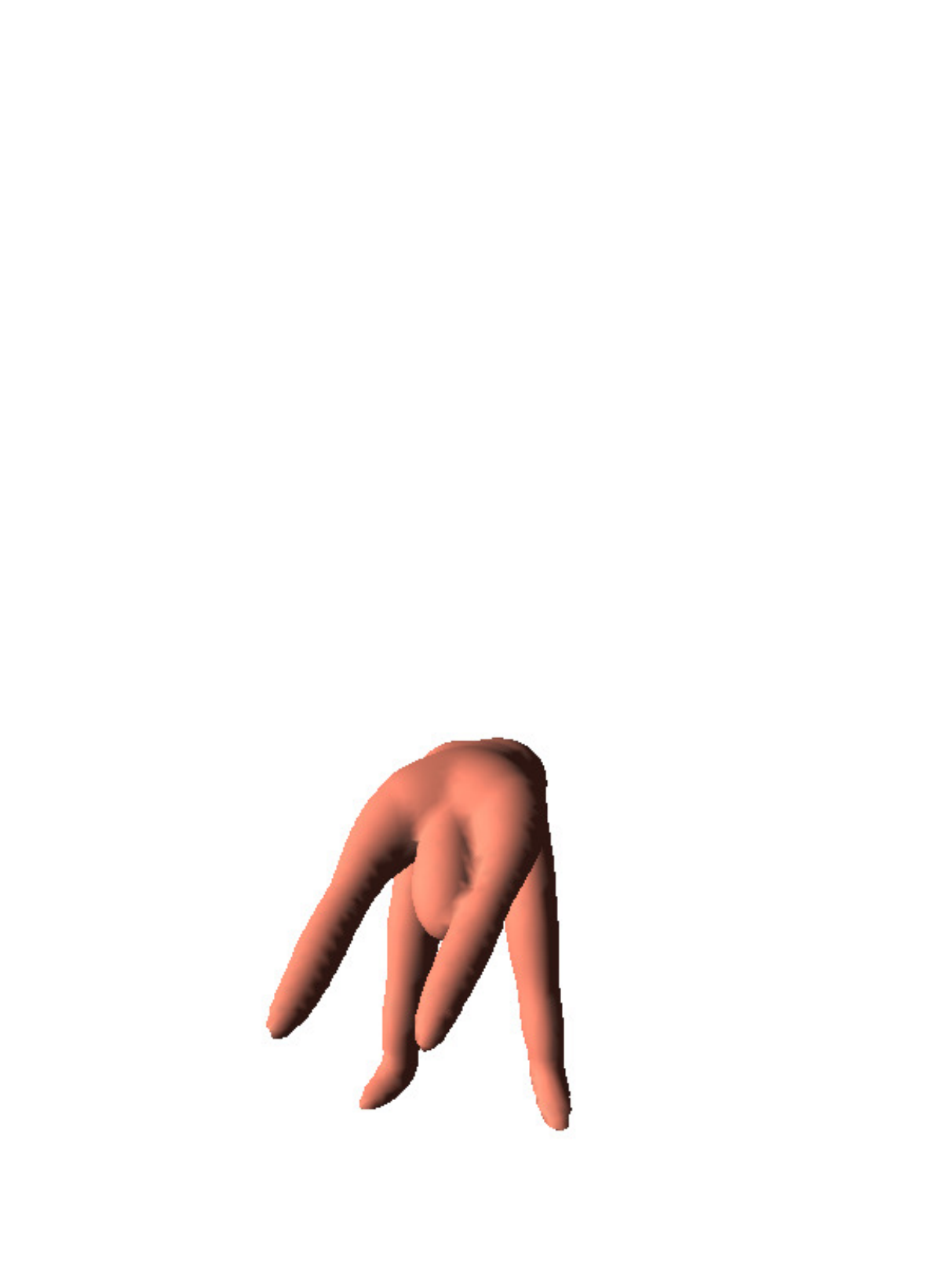}}&
\frame{\includegraphics[width=0.12\textwidth]{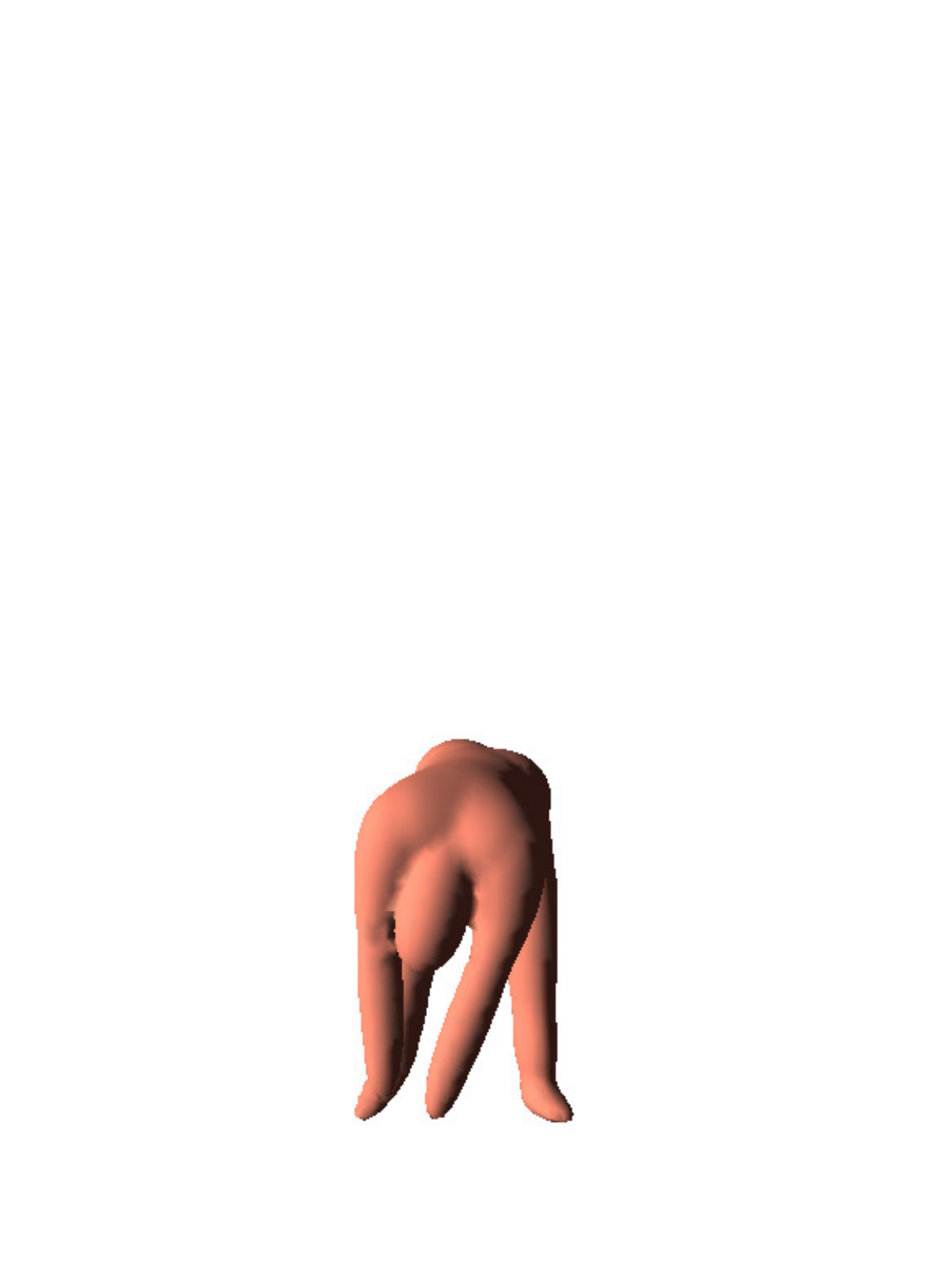}}&
\frame{\includegraphics[width=0.12\textwidth]{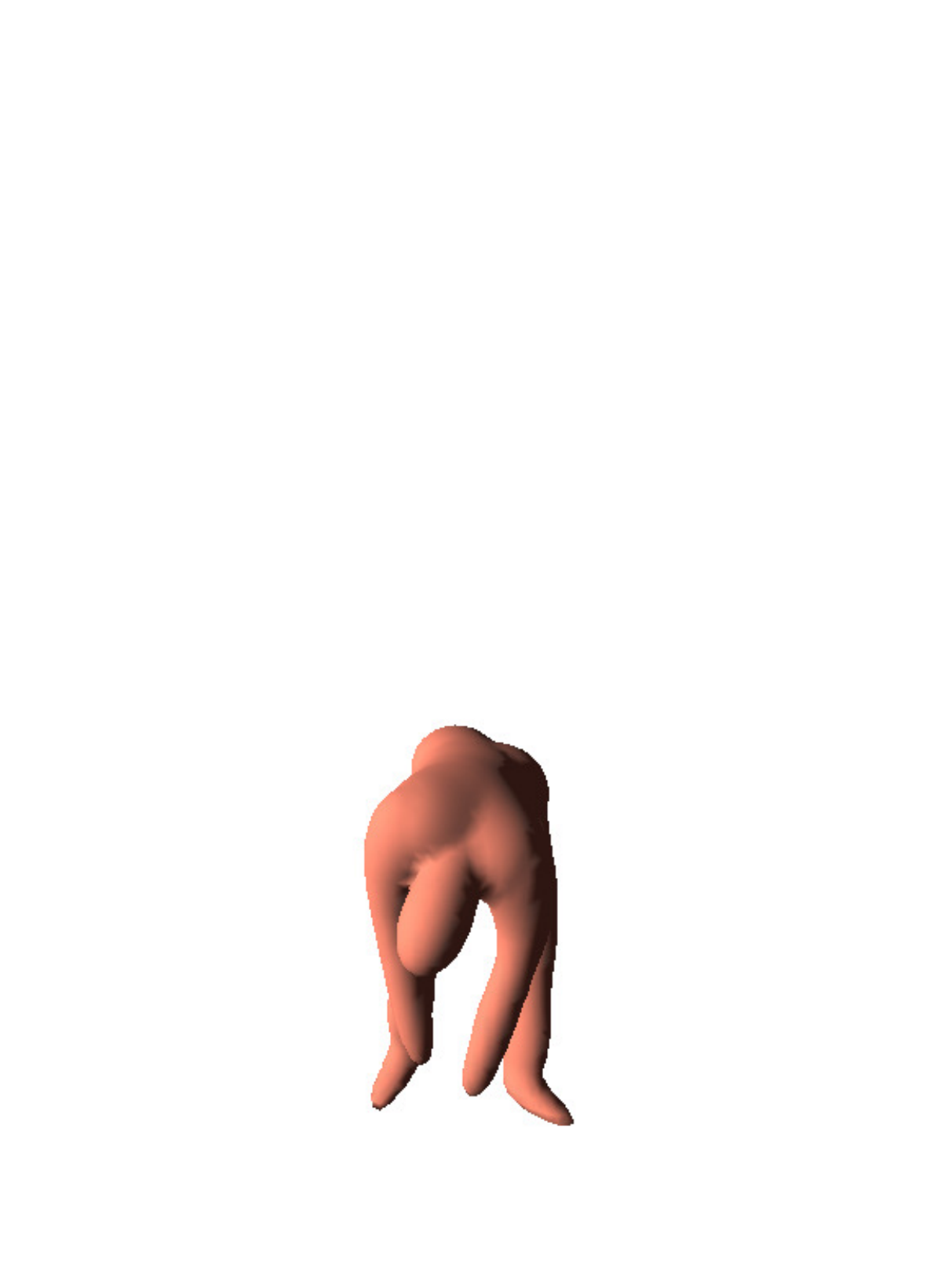}}
\\
\end{tabular}
\caption{The leaning sequence: Images and silhouettes associated with
  the first camera, and the fitted implicit surface.} 
\label{fig:Erwan-leaning-rawdata}
\end{center}
\end{figure*}

\subsection{Experiments with multiple-video data}
\label{section:experiments-video}

The experimental data that we used for validating the human motion
tracker was gathered with six calibrated and finely synchronized cameras. 
Each camera delivers 780$\times$580 color images at 28 frames per
second with a synchronization accuracy within 1 mili-second. The
figures below show these image sequences sampled at 14 frames per second.

We
applied articulated human motion tracking  to two multiple-image
sequences, the {\em taekwendo} sequence shown 
on Figure~\ref{fig:Ben-taekwendo-rawdata} and the {\em leaning}
sequence shown of Figure~\ref{fig:Erwan-leaning-rawdata}. The first
data set is composed of 6$\times$700 frames while the second one is composed of
6$\times$200 frames.
We
used the same body-part dimensions for the two characters.
One may notice that the
silhouettes have holes and missing pieces, which results in the presence of
3-D outliers. 
The top row in Figure~\ref{fig:Ben-taekwendo-implicit-model} shows the
3-D points and normals that were reconstructed from the imperfect
silhouettes; The middle row shows the articulated implicit surface
resulting  from application of our method, while the bottom row shows
the same surface resulting from application of our method in
conjunction with the algebraic distance.
Obviously, in
this last case, there
is a discrepancy between the data and the fitted model: the recovered motion of
the right feet and the right thigh are incorrect.
Similarly, figure~\ref{fig:Erwan-leaning-rawdata}
shows a sample of the {\em leaning} sequence, the
corresponding silhouettes, and the fitted model using the proposed method.

\section{Conclusions} 
\label{section:conclusion}

In this paper we described a method for tracking articulated
motion with several cameras. We introduced a new metric that measures
the discrepancy between
observations (composed of both 3-D points and 3-D normals) and an
articulated implicit
surface. This metric is more powerful than previously used distance
functions because it allows for less ambiguous associations between
the data and the model. Moreover, it is well suited when one deals
with either visual-hull or visual-shape representations of the data. 

We cast the data-to-model fitting process into a robust probabilistic
framework. We showed that there is a strong similarity between the
mathematical representation of an implicit surface and a mixture of
Gaussian distributions. 
We explored this similarity and we showed that
the articulated motion tracking problem can be formulated as maximum
likelihood with hidden variables. We added a uniform component to the
mixture to account for outliers.
We formally derived an algorithm that computes ML
estimates for the motion parameters within the framework of
expectation-maximization (EM).  
Therefore, the tracker may well be interpreted in the framework of
robust data
clustering, where the observations are assigned to one of the
ellipsoids, or to an outlier component.

There are many questions that remain open and that we plan to
investigate in the near future: The algorithm may be trapped in local
minima if it is not properly initialized; There are some similarities between
our robust tracker and the use of M-estimators and these similarities diserve
further investigation. There are other interesting issues such as:
A thorough and
quantitative evaluation of the results and their comparison with
marker-based motion capture systems and the possibility to capture several
articulated motions at once. 

\vspace{-5mm}
\bibliographystyle{plain}
%\input{bibfile}

%\bibliography{../bibfiles/boyer,../bibfiles/triggs,../bibfiles/horaud,../bibfiles/books,../bibfiles/general,../bibfiles/manuals,../bibfiles/articulated,../bibfiles/silhouette,../bibfiles/guillaume,../bibfiles/new-citations}

\end{document}